\documentclass[journal]{IEEEtran}
\usepackage{url}
\usepackage{xcolor}
\usepackage{graphicx}
\usepackage{amsmath}
\usepackage{amssymb}

\ifCLASSINFOpdf
\else
\fi

\begin{document}

\title{Advances in adversarial attacks and defenses in computer vision: A survey}
\author{Naveed~Akhtar,  Ajmal~Mian, Navid Kardan and Mubarak~Shah
\thanks{Naveed~Akhtar and Ajmal~Mian are with the Department of Computer Science and Software Engineering, University of Western Australia, 35 Stirling Highway, Crawley 6009, WA, Australia.}
\thanks{Mubarak Shah and Navid Kardan are with the Center for Research in Computer Vision, University
of Central Florida, Orlando, FL 32816, United States.}
}


\maketitle

\begin{abstract}
Deep Learning (DL) is the most widely used tool in the contemporary field of computer vision. Its ability to accurately solve complex problems is employed in vision research to learn deep neural models for a variety of tasks, including security critical applications. However, it is now known that DL is vulnerable to adversarial attacks that can manipulate its predictions by introducing visually imperceptible perturbations in images and videos. Since the discovery of this phenomenon in 2013~\cite{szegedy2013intriguing},  it has attracted significant attention of  researchers from multiple sub-fields of machine intelligence. In \cite{akhtar2018threat}, we reviewed the contributions made by the computer vision community in adversarial attacks on deep learning (and their defenses) until the advent of year 2018. Many of those contributions have inspired new directions in this area, which has matured significantly since witnessing the first generation methods. Hence, as a legacy sequel of~\cite{akhtar2018threat},  this literature review focuses on the advances in this area since 2018. To ensure  authenticity, we mainly consider peer-reviewed contributions published in the prestigious sources of computer vision and machine learning research.  
Besides a comprehensive literature review, the article also provides concise definitions of technical terminologies for non-experts in this domain. 
Finally, this article discusses challenges and future outlook of this direction based on the literature reviewed herein and~\cite{akhtar2018threat}.   
\end{abstract}


\IEEEpeerreviewmaketitle

\section{Introduction}
\IEEEPARstart{D}{eep Learning} (DL)~\cite{lecun2015deep} is a data driven technology that can precisely model  complex mathematical functions over large data sets. It has recently provided scientists with numerous  breakthroughs in machine intelligence applications. From analysing mutations in DNA~\cite{xiong2015human} to reconstruction of brain circuits~\cite{helmstaedter2013connectomic} and exploring cell data~\cite{amodio2019exploring}; deep learning methods are currently advancing our knowledge for many cutting-edge scientific problems.
Thus, it is not surprising that multiple contemporary sub-fields of  machine intelligence are fast adopting this technology as `the tool' to solve their long-standing problems. Along speech recognition~\cite{hickok2007cortical} and natural language processing~\cite{manning1999foundations}, computer vision is one of the sub-fields that currently relies heavily on deep learning.  

The rise of deep learning in computer vision was triggered by the seminal work of Krizhevsky  et  al.~\cite{krizhevsky2012imagenet} in 2012, reporting a record performance improvement on a hard image recognition task~\cite{deng2009imagenet} using a Convolutional Neural Network (CNN)~\cite{lecun1989backpropagation}. 
Since \cite{krizhevsky2012imagenet}, the computer vision community has contributed significantly to deep learning research, which has led to increasingly powerful neural networks~\cite{he2016deep}, \cite{huang2017densely}, \cite{szegedy2016inception} that can handle a large number of layers in their architectures -  establishing the essence of `deep' learning.   
The advances made in the context of computer vision have also enabled deep learning to solve complex problems of Artificial Intelligence (AI). For instance, one of the crowning achievements of the modern AI, i.e.~tabula-rasa learning~\cite{silver2017mastering} owes a fair share to Residual Learning~\cite{he2016deep}, which originated in the field of computer vision.  

\begin{figure}
    \centering
    \includegraphics[width = 2in]{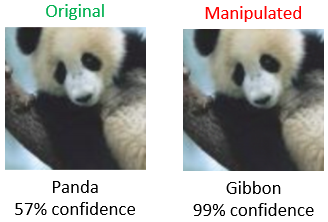}
    \vspace{-2mm}
    \caption{Attacking a deep visual model (GoogLeNet~\cite{szegedy2015going} here) by imperceptible image manipulation results in incorrect prediction with high confidence. FGSM attack~\cite{goodfellow2014explaining} is used here to manipulate the image. }
    \label{fig:teaser}
    \vspace{-3mm}
\end{figure}

Owing to the (apparent) super-human abilities of deep learning~\cite{silver2017mastering}, computer vision-based AI is believed to have reached the maturity required for deployment in safety and security critical systems.
Auto-pilots of vehicles~\cite{Tesla:2020}, facial recognition in ATMs~\cite{ATM:2020} and Face ID technology of mobile devices~\cite{FaceID:2020} are a few fore-running real-world examples that portray the developing faith of modern societies in computer vision solutions. With highly active deep learning-based vision research for autonomous vehicles~\cite{grigorescu2020survey}, face recognition~\cite{zulqarnain2018learning}, \cite{masi2018deep}, robotics~\cite{sunderhauf2018limits} and surveillance systems~\cite{najafabadi2015deep} etc., we can anticipate the \textit{omnipresence} of deep learning in security critical computer vision applications. However, serious concerns are now  emerging for this prospect due to an unsought discovery of adversarial vulnerability of deep learning \cite{szegedy2013intriguing}. 



\begin{figure*}[t!]
    \centering
    \includegraphics[width = 0.95\textwidth]{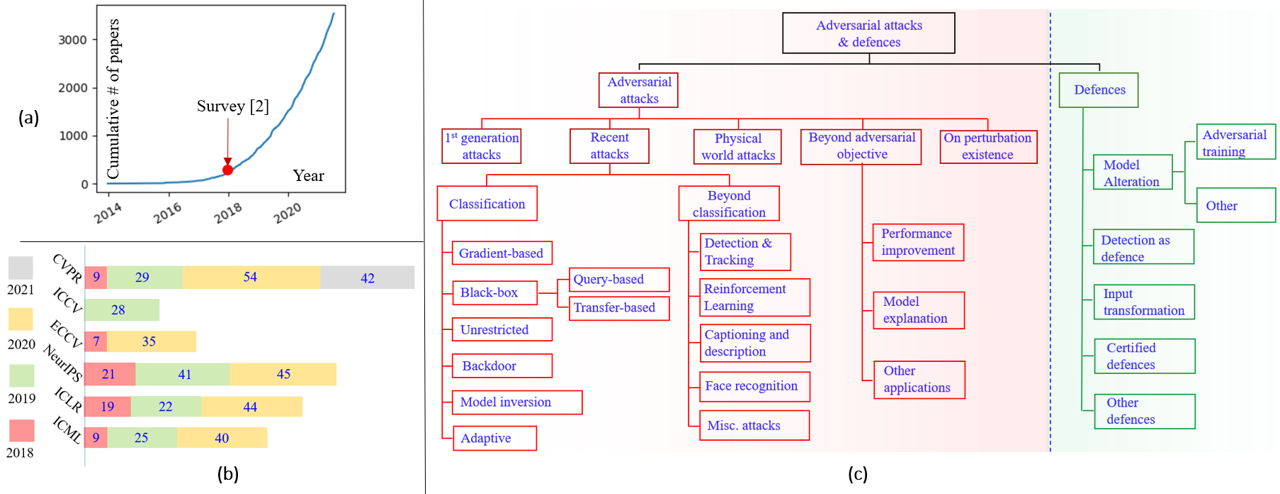}
    \caption{\textbf{(a)} Cumulative number of adversarial attacks and defense papers appearing on arXiv in recent years (data from~\cite{Carlini:20}). Over 3,000 papers have appeared since the first survey article~\cite{akhtar2018threat}. \textbf{(b)} An increasing number of publications in this direction is experienced by the leading research sources of computer vision and machine learning. The bar chart indicates the total number of papers appearing per year which include `adversarial', `attack' or `defense' keyword in their title, while the paper-content directly focuses on adversarial attack or defense problem. \textbf{(c)} Structuring of the literature reviewed in the  article. The survey covers both aspects of attacks and defenses with emphasis on the attack methods. 
    }
    \label{fig:intro}
    \vspace{-3mm}
\end{figure*}

It was discovered by Szegedy et al.~\cite{szegedy2013intriguing} that deep neural network predictions can be manipulated with extremely low magnitude input perturbations.  
For images, these perturbations can be restricted to the imperceptible regime of human vision system, yet they can completely alter the output predictions of a deep visual model (see Fig.~\ref{fig:teaser}).
Originally, these manipulative signals were discovered for the image classification task~\cite{szegedy2013intriguing}. However, their existence is now well-established for a variety of mainstream computer vision problems, e.g.~semantic segmentation~\cite{arnab2018robustness}, \cite{he2019segmentations}; object detection~\cite{tu2020physically}, \cite{zhang2019towards}; and  object tracking~\cite{jia2019fooling}, \cite{chen2020one}. The literature highlights numerous characteristics  of adversarial perturbations, that make them a real threat to deep learning as a pragmatic technology. For instance, it is repeatedly observed that the attacked models generally show high confidence on the wrong predictions of the manipulated images~\cite{akhtar2018threat}, \cite{goodfellow2014explaining}. It is  also established that the same perturbation can often fool multiple models~\cite{zheng2020efficient}, \cite{zhou2018transferable}. The literature  has also witnessed pre-computed perturbations, known as universal perturbations, that can be added to `any' image to fool a given model with high probability~\cite{moosavi2017universal}, \cite{akhtar2019label}. These facts have profound implications for security critical applications, especially when it is widely believed that deep learning solutions have predictive prowess that can surpass human abilities 
~\cite{silver2017mastering},~\cite{vinyals2019grandmaster}. 

Due to its critical nature, the topic of adversarial attacks (and their defenses) has  received considerable attention of the research community in the last five years.   
In \cite{akhtar2018threat}, we surveyed the contributions surfaced in this direction until the advent of 2018. Most of those works can be seen as the \textit{first-generation} techniques that explore the core algorithms and techniques to fool deep learning or defend it against the adversarial attacks. Some of those algorithms have inspired streams of followup methods that further refine and adapt the core attack and defense techniques. These \textit{second-generation} methods are also found to focus more on other vision tasks instead of just the classification problem, which is the main topic of interest in early contributions in this direction.

Since 2018, there has been an ever increasing number of publications in this research direction (see Fig.~\ref{fig:intro}-a,b). Naturally, these publications also include instances of literature reviews, e.g.~\cite{yuan2019adversarial}, \cite{hao2020adversarial}, \cite{ozdag2018adversarial}, \cite{zhou2019adversarial}, \cite{vakhshiteh2020threat}. The literature survey we provide here differs from the existing reviews 
in many ways. This article is unique in that it is a legacy sequel of~\cite{akhtar2018threat} -  the first-ever  peer-reviewed literature survey 
on  this topic. 
Subsequent reviews, e.g.~\cite{zhou2019adversarial} are often found to be closely following \cite{akhtar2018threat}; or building on \cite{akhtar2018threat} for specific problems~\cite{vakhshiteh2020threat}. In recent years, this  direction has matured significantly within the  field of computer vision. By building on the insights of~\cite{akhtar2018threat} and subsequent literature, we are able to  provide more precise definitions of the technical terminologies for this fast developing research direction. This also resulted in a more coherent structure of  literature reviewed in this article, for which we provide concise discussions based on the current understanding of the terminologies by the research community. Moreover, we focus on peer-reviewed publications appearing in the prestigious research publication venues of computer vision and machine learning. Focusing on the leading contributions allows us to provide a more clear outlook of this  direction for computer vision and machine learning researchers. Not to mention, this article reviews the most recent contributions of this fast evolving area to provide the most comprehensive review in this direction to date.

The rest of the article is organized as follows. In Section~\ref{sec:def}, we provide  definitions of technical terminologies used in the rest of the article. In Section \ref{sec:PF}, we formulate the broader problem of adversarial attacks. The first generation of the attacks are discussed in Section~\ref{sec:firstGen}, followed by the recent attacks focusing on the classification problem in Section~\ref{sec:RA}. We focus on recent attacks beyond classification problem in Section~\ref{sec:BeyondClassification}, and on the attacks tailored to the Physical world in Section~\ref{sec:RealWorld}. Contributions focusing more on the theoretical aspect of the existence of adversarial examples are discussed in Section~\ref{sec:On}. Recent defense methods are the topic of Section~\ref{sec:defense}. The article reflects on the literature trends in Section~\ref{sec:Disc}, where it also provides a discussion on the outlook of this research direction and future venues. Finally, we conclude in Section~\ref{sec:Conc}. 
%


\vspace{-1mm}
\section{Definition of terms}
\label{sec:def}
To provide a clear discussion on the literature, it is imperative to first specify precise   definitions of the  technical terminologies commonly appearing in publications. Currently, the domain of adversarial machine learning is evolving rapidly. Hence, understanding of the related technical terms is also evolving in the research community. Arranged alphabetically below, we provide definitions of the frequently encountered  terminologies in the related literature, as widely understood by the computer vision (and machine learning) community. The same definitions of the concepts are followed in the rest of this  article.

\begin{itemize}
    \item \textit{Adversarial example/image} is an image that is intentionally manipulated to cause incorrect model prediction. It is generally computed by adding \textit{adversarial perturbation} to a natural image. 
    \textit{Clean}, \textit{natural} or \textit{benign} image are the commonly  used terms to describe the opposite of an \textit{adversarial} image.
     \item \textit{Adversarial perturbation} is the component of an adversarial image that causes the  incorrect prediction. Commonly, it is a low magnitude additive noise-like signal. However, exceptions are possible.
     \item \textit{Adversarial training} is a process that injects adversarial examples in the training data of a model to make it adversarially robust.
     \item \textit{Adversary} is the agent (i.e.~the attacker)  creating an adversarial example. Alternatively, the adversarial signal/perturbation is also referred to as the adversary, albeit much less often.   
     \item \textit{Attack detector} is an external mechanism for a model to (only) identify an input as adversarial or clean. 
    \item \textit{Black-box attack} assumes no knowledge of the target model. More strictly, the adversary is unaware of its training process and parameters. One category of black-box attacks allows \textit{probing} the deployed target models with queries. This setup is more commonly known as \textit{query-based} attack. To distinguish from the query-based attacks, other black-box attacks are sometimes also referred to as   \textit{Zero-knowledge attacks}. Opposite of black-box attack is \textit{white-box} attack - see the definition below. 
    \item \textit{Data membership attack} aims to identify if a sample was used in the training of a model or not. 
    \item \textit{Defense/adversarial defense} is a broader term used for any mechanism of inducing inherent robustness in a model, or  external/internal mechanisms to detect adversarial signals, or  image processing to negate adversarial effects of input manipulations. \textit{Adversarial robustness} is the preferred alternate term for the techniques focusing on inducing inherent resilience in the models, e.g.~with adversarial training.     
    \item \textit{Digital attack} assumes that the adversary has full access to the actual digital input to the model. Most of the existing adversarial attacks are digital attacks. The opposite of digital attack is \textit{Physical (world) attack} - see definition below.  
    \item \textit{Evasion attack} is a broader term for the adversarial attacks that fool pre-trained models into misclassifying input images at `test time'. Poisoning attack (see below) is its close antonym that poisons a model during `training'.  
    \item \textit{Fooling rate/ratio} is the commonly used evaluation metric, defined as the percentage of adversarial images on which the target model prediction is incorrect. 
    \item \textit{Gradient-based attacks} involve gradient computation of a model's cost surface (or intermediate internal representation) with respect to the input. White-box attacks are predominately gradient-based.  
    \item \textit{Gradient-free attacks} do not involve gradient computation of any model. 
    \item \textit{Gray-box attack} assumes partial knowledge of the target model. However, since partial knowledge may actually lead to more knowledge, we prefer the term \textit{restricted knowledge white-box} over the \textit{gray-box} in this review. Under this nomenclature, gray-box attacks form a sub-category of the white-box attacks. 
    \item \textit{Image-specific attack} is computed  to fool a target model on a specific image. Close antonyms for this term are \textit{universal attack} and \textit{label universal attack}. 
    \item \textit{Insertion attacks} insert an adversarial object (or a well-localised visible pattern) in an image, e.g.~adversarial patch to alter the model prediction.  
    \item \textit{Label universal (adversarial) attack} aims at class-specific fooling. It computes an  additive perturbation that has a pronounced effect on all samples of a selected class. 
    \item \textit{Model extraction attack} aims at recovering information about a target model (e.g.~its classification boundaries) to subsequently use the information for fooling it. 
    \item \textit{Model inversion attack} aims at reconstructing individual training samples of the target model.  
    \item \textit{Norm-bounded perturbations} restrict the $\ell_p$-norm of additive adversarial perturbations to control their perceptibility in adversarial examples. An overwhelming majority of the additive adversarial perturbations is norm-bounded.     
    \item \textit{One-step methods} compute perturbations in a single step, as opposed to \textit{iterative methods} that use multiple iterations in their algorithm. These terms are generally more relevant to white-box attacks. 
    \item \textit{Physical (world) attacks} do not assume any access to the digital representation of the target model's input. Adversarial examples are `clean' images of e.g.~physically modified or adversarially illuminated objects. 
    \item \textit{Poisoning attack} causes a model to misbehave when exposed to a trigger in the input. This (mis-)behavior is programmed into the model by manipulating the training process with tampered training data or algorithm. Generally,  \textit{trojan} or \textit{backdoor} attack are used as synonyms for poisoning attack. This article largely focuses on the attacks (and their defenses) launched on clean pre-trained models. Hence, poisoning attack is not a direct topic for this survey.
    \item \textit{Quasi-imperceptible perturbations} introduce slight visual impairment to images. This is in contrast to the  imperceivable changes induced by imperceptible perturbations.   
    \item \textit{Query-based attack} is a form of black-box attack where the attacker is able to query the target model and exploit its output to optimize adversarial image(s). It either treats the target model as an \textit{oracle} or learns a substitute model (see below) to be used as an oracle to subsequently generate adversarial images.
    A \textit{decision/boundary-based} attack is a specific form of query-based attacks that assumes knowledge of only the predicted labels (not confidence scores) of the target model. The query-based attacks that also exploit confidence scores of the target model are termed \textit{score-based} attacks. 
    \item \textit{Real-world attacks} are evaluated in practical conditions by attacking real-world systems, as opposed to the bare models in laboratory setup. These attacks may still be digital or physical. 
    \item \textit{Targeted attack} forces the output of a model to pre-specified prediction of adversary's choice, as opposed to random incorrect prediction in the case of \textit{non-trageted attack}.  
    
    \item \textit{Target image} is the clean image being manipulated by the adversary. 
    \item \textit{Target model} is the model under attack.
    \item \textit{Target label} is the (desired) incorrect label of the adversarial example. The term is more relevant for targeted attacks.
    \item \textit{Threat model} refers to the assumed collective adversarial conditions against which a defense mechanism is designed and tested to verify  its effectiveness. 
    \item \textit{Transferability} is the ability of an adversarial example/perturbation to generalise beyond the model for which it was originally computed.  
    \item \textit{Substitute model} is a model trained by an adversary to replicate the prediction behavior of the target model. \textit{Surrogate model} and \textit{auxiliary model} are the commonly used synonyms for the term substitute model.
    \item \textit{Universal (adversarial) perturbations} are image-agnostic manipulative signals that can alter the model prediction on any input with high probability. 
    \item \textit{Unrestricted adversarial attacks}  replace a natural image with a (synthetically) generated adversarial image, such that the latter has the same semantic meaning as the former for humans but not for the target model\footnote{This understanding of the term is slightly different from~\cite{brown2018unrestricted} and relates more to \cite{song2018constructing} that allows a clearer delineation between the unrestricted and conventional adversarial examples. }. 
    \item \textit{White-box attack} assumes complete knowledge of the target model. We refer to the attacks that assume partial knowledge of the target model or its training process, as \textit{restricted knowledge white-box attacks}. Such attacks differ from the black-box attacks in that the latter only assume the knowledge of `prediction' made by the model. The prediction may include a single/set of labels or a single/set of confidence scores. Any further, but incomplete knowledge makes the attack restricted knowledge white-box attack.
 \end{itemize}
\section{Adversarial attacks: The formal problem}
\label{sec:PF}

Let $\mathcal{M}(.)$ be the target deep visual model such that $\mathcal{M}({\bf I}): {\bf I} \rightarrow \boldsymbol\ell$, where ${\bf I} \in \mathbb R^m$ is a natural image 
and $\boldsymbol\ell \in \mathbb Z^+$ is the output  of the model. In the most common form of  adversarial attacks, the adversary seeks a signal $\boldsymbol\rho \in \mathbb R^m$ to achieve $\mathcal{M}({\bf I} + \boldsymbol\rho) \rightarrow \tilde{\boldsymbol\ell}$, where $\tilde{\boldsymbol\ell} \neq {\boldsymbol\ell}$. To ensure that the manipulation to a clean image is humanly imperceptible, the perturbation $\boldsymbol\rho$ is often norm-bounded, e.g.~by enforcing $|| \boldsymbol\rho||_p < \eta$, where $||.||_p$ denotes the $\ell_p$-norm of a vector and `$\eta$' is a pre-defined scalar. More concisely, the adversary seeks $\boldsymbol\rho$ that satisfies:
\begin{align}
    \mathcal{M}({\bf I} + \boldsymbol\rho) \rightarrow \tilde{\boldsymbol\ell}~~\text{s.t.}~ \tilde{\boldsymbol\ell} \neq {\boldsymbol\ell}, ||\boldsymbol\rho||_p < \eta.
    \label{eq:conventional}
\end{align}

The formulation above underpins the most prevailing contemporary understanding of the adversarial attacks. Yet, it does not encompass all attacks. For instance,  unrestricted adversarial examples~\cite{brown2018unrestricted}, \cite{song2018constructing}, where the adversary is neither restricted 
to manipulate the original image (i.e.~the image itself can be replaced) nor concerned with limiting the perturbation norm, can not be described by the constraint in~(\ref{eq:conventional}). Similarly,
the addition of a localized, but perceivable adversarial pattern in an image (e.g.~adversarial patch~\cite{brown2017adversarial}) is not accounted for by~(\ref{eq:conventional}).  
Hence, for comprehensiveness, we also consider a more broader constraint, given as 
\begin{align}
    \mathcal{M}(\tilde{\bf I}) \rightarrow \tilde{\boldsymbol\ell}~~\text{s.t.}~ \tilde{\boldsymbol\ell} \neq {\boldsymbol\ell},~\tilde{\bf I} \in \mathcal S_{{\bf I}},~\mathcal M\left({\bf I} \sim  \{\mathcal S_{\bf I} - \tilde{\bf I}\}\right) = \boldsymbol\ell,
    \label{eq:newConstraint}
\end{align}
where $\mathcal S_{{\bf I}}$ is the set of images   \textit{perceived} as clean or allowed by humans to produce the desired output $\boldsymbol\ell$. For the sake of brevity, we are assuming a single adversarial sample in $\mathcal S_{{\bf I}}$ in (\ref{eq:newConstraint}). 
The conventional view of additive perturbations (in Eq.~\ref{eq:conventional}) becomes a special case of this constraint where $\tilde{\bf I} = {\bf I} +\boldsymbol\rho$ and  $\tilde{\bf I} \in \mathcal S_{{\bf I}}$ is ensured by restricting the perturbation norm. 
Since (\ref{eq:newConstraint}) does not deal with $\boldsymbol\rho$ explicitly, one must articulate any additional constraint over $\boldsymbol\rho$ to specify an attack under (\ref{eq:newConstraint}) - as we have done above for the imperceptible perturbation.               

Adversarial examples for deep visual models were  originally discovered for the image classification task~\cite{szegedy2013intriguing}, where \textit{additive} perturbations were used to launch the attack. 
Consequently, a vast majority of the existing attacks  leverage some form of the additive perturbations to manipulate the model output. Moreover, image classifiers still remain the most popular target models for attacks.   
This trend partially owes to the fact that classification is one of the fundamental tasks in pattern recognition. Thus, it is important to  explicitly, though briefly,  discuss the broad concept of adversarial attacks on deep image classifiers under the above formulation.  

For the image classifiers, an output is a class label $\ell \in \mathbb Z^+$. The nature of the task makes it more interesting to change this label to a pre-specified incorrect label $\tilde{\ell} \in \mathbb Z^+$ by the attack, which motivates the \textit{targeted} adversarial attacks on classifiers. A \textit{non-targeted} attack on a classifier can  also be considered as a special case of the targeted attacks, where $\tilde{\ell}$ is chosen at random. Whereas \textit{image-specific} attacks lead to misclassification of individual images, it is also possible to compute additive  perturbations $\boldsymbol\rho$ that cause incorrect label predictions on a large number of images. Such \textit{universal} perturbations were first reported by Moosavi-Dezfooli~\cite{moosavi2017universal}.  Here, we discuss the notions of image-specific vs universal, and targeted vs non-targeted attacks in the context of classifiers for a clear understanding of the text to follow immediately. Nevertheless, these concepts are more general and can also be applied to other computer vision tasks.
\section{First-generation attacks}
\label{sec:firstGen}
In the context of this survey, as a legacy sequel of~\cite{akhtar2018threat}, the first generation of adversarial attacks include the most influential contributions surfacing before 2018, which inspired series of followup methods. These attacks focus more on the  fundamental algorithms to compute adversarial images, using image classification task as the test bed. We discuss these methods upfront as a separate section for two main reasons. First, by organizing the discussion on these methods in a (roughly) chronological order, we also provide the readers with a historical account of this research direction. Second, describing these seminal  works  early provides a more clear understanding of the inspiration of the more recent techniques.  

\begin{figure}[t!]
    \centering
    \includegraphics[width = 0.45\textwidth]{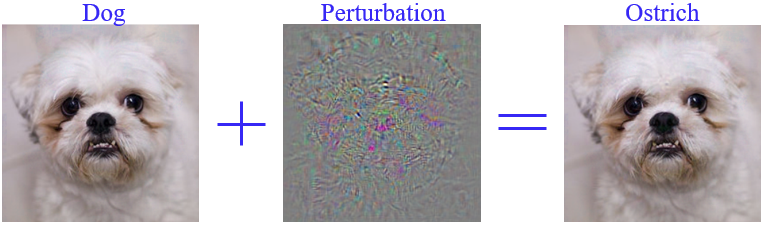}
    \caption{Szegedy et al.~\cite{szegedy2013intriguing} were the first to demonstrate imperceptible perturbations to images to fool deep learning. Here, the image of a `dog' is confused as `ostrich' by AlexNet~\cite{krizhevsky2012imagenet} when the shown perturbation is added to it. The perturbation is exaggerated for visualisation.}
    \label{fig:Szegedy}
\end{figure}

\vspace{1mm}
\noindent{\bf \textit{The L-BFGS Attack:} } 
Szegedy et al.~\cite{szegedy2013intriguing} first discovered the vulnerability of deep visual models to adversarial perturbations by solving for the following optimisation problem:
\begin{align}
\min_{\boldsymbol\rho} || \boldsymbol\rho ||_2 \hspace{2mm} \text{s.t.}~\mathcal M( {\bf I}  +  \boldsymbol\rho) = \tilde{\ell};~{\bf I}  +  \boldsymbol\rho \in [0,1]^m.  
\label{eq:Intriguing}
\end{align}
The above is a hard problem, for which an approximate solution is computed by Szegedy et al.~with the
Limited Memory  Broyden–Fletcher–Goldfarb–Shanno (L-BFGS) algorithm, which is a quasi-Newton algorithm involving computation of inverse Hessian~\cite{fletcher2013practical} - inspiring the name of the attack adopted in the subsequent literature. 
To solve (\ref{eq:Intriguing}), the  constraint $\underset{\boldsymbol\rho}{\min}||\boldsymbol\rho ||_2$ is combined using a Lagrangian multiplier `$c$' and the solution is computed by estimating the smallest $c>0$ for which the minimizer $\boldsymbol\rho$ of the problem~(\ref{eq:approxSzegedy}) satisfies $\mathcal M({\bf I}  +  \boldsymbol\rho) = \tilde{\ell}$.
\begin{align}
\min_{\boldsymbol\rho}~~c|\boldsymbol\rho| + \mathcal{L}({\bf I}  +  \boldsymbol\rho, \tilde{\ell})~~s.t.~{\bf I}  +  \boldsymbol\rho \in [0,1]^m,
\label{eq:approxSzegedy}
\end{align}
where $\mathcal L(.,.)$ is the classifier loss. Manipulation of a clean image with the additive perturbation resulting from (\ref{eq:approxSzegedy}) remains imperceivable to the human visual system, see Fig.~\ref{fig:Szegedy}. This observation had a profound impact on the vision research community, which was fast developing the impression that deep visual features well approximate the perceptual differences in images with Euclidean distances. Discovery of adversarial perturbations that could  completely alter the decisions of deep visual models with minuscule Euclidean norm revised this impression.  
Szegedy et al.~also demonstrated that their  adversarial attack transfers well between different deep visual classifiers. This intriguing vulnerability of deep learning to adversarial attacks attracted a wide interest of  researchers in the subsequent years.    

\vspace{1mm}
\noindent{\bf \textit{The FGSM Attack:} }
It was originally observed by Szegedy et al.~\cite{szegedy2013intriguing} that including adversarial images in the training data of a classifier improves its robustness to adversarial examples. Reinforced by multiple followup works, this observation is the main motivation behind the idea of \textit{adversarial training} in the  literature. However, solving (\ref{eq:approxSzegedy}) for a large number of images  is computationally prohibitive. This inspired the Fast Gradient Sign Method (FGSM)~\cite{goodfellow2014explaining} to efficiently compute  adversarial perturbations as:
\begin{align}
\boldsymbol\rho = \epsilon~\text{sign}\left( \nabla_{{\bf I}}~\mathcal J_{\boldsymbol\theta} ({\bf I}, \ell)  \right),
\label{eq:FGSM}
\end{align}
where $\mathcal J_{\boldsymbol\theta}(.,.)$ 
is the cost for the model with parameters $\boldsymbol\theta$, $\nabla_{{\bf I}}$ computes its gradient \textit{w.r.t.}~${\bf I}$, sign(.) denotes the sign function which is applied to each element in a vector, and $\epsilon$ is a pre-fixed scalar value to control perturbation perceptibility. The adversarial image is finally computed as $\tilde{\bf I} = {\bf I} +\boldsymbol\rho$. 

The FGSM is a one-step gradient-based method that computes norm-bounded perturbations, focusing on the `efficiency' of perturbation computation rather than achieving high fooling rates. Goodfellow et al.~\cite{goodfellow2014explaining} also used this attack to corroborate their linearity hypothesis, which considers the linear behavior of the modern neural networks in high dimension spaces (induced by e.g.~ReLUs) as a sufficient reason for their vulnerability to adversarial perturbations. They also advocated this behavior as a major cause of transferability of the attacks between different modern networks, as their  architectures pervasively allow such linearity for  training efficiency.  
At the time, the linearity hypothesis was in sharp contrast to the developing idea that adversarial vulnerability was a result of high `non-linearity' of the complex modern networks.     

The FGSM~\cite{goodfellow2014explaining} is among the most influential attacks in the existing literature, especially in the white-box setup. Its core concept of performing gradient ascend over the model's loss surface to fool it, is the basis for a plethora of adversarial attacks.
Many follow-up attacks can be strongly related to the original idea of FGSM. For instance, the Fast Gradient Value Method (FGVM) of Rozsa et al.~\cite{rozsa2016adversarial} mainly removes the sign function from (\ref{eq:FGSM}) to launch the attack. Similarly, ignoring the sign function, Miyato et al.~\cite{miyato2018virtual} normalised the gradient with its $\ell_2$-norm to launch the attack. Kurakin et al.~\cite{kurakin2016adversarial} also analysed the normalisation with $\ell_{\infty}$-norm. They also extended the FGSM to I-FGSM - its iterative variant, which is subsequently enhanced to incorporate momentum during the iterative optimisation by Dong et al.~\cite{dong2018boosting}. Their technique is  known as Momentum Iterative (MI-)FGSM. Diverse Input I-FGSM, i.e.~DI$^2$-FGSM~\cite{xie2019improving} is another example of the attacks that directly builds on FGSM. The main idea of \cite{xie2019improving} is to diversify the input used in each iteration of the iterative FGSM by applying image transformations, such as random resizing and padding, with a fixed probability. This diversification is claimed to facilitate better transferability of the resulting attack in a  black-box setup. The authors also extend DI$^2$-FGSM to 
 M-DI$^2$-FGSM by incorporating the momentum following~\cite{dong2018boosting}.  

In the above discussion, we consider FGSM as the first generation attack that inspired the followup works. It is emphasized that the discussed follow-up contributions do not form an exhaustive list of the methods that largely  build on FGSM by far. Other such methods will keep appearing in the remaining article. 

\begin{figure}
    \centering
    \includegraphics[width = 0.45\textwidth]{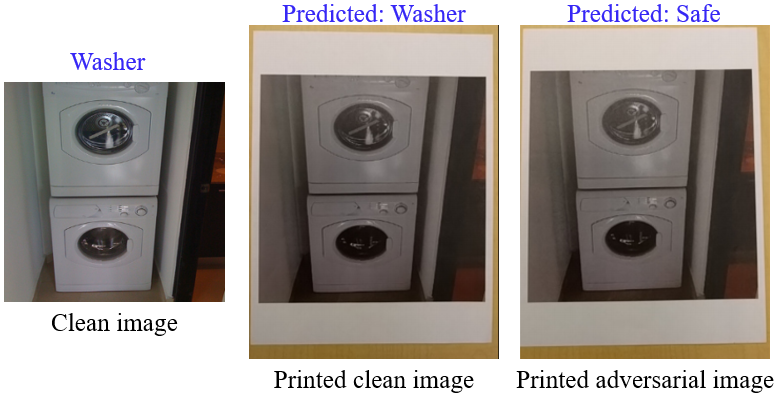}
    \caption{Kurakin et al.~\cite{kurakin2016adversarial} first demonstrated adversarial attack in the physical world by fooling a classifier on a printed adversarial image. Printed clean image of `Washer' is predicted correctly, but printed adversarial image is predicted as `Safe' by the TensorFlow Camera app used by \cite{kurakin2016adversarial}.}
    \label{fig:Kurakin1}
\end{figure}

\vspace{1mm}
\noindent{\bf \textit{The BIM \& ILCM Attacks:}} 
Though closely building on the FGSM~\cite{goodfellow2014explaining} as the original concept of iterative FGSM, the Basic Iterative Method (BIM)~\cite{kurakin2016adversarial} is also an influential contribution that introduced the Physical World attacks. The attack, which is essentially the iterative FGSM algorithm, computes an adversarial image by repeating
\begin{align}
\tilde{\bf I}_{i + 1} = \text{Clip}_{\epsilon} \left \{ \tilde{\bf I}_i + \alpha~\text{sign} (\nabla_{{\bf I}} \mathcal J_{\boldsymbol{\theta}}(\tilde{\bf I}_i, \ell) \right \},
\label{eq:BIM}
\end{align}
where `$i$' indicates the $i^{\text{th}}$ iteration, Clip$_\epsilon \{.\}$ performs clipping at $\epsilon$, and $\alpha$ is a pre-selected fixed scalar. Kurakin et al.~\cite{kurakin2016adversarial} fooled the ImageNet inception model~\cite{szegedy2016rethinking} on a mobile device by imaging \textit{printed} adversarial images in the physical world, see Fig.~\ref{fig:Kurakin1}. This idea also played its role in inspiring physical world attacks. The notion of targeted adversarial attacks can also be traced back to \cite{kurakin2016adversarial} and  \cite{kurakin2016adversarial_scale}, where it is shown that the log-probability of prediction for a target class of adversarial image can be maximised by modifying  (\ref{eq:BIM}) by changing addition to subtraction and replacing  ${\ell}$ by $\tilde{\ell}$ as:
\begin{align}
\tilde{\bf I}_{i + 1} = \text{Clip}_{\epsilon} \left \{ \tilde{\bf I}_i - \alpha~\text{sign} (\nabla_{{\bf I}} \mathcal J_{\boldsymbol{\theta}}(\tilde{\bf I}_i, \tilde{\ell}) \right \}.
\label{eq:ILCM}
\end{align}
For a classifier with cross-entropy loss, solving (\ref{eq:ILCM}) maximizes the confidence of the model on $\tilde{\ell}$ for the image $\tilde{\bf I}$. Originally, the authors proposed to use the label of the least-likely class of the clean image (as predicted by the model) as $\tilde{\ell}$ to compute interesting fooling outcomes. Hence, the technique is also referred to as Iterative Least-likely Class Method (ILCM).   

\vspace{1mm}
\noindent{\bf \textit{The PGD Attack:}} The Projected Gradient Descent (PGD) attack is widely considered as one of the most powerful attacks in the literature, while referring to the seminal work of Madry et al.~\cite{madry2017towards} as its origin. However, Madry et al.~also refer to the iterative FGSM~(\cite{kurakin2016adversarial}, \cite{kurakin2016adversarial_scale}) as a PGD method because Projected Gradient Descent is a standard optimization technique that projects gradients to a ball.  Specifically, the authors  see the iterative FGSM as the $\ell_{\infty}$-bounded PGD, in which the $\ell_{\infty}$-norm of the perturbation is bounded by the clipping operation - the projection. The main contribution of~\cite{madry2017towards} comes in the form of looking at the adversarial robustness of deep models through the lens of robust optimisation, 
 thereby defining adversarial training of deep models as a formal min-max optimisation problem below:
\begin{align}
    \min_{\boldsymbol\theta} \boldsymbol\rho(\boldsymbol\theta), ~~\text{s.t.}~~\boldsymbol\rho(\boldsymbol\theta) = \underset{({\bf I}, \ell) \sim \mathcal I}{\mathbb E} \left[\max_{\boldsymbol\rho} \mathcal L(\boldsymbol\theta, \tilde{\bf I}, \ell)  \right],
\end{align}
where $\mathbb E[.]$ is the Expectation operator and $\mathcal{I}$ is a distribution defined over the input images. This view allowed the authors to identify PGD as possibly the strongest first-order attack. 
  
From the above view, we can also look at the variants of I-FGSM discussed in the previous sections as variants of PGD. In turn, PGD can be related to FGSM. However, a crucial finding by  Madry et al.~\cite{madry2017towards} makes PGD more appealing than FGSM for adversarial training. That is, the phenomenon of `label leaking', observed in FGSM-based adversarial training~\cite{kurakin2016adversarial_scale}, does not occur for PGD-based adversarial training. In plain words,  label leaking occurs when  adversarially trained model ends up with higher prediction accuracy for adversarial images, as compared to the clean images. FGSM results in a restricted set of adversarial examples, which can lead to over-fitting in adversarial training, thereby causing  label leaking. Considering that a major objective of FGSM is to compute samples for better adversarial training, avoiding label leaking is a significant advantage of PGD.   Madry et al.~\cite{madry2017towards} also showed that adversarial training with PGD - the strongest first-order attack - automatically makes the model robust against the weaker first-order attacks, e.g.~FGSM. Nevertheless, being an iterative technique, PGD is computationally expensive.



\vspace{1mm}
\noindent{\bf JSMA \& One-pixel Attack:}  Whereas most of the early attacks focused on perturbing a clean image holistically while enforcing perturbation imperceptibility by restricting the  $\ell_2$ or $\ell_{\infty}$ norms of the perturbations, the Jacobian-based Saliency Map Attack (JSMA)~\cite{papernot2016limitations} and One-pixel attack~\cite{su2019one} deviate from this practice by restricting the perturbations to smaller regions of the image. Contrary to the convention of computing the backward-gradient of the network for perturbation estimation, as done by e.g.~FGSM and its variants, JSMA computes the forward-gradient of a network  $\mathcal M(.)$ as:
\begin{align}
    \nabla \mathcal M({\bf I}) = \frac{\partial \mathcal M({\bf I})}{\partial{\bf I}} = \left[ \frac{\partial \mathcal M_j ({\bf I})}{\partial x_i}  \right], 
    \label{eq:Jacob}
\end{align}
where $j \in 1,..., M$ for the M-dimensional function represented by $\mathcal M(.)$, $i \in 1,..., N$ for the $N$-dimensional vecrotized form of ${\bf I}$, whose $i^{\text{th}}$ element is denoted as $x_i$. Essentially, (\ref{eq:Jacob}) computes the Jacobian of the function learned by the network. Later, an adversarial extension of the saliency map~\cite{simonyan2013deep} is used by \cite{papernot2016limitations} to modify only a few selected pixels that are most influential in terms of altering the model prediction. 

Su et al.~\cite{su2019one} demonstrated that a deep visual model can even be fooled by restricting the perturbation to a single pixel. However, this is generally more effective  for the smaller image sizes, e.g.~$64\times 64$ or smaller. They used Differential Evolution (DE)~\cite{das2010differential} to estimate the location and RGB value of the pixel to be modified in the image to create an adversarial image, where the fitness criterion of the evolution is defined by accounting for the model prediction. Interestingly, the use of DE in contrast to model gradients, inherently makes their  attack a query-based black-box attack. The authors also analysed the cases of a few pixel modifications, e.g.~altering 5 instead of a single pixel for fooling. Although not originally emphasized as such, both JSMA and One-pixel attacks can be casted as optimisation problems with external constraints over the $\ell_0$-pseudo norm of the perturbations.

\vspace{1mm}
\noindent{\bf The DeepFool Attack: } Instead of restricting the perturbation norms to pre-fixed values, Moosavi-Dezfooli et al.~\cite{moosavi2016deepfool} specifically aimed at minimising the norm of the adversarial perturbation by solving:
\begin{align}
    \Delta({\bf I}; {\ell}):= \min_{\boldsymbol{\rho}} ||\boldsymbol\rho||_2~~\text{s.t.}~~\tilde{\ell} \neq \ell.
    \label{eq:deepfool}
\end{align}
The main motivation behind computing the perturbations with minimal norm was to effectively quantify the adversarial robustness of the target models, where the robustness measure was defined as:
\begin{align}
    \rho_{\text{adv}} = \mathbb E_{\bf I} \frac{\Delta({\bf I}; {\ell})}{||{\bf I}||_2},
    \label{eq:robust}
\end{align}
where $\mathbb E_{\bf I}$ is the expectation over the data distribution. 

\begin{figure}
    \centering
    \includegraphics[width = 0.4\textwidth]{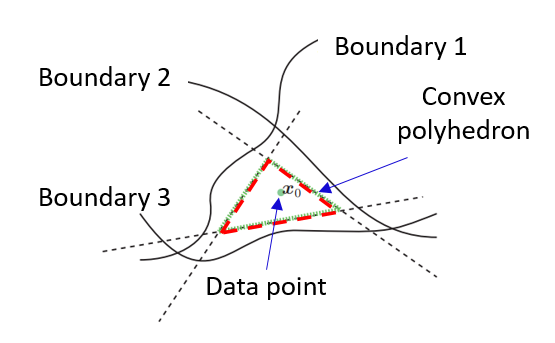}
    \caption{The DeepFool algorithm~\cite{moosavi2016deepfool} linearizes the decision boundaries around a data point to form a convex polyhedron to gradually push the point over the closest boundary for minimal perturbation.}
    \label{fig:deepfool}
\end{figure}

DeepFool is the algorithm that computes $\boldsymbol\rho$ in (\ref{eq:deepfool}) to compute the robustness defined by (\ref{eq:robust}). The iterative algorithm linearizes the class boundaries around the current image to form a convex polyhedron and pushes the image towards the closest hyperplane to change the class label, see Fig.~\ref{fig:deepfool}.  The image gets updated in each iteration with the additive perturbation. Though originally proposed to quantify model robustness, DeepFool is now generally seen as an effective image-specific adversarial attack, while overlooking the quantification aspect.  

\vspace{1mm}
\noindent{\bf The C\&W Attack:} The discovery of adversarial vulnerability of deep learning~\cite{szegedy2013intriguing} also started a parallel research direction of defenses against adversarial attacks on deep learning in 2015-16. Defensive distillation~\cite{papernot2016distillation} was a prominent technique that promised an effective solution to the problem, by building on the insights of knowledge distillation in deep networks~\cite{hinton2015distilling}. However Carlini \& Wagner~\cite{carlini2017towards} developed a set of attacks that computes norm-restricted additive perturbations that completely break defensive distillation. It is also shown that their attack is successful in fooling a defensively distilled network under black-box settings, where the perturbation is computed using an unsecured white-box model. Transferability of their attack in this setting significantly undermines the efficacy of defensive distillation.

To compute the adversarial perturbation, Carlini and Wagner solve the following optimisation problem:
\begin{align}
    \min ||\boldsymbol\rho||_p + c . f ({\bf I} + \boldsymbol\rho),~~\text{s.t.}~ {\bf I} + \boldsymbol\rho \in [0, 1]^m,
\end{align}
where $f(.)$ is a function satisfying $\mathcal M(\tilde {\bf I}) \rightarrow \tilde\ell,~\iff f ({\bf I} + \boldsymbol\rho) \leq 0$. A range of analytical forms of $f(.)$ are discussed by the authors to compute the desired perturbations. Carlini and Wagner~\cite{carlini2017towards} bounded the perturbations in their $\ell_2$, $\ell_{\infty}$ and $\ell_0$-pseudo norms, which gave rise to a set of attacks. The authors later showed that their attacks are also effective against other defense techniques~\cite{carlini2017adversarial}. The Carlini \& Wagner (C\&W) attack is generally considered a very strong attack, however, it does have a higher computational cost. 

\begin{figure}[t!]
    \centering
    \includegraphics[width = 0.35\textwidth]{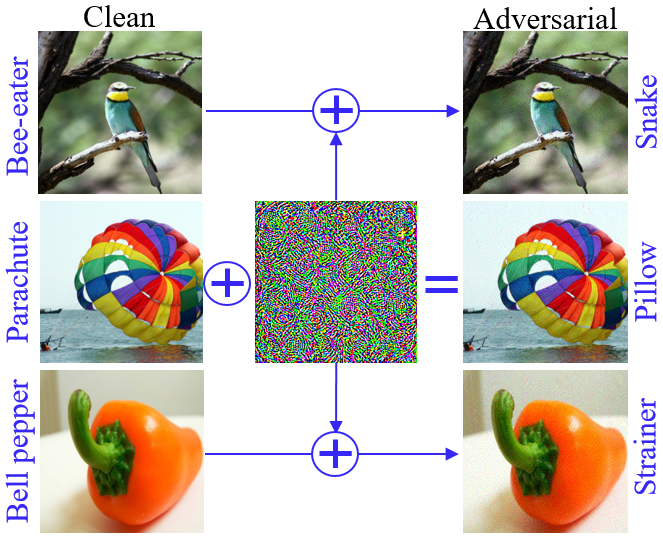}
    \caption{A single Universal Adversarial Perturbation~\cite{moosavi2017universal} can fool a model on multiple images. Fooling of GoogLeNet is shown here. These perturbations often transfer well across different models.}
    \label{fig:Universal}
\end{figure}

\vspace{1mm}
\noindent{\bf Universal Adversarial Perturbations: }
The above-mentioned methods compute adversarial perturbations that fool a target model on a specific image. Moosavi-Dezfooli et al.~\cite{moosavi2017universal} focused on computing image-agnostic perturbations that could fool the model on \textit{any} image with a high probability, see Fig.~\ref{fig:Universal}. Dubbed `universal' for their transferability across different images (as opposed to models), these perturbations aim at satisfying the following constraint:
\begin{align}
\underset{{\bf I} \sim \mathcal I}{\mathrm{\text P}} \Big( \mathcal M({\bf I}) \neq \mathcal M({\bf I} + \boldsymbol \rho) \Big) \geq \delta ~~~\text{s.t.}~~||\boldsymbol\rho||_p \leq \eta,
\label{eq:universal}
\end{align}
where P(.) is the probability, $\mathcal I$ denotes the distribution of clean images and $\delta \in (0, 1]$ is a predefined scalar, deciding the acceptable fooling ratio for the perturbations. The resulting universal adversarial perturbations are shown to be effective with both $\ell_2$ and $\ell_{\infty}$ bounds over their respective norms. It can be observed from the experiments of \cite{moosavi2017universal} that perturbations bounded to around $4\%$ of the respective image norms are able to achieve a significant fooling ratio (of $\sim 80\%$) for popular ImageNet models, e.g.~ResNet~\cite{he2016deep}, Inception~\cite{szegedy2016inception}. However, a $4\%$ distortion in an image is often slightly perceivable to the human visual system, hence the authors termed the perturbations to be quasi-imperceptible.  

The universal adversarial perturbations are also able to transfer well across different models. In a sense, this property makes them `doubly universal', as suggested by the  authors~\cite{moosavi2017universal}. However, since the estimated  perturbations depend on parameters $\delta$ and $\eta$, $(\delta, \eta)$-universal perturbations is a more qualified term for these signals.  Moosavi-Dezfooli et al. compute these perturbations by building on the concept of Deepfool~\cite{moosavi2016deepfool}, where a single image is gradually pushed out of the decision boundary of its class. In the case of universal perturbations, the iterative algorithm sequentially pushes all the data points out of their respective class regions, 
while accumulating the (label changing) perturbations by back-projection them  onto an $\ell_p$-ball of radius $\eta$. It is shown in the original paper that computing universal perturbations with as little as 2000 training images can still achieve $\sim 50\%$ fooling ratio for ImageNet models.

\section{Recent attacks on classifiers}
\label{sec:RA}
Mainly building on the core concepts of the  first-generation attacks, there have been a multitude of more recent attacks on image classifiers. We cover those attacks in this section as per the structure illustrated in Fig.~\ref{fig:intro}(c).

\subsection{Advanced gradient based attacks}
There is still a variety of contributions that are intended to improve the core strategy of gradient ascend for adversarial attacks. Naturally, these methods can be seen as downstream fine-tuning of first generation attacks like FGSM or PGD. For instance, Dong et al.~\cite{dong2020robust} proposed to focus the gradient-based perturbation computed in an FGSM-like manner on the salient regions of images with the help of super-pixel guided attention. Such perturbations are claimed to be more robust against image processing based defenses.  Similarly, Guo et al.~\cite{guo2020backpropagating} focused on improving the transferability of gradient-based attacks by backpropagating the computed gradients \textit{linearly} through the model. Their gradient  backpropagation  mimics the scenario in which nonlinear activations are not encountered in the forward pass. Their modification is claimed to achieve better transferability of gradient-based attacks on large scale models.

Dong et al.~\cite{dong2020greedyfool} proposed a so-called GreedyFool algorithm that performs a sparse distortion in the input image based on gradients of its pixels. With improved sparsity, the perceptibly of their gradient-based perturbations becomes lower.  Sriramanan et al.~\cite{sriramanan2020guided} proposed a Guided Adversarial 
Margin Attack (GAMA) that introduces a relaxation term in the standard losses (e.g.~cross-entropy) of gradient-based attacks, e.g.~PGD. It is claimed that this modification allows the attack to find better gradient directions, thereby increasing its  efficacy. 
Similarly, Tohsiro et al.~\cite{tashiro2020diversity} devised a gradient-based strategy called Output Diversity Sampling (ODS) that is claimed to improve attacks in both white and black-box setups. Many adversarial attacks use random sampling of distributions, e.g.~for initializing optimization process or updating query (in black-box setup). The ODS is mainly directed to provide a better sampling scheme for such attacks.  


In \cite{rony2019decoupling}, decoupling of the direction and norm of $\ell_2$-norm bounded gradient-based perturbations was proposed to make the attack more lethal. The resulting attack is commonly referred to as Decoupled Direction and Norm (DDN) attack.  In \cite{yao2019trust}, Yao et al.~recommended to upgrade the first generation gradient-based attacks with Trust Regions~\cite{conn2000trust}. During optimization, trust regions around the current point in the loss landscape finds  descent/ascent directions  that reduce errors due to the local nature of decisions. 
It is shown that multiple  first-generation attacks can be improved for norm reduction and computational efficiency using trust regions. 
In \cite{Phan_2021_CVPR}, Phan et al.~argue to also consider the influence of image processing pipeline of cameras in attacks. They develop a gradient-based attack by differential approximation of this pipeline such that their perturbations are able to fool classifiers by images from one camera pipeline and not for another.  

We emphasize that although we categorize only a few methods under advanced gradient attacks, nearly all white box  (and transfer-based) attacks can be placed under this title, because those attacks inadvertently deal with model gradients rather directly. However, we introduce those attacks under subcategories more suited to their objective or threat models. Our intention to include a separate subsection for `advanced' gradient attacks is to emphasize on the fact that improving the core gradient ascend scheme for attacks is still an active direction in this domain. The gradient based attacks, which are inherently white box, are generally the easiest to compute. Hence, they are the hardest to defend against. This makes them a useful tool to analyze model robustness.   


\subsection{Black-box attacks}
From a pure adversarial perspective, black-box attacks form the most pragmatic category, because they assume no (or minimal) knowledge of the target model. Their practicality is making them highly popular in the recent literature. We review the recent black-box attacks along the directions of query-based and transfer-based attacks.

\subsubsection{Query-based attacks}
\begin{figure}[t!]
    \centering
    \includegraphics[width = 0.45\textwidth]{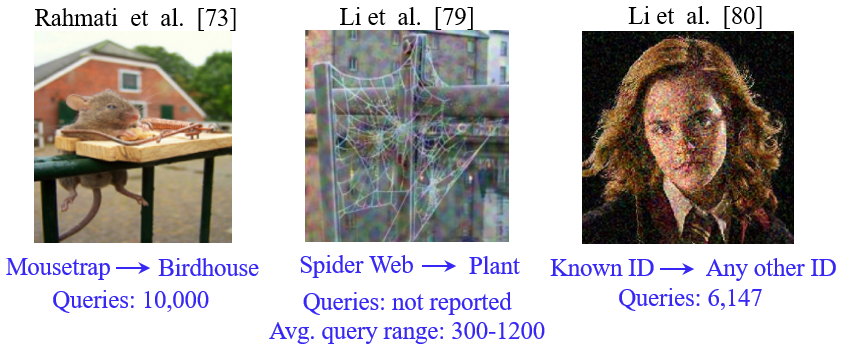}
    \caption{Representative examples of query-based adversarial examples (attacks selected randomly): Generally, a larger number of queries is required for smaller perturbation perceptibility. For \cite{li2020projection}, we provide average query range, as queries for the shown image are not reported. Perturbed images are  taken directly from \cite{rahmati2020geoda}, \cite{li2020projection}, \cite{li2020qeba}.}
    \label{fig:Querybased}
    \vspace{-4mm}
\end{figure}

These attacks query the target model and use their outputs to construct adversarial images. 
Generally, their objective is to achieve minimal distortion in adversarial samples while maintaining model fooling. Queries are normally utilized for  refining stronger  perturbations for imperceptibility, see Fig.~\ref{fig:Querybased}. Due to their practicality, decision/boundary-based attacks in this category are overwhelmingly popular as compared to their score-based counterpart. 

Recently, Rahmati et al.~\cite{rahmati2020geoda} introduced a  framework exploiting the decision boundary geometry  to launch a  black-box attack  with a small number of queries to the target model that returns only the top-1 label. The attack exploits the smaller `mean' curvature of the decision boundaries near the data point to estimate the normal vector, along which the data point can be efficiently nudged to the other side of the decision boundary by adding perturbations with small $\ell_p$-norm for $p\geq 1$. The authors also show that the computed perturbation converges to the minimal norm for $p = 2$ for curvature-bounded decision boundaries. Better performance in terms of the number of queries and perturbation norm are reported as compared to the Boundary attack~\cite{brendel2017decision}, HopSkipJump attack~\cite{chen2020hopskipjumpattack} and the qFool attack~\cite{liu2019geometry}.

The Customized Adversarial Boundary (CAB) attack~\cite{shi2020polishing} reduces the number of queries by customising  adversarial noise distribution with the queries in query-history, and initializing with perturbations already aimed for transferable attacks. Similarly, to improve query efficiency, a technique to extract generalizable prior using the earlier queries with meta learning is proposed in  \cite{du2019query}. Another effort to improve query efficiency includes Projection \& Probability-Driven Black-box Attack (PPBA)~\cite{li2020projection} that restricts the solution space of the problem with low-frequency constrained sensing matrix -  a concept inspired by compressive sensing theory. Li et al.~\cite{li2020qeba} proposed a Query Efficient Boundary-based Black-box Attack (QEBA), that iteratively adds perturbation to a source image to retain its original label, but alters the image to form a perceptibly clean target image of a different object.

In \cite{ru2019bayesopt}, a Bayesian optimisation based attack is proposed. One method to reduce the number of queries it to search for adversarial images in a lower dimensional latent space as compared to the original image space. In that case, estimating the correct dimensionality of the latent space becomes a problem of its own. Ru et al.~\cite{ru2019bayesopt} employ non-parametric Bayesian strategy to resolve that by exploiting Gaussian Processes~\cite{akhtar2018hyperspectral} based surrogate models to generate queries. Cheng et al.~\cite{cheng2019sign} also claimed a query efficient attack, altering the optimization objective of their previous work~\cite{cheng2018query} that performed a binary search to estimate the gradient of the target model using query results. Later, they improved the attack by drastically reducing the number of queries by focusing on estimating gradient signs instead of gradients \cite{cheng2019sign}. In another attempt to decrease the number of queries, Cheng et al.~\cite{cheng2019improving} also  introduced  a prior-guided random gradient-free method.



A TRansferable EMbedding based Black-box Attack (TREMBA) \cite{huang2019black} trains an encoder-decoder model to learn a low-dimensional embedding space, where an adversarial example is searched for a given target model in a query-based setup. This process is claimed to reduce the number of queries significantly due to reduction in the search space of queries. 
Another method looking at the problem from search space perspective performs the attack as a  progressive binary search using the gradient signs (instead of magnitude) \cite{al2019sign}. The attack shows fooling of MNIST models with as little as 12 queries.  
Ilyas et al.~\cite{ilyas2018prior} revisited the zeroth-order optimization (zoo) and proposed a query-based attack using bandit optimization that exploits prior information about the target model gradient. From zoo perspective, Zhao et al.~\cite{zhao2019design} also proposed to augment the optimization with an ADMM-based framework.

Query-based black-box attacks are attracting significant interest of the research community in the recent literature. There are multiple other recent works that deal with these kinds of attacks, e.g., \cite{andriushchenko2020square}, \cite{chen2020boosting}, \cite{xia2020improving}, \cite{cheng2018query}, \cite{Wang_2021_CVPRDelving}. 
Mostly, the current literature is dealing with  decision-based attacks~\cite{brunner2019guessing}, \cite{tolias2019targeted}, \cite{dolatabadi2020advflow}, \cite{Maho_2021_CVPR}, \cite{Li_2021_CVPRQAIR}. However, score-based attack schemes are also frequently encountered in the recent  literature~\cite{yang2020learning}, \cite{Ma_2021_CVPRSimulating}.


\vspace{1mm}
\subsubsection{Transfer-based attacks}
Among the black-box attacks, transfer-based attacks are even more popular than the query-based attacks. This is because transfer based attacks do not require to query the black-box model and hence avoid suspicion altogether. The core idea behind transfer-based attacks is to compute perturbation on local surrogate models such that the perturbations will also effectively fool the remote target model. Popularity of these attacks also owes to the fact that the insights from white-box setup can often be readily leveraged for these attacks. The main objective of the methods appearing in this direction then is to amplify the intrinsic transferability of perturbations, for which different strategies are adopted.  


Recently, Wu et al.~\cite{wu2020boosting} proposed to boost the transferability of perturbations by focusing them more on the salient image regions, where the regions are computed with Grad-CAM~\cite{selvaraju2017grad}. Improving perturbation transferability by manipulating the internal representation of the models is studied in \cite{li2020yet}. Similarly, Huang et al.~\cite{huang2019enhancing} fine-tuned  adversarial examples using representations of pre-specified layers of the source model to improve attack transferability. A concept of `Adversarial Example Game' is introduced in \cite{bose2020adversarial} that trains a generator for a transfer-based attack by training it against a discriminator for a hypothesis class of the target classifier. Since the underlying attack generation method does not assume details of the target (remote model), this setup is termed No-box attack in \cite{li2020practical}.

\begin{figure}
    \centering
    \includegraphics[width = 0.49\textwidth]{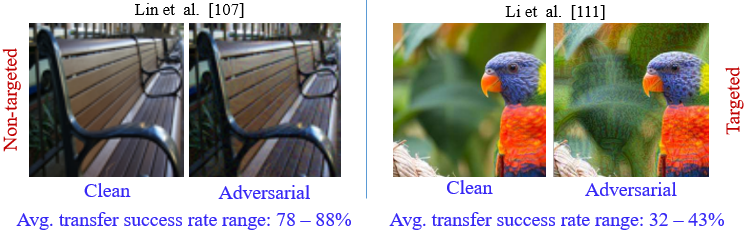}
    \caption{Typical examples of transfer-based perturbations (chosen randomly). Due to the harder objective of targeted transfer-based attacks, success rates are generally low (e.g.~$<50\%$) while perturbations are often perceptible. The reported average success rates across ImageNet models are taken from the original papers, which fall in the typical range of transfer-based attack success rates in the literature. Images are taken from \cite{lin2019nesterov} and \cite{li2020towards}.}
    \label{fig:TransferBased}
    \vspace{-3mm}
\end{figure}

From the perspective of enhancing perturbation transferability, Lin et al.~\cite{lin2019nesterov} exploit Nesterov gradient acceleration~\cite{nesterov1983method} with  iterative FGSM for computing more generalizable, and hence transferable perturbations. The authors also introduced a scale-invariant attack method that induces an ensemble of models from an original model using data transformations that preserve the original loss of the model. Adversarial examples computed with these models are shown to exhibit better transferability. Lu et al.~\cite{lu2020enhancing} demonstrated the possibility of fooling deep learning across  different computer vision tasks. Analysing image classification, object detection, semantic segmentation and content detection as the tasks, they showed transferability of adversarial examples across them with rather modest perturbations. This is mainly achieved by reducing the dispersion in the feature maps of the internal layers of the surrogate model with the help of a specialized loss.   
Inkawhich et al.~\cite{inkawhich2019feature} also claimed that feature space perturbations are particularly helpful in computing  adversarial examples that are more transferable across models.

There are also examples of targeted transferable attacks. For instance, 
Li et al.~\cite{li2020towards} proposed to make gradient-based targeted attacks more transferable by identifying two characteristics of white-box targeted attacks that restrict their transferability. First, reduction in gradient magnitude across iterations - leading to noise curing. Second, proximity of the adversarial examples to the true class region. The first issue  is handled in \cite{li2020towards} by allowing adaptive gradient magnitude in optimisation.  Whereas the second is mitigated by metric-learning based regularization.
Inkawhich et al.~\cite{inkawhich2020transferable} claimed state-of-the-art results for transferable targeted attacks on pristine ImageNet models. Instead of the classification layer, their method focuses on modeling layer-wise and class-wise feature distributions of a white box model and uses this information to alter the label of an adversarial image.

The direction of transferable attacks is also expanding in terms of the target tasks and underlying objective. For example, Wang et al.~\cite{wang2020transferable} demonstrated successful transferable attacks for the task of person re-identification. We also find an example of  improving transferability of  universal adversarial perturbations~\cite{li2019regional}. Moreover, training surrogate models in a data-free manner for transferable attacks was proposed in   \cite{zhou2020dast}. The core idea is to use a generator to construct synthetic images and label those with the target model (similar to query-based attacks), and train the substitute model with those images to better replicate the decision boundaries of the target model. 
Other recent examples focusing directly or indirectly on improving transferablity of perturbations include \cite{zou2020improving}, \cite{dong2019evading}, \cite{shi2019curls}, \cite{Wang_2021_CVPR}, \cite{Wu_2021_CVPRImproving}.





\subsection{Unrestricted adversarial attacks}
\begin{figure}[t!]
    \centering
    \includegraphics[width = 0.35\textwidth]{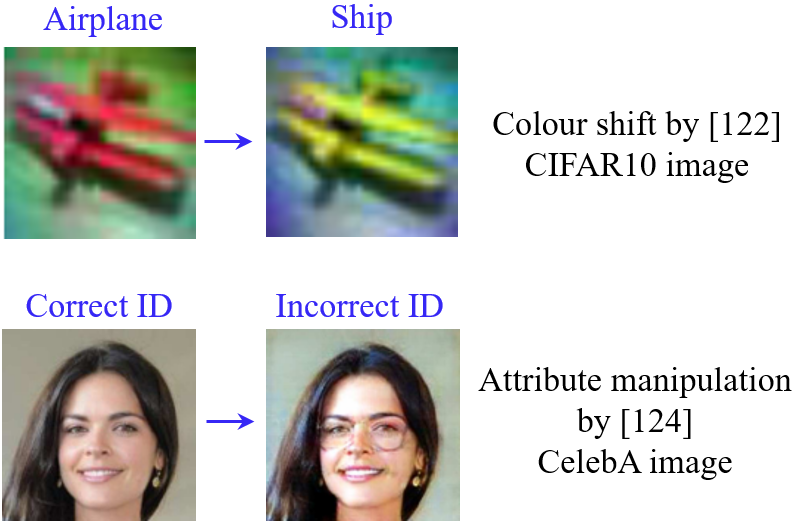}
    \caption{Examples of unrestricted attacks. Images taken from \cite{hosseini2018semantic}, \cite{joshi2019semantic}. }
    \label{fig:Unrestricted}
    \vspace{-3mm}
\end{figure}

Whereas the majority of mainstream  attacks induce perturbation imperceptibility in adversarial images by restricting the $\ell_p$-norm of perturbations, it is sometimes argued that the perturbation norm is not a good indicator of the perceptual difference between the two  images~\cite{sharif2018suitability}. The works related to achieving perturbation imperceptibility based on preserving semantics of the target image~\cite{hosseini2018semantic}, \cite{eykholt2018robust}, \cite{joshi2019semantic}, \cite{sharif2019general} and preserving the structural information~\cite{croce2019sparse}, \cite{wong2019wasserstein} are motivated by this argument, see Fig.~\ref{fig:Unrestricted}. 
In \cite{bhattad2019unrestricted}, unrestricted perturbations are introduced by manipulating the image color and texture to make them  adversarial. It is claimed that such unrestricted perturbations are generally robust to defenses like feature squeezing, JPEG compression and adversarial training.  On the other hand, compression and adversarial training are sometimes found effective against norm-bounded attacks like FGSM~\cite{goodfellow2014explaining}.
Shamsabadi et al.~\cite{shamsabadi2020colorfool} demonstrated that it is possible to selectively manipulate image colors imperceptibly by operating on the decorrelated  $a$, $b$ channels of the \textit{Lab} color space~\cite{ruderman1998statistics}. By changing image colors only to natural colors, and restricting manipulation to perceptually less sensitive regions in images, they computed transferable unrestricted adversarial examples that appear natural to humans. 

Zhao et al.~\cite{zhao2020towards} recently proposed to use the perceptual color distance CIEDE2000~\cite{luo2001development}  to control the imperceptibility of perturbations. The CIEDE2000 distance is known to align better with human perception. Zhao et al.~demonstrated that accounting for the perceptual color distance while perturbing images can  allow larger perturbations (having higher $\ell_p$-norm) to remain imperceptible. The authors extended the C\&W attack~\cite{carlini2017towards} to its variant that accounts for the perceptual color distance, Per-C\&W. Their results show that higher confidence on incorrect labels and better transferability of attacks is possible by considering the perceptual color distance, without sacrificing perturbation imperceptibility. The proposed Per-C\&W attack still computes a norm-bounded perturbation though, and the resulting image is not an unrestricted adversarial example. 
Another example of unrestricted perturbation attack is the semantic adversarial attack that manipulates image attributes with parametric conditional generative models~\cite{joshi2019semantic}, \cite{qiu2020semanticadv}. Incidentally, we can also categorise the emerging deepfakes~\cite{korshunov2018deepfakes}, \cite{Chen_2021_CVPRMagDR} as unrestricted attacks.
In a recent work, Hendrycks et al.~\cite{Hendrycks_2021_CVPR} also reported two sets of natural images for which ImageNet models have extremely low accuracies ($< 5\%$). Named ImageNet-A (for adversarial) and ImageNet-O (for out-of-distribution), these images are termed natural adversarial examples by the author.

\subsection{Backdoor attacks}
Whereas adversarial attacks manipulate images during test time, backdoor attacks embed a backdoor or Trojan in the model during training. The targeted model normally shows high accuracy for clean input, however, its output is easily manipulated by embedding an attacker-defined trigger in the input. Although this article does not directly deal with backdoor or Trojan attacks, we still include recent papers in the surveyed venues due to the proximity of this research direction to adversarial attacks and for the sake of comprehensiveness of our survey. For a more detailed review of backdoor attacks appearing in other venues, we refer to \cite{li2020backdoor}, \cite{liu2020survey}. 

Generally, backdoors are embedded in the victim model by including trigger patterns in the training data so that the model learns a false association of a label with the trigger pattern. An issue with such triggers in training data is that the trigger patterns are often conspicuous, leading to easy detection of triggers with visual inspection.  
Liu et al.~\cite{liu2020reflection} recently proposed to use reflection patterns as triggers. Casting the triggers as natural looking shadows makes them harder to detect. 
Often, triggers in the backdoor attacks are uniform across input images. However, Nguyen and Tran~\cite{nguyen2020input} proposed a generator-based backdoor attack that allows using different trigger patterns based on the context in the image. This makes detection of the trigger pattern even harder.
Xie et al.~\cite{xie2019dba} introduced a distributed backdoor attack on Fedrated  Learning~\cite{smith2017federated} in which the trigger is distributed among different parties providing the training data. This is in contrast to the centralized poisoning of data that appears in conventional supervised learning \cite{bhagoji2019analyzing}, \cite{bagdasaryan2020backdoor}.

A method claiming effective targeted poisoning was proposed in \cite{guo2020practical} for the practical setups where minimal assumptions can be made about the target network. That technique uses a pre-trained network to learn an attack model that can be directly used to generate images that would fool the victim model.
In another example of backdoor attacks, Rakin et al.~\cite{rakin2020tbt} generated a Trojan trigger to locate and flip the vulnerable bits of a DNN in DRAM to make it misbehave. It is noted in \cite{zhao2020clean} that static backdoor attack on images do not work well for videos. Hence,  a specialized backdoor attack for the task of video recognition was proposed. Similar to the concept of universal adversarial perturbations~\cite{moosavi2017universal}, their method uses a universal trigger to perform Trojan attack on video models.


The current literature is also witnessing multiple methods to secure deep learning models against the backdoor attacks. For instance, 
Kolouri et al.~\cite{kolouri2020universal} introduced  a `universal litmus patterns' (ULP) for detecting a backdoor in pre-trained models. The detection is done by binary classification of the response of logit layers of the model in question for multiple geometric ULPs. The geometric ULPs are pre-defined, which are computed by an optimization problem  inspired by universal adversarial perturbations~\cite{moosavi2017universal}. 
Along the line of defense against Trojan attacks, Wang et al.~\cite{wang2020attack} analyzed the possibility of detecting backdoors in the context of Federated Learning. They claimed that the detection is ``unlikely'' - assuming first-order oracles or polynomial time. Building on this theoretical insight they introduced a new family of backdoor attacks, called edge-case backdoors, which forces model fooling on the inputs living on the tail of the input distribution.  By doing so, they make the detection of their attack very hard.


\subsection{Model inversion}
Model inversion aims to reconstruct  training data or its markers from a trained model~\cite{fredrikson2014privacy}. These attacks raise serious privacy concerns. Although  model inversion problem is currently not as popular in the computer vision literature  as additive perturbations, the attack is highly relevant for visual models in practical adversarial setups.

Since the discovery of the model inversion phenomenon~\cite{fredrikson2014privacy}, there have been multiple attempts to formalize the underlying problem for systematic investigation of the issue. For example, \cite{wu2016methodology} uses the notion of influence from Boolean analysis to  characterized inversion of Boolean functions. Similarly, \cite{yeom2018privacy} formulates the risk faced by the model to reveal individuals in training data. It is shown that the risk increases with over-fitting. To an extent, the model inversion problem can be related to feature visualization \cite{nguyen2016synthesizing} or the recently introduced attack to explain~\cite{akhtar2021attack}. However, the overall adversarial objective of model inversion remains different from these frameworks which are more focused on model explanation.

Recently, Zhang et al.~\cite{zhang2020secret} proposed a generative model-inversion attack that trains a GAN to estimate the distributional prior of the target model's training data. Combining the prediction loss of the target model with the loss of the discriminator, the trained generator is shown to produce high quality training samples of the target model, especially for the  face models.  Interestingly, the authors showed that highly predictive models establish stronger correlation between learned features and the sample labels. This is exactly what can be leveraged to do better in launching an inversion attack. An implication of this fact is that more accurate models might be more vulnerable to inversion attacks.



\subsection{Adaptive attacks}
\label{sec:break}
It is now known that defenses against adversarial attacks can also be broken with counter-counter measures. For instance, in  \cite{carlini2017adversarial} and  \cite{athalye2018obfuscated}, we see multiple defenses broken with subsequent stronger attacks. This has prompted the research community to evaluate defenses against adaptive attacks~\cite{tramer2020adaptive}. Adaptive attacks are designed to specifically fool a defense mechanism. Although, research community is fast adapting the convention of evaluating defenses against adaptive attacks, Tramer et al.~\cite{tramer2020adaptive} showed that these evaluations are still far from providing   guaranteed robustness against such attacks. 
The authors demonstrate this by circumventing thirteen recent defenses published in the proceedings of ICML, ICLR and NeurIPS. These defenses include
\cite{xiao2019enhancing}, \cite{roth2019odds}, \cite{li2019generative}, \cite{bafna2018thwarting}, \cite{pang2019rethinking}, \cite{verma2019error}, \cite{pang2019improving}, \cite{sen2020empir}, \cite{yang2018characterizing},
\cite{pang2019mixup},
\cite{yang2019me}, \cite{yin2019adversarial} and \cite{yu2019new}. A key takeaway from \cite{tramer2020adaptive} is that adaptive attacks should be hand-designed to specific defenses to be more effective, instead of automated attack adaption. 

Although certifiable defenses are sometimes assumed robust to adaptive attacks, we also witness  instances in the literature for adaptive attacks against certified defenses (see Section~\ref{sec:certified}). For example, 
Ghiasi et al~\cite{ghiasi2020breaking} proposed a ``Shadow Attack'' to break certifiable defenses. Through generating the perturbation outside the certified $\ell_p$ bounds, their method produces a ``spoofed'' certificate, which results in visually imperceptible adversarial perturbations to break the defense. Whereas the underlying tools to develop adaptive attacks (i.e.~counter-counter measures) are generally similar to conventional adversarial attacks, it is normally the objective of circumventing a specific (type of) defense that characterizes adaptive attacks. In recent years, these attacks are studied mainly in the context of developing robust defense techniques.

\subsection{Miscellaneous attacks}
The above sections reviewed literature related to the attacks on classifiers along popular directions. There are also other multiple interesting attacks related to the classification problem that do not fall under the above-mentioned subcategories. We provide a summary of those attacks in this section. 

In \cite{zhou2020adversarial}, the authors devise an  adversarial ranking attack, where the attacker can raise or reduce the rank of the potential label for the image. The unique objective of this attack differentiates it from the conventional fooling attacks. The literature has also seen attempts to fool deep neural networks by exploiting their storage on Dynamic Random Access Memory (DRAM)~\cite{razavi2016flip}, \cite{hong2019terminal}, \cite{rakin2019bit}. These attacks are particularly interesting for deep learning in practice.
In another interesting work, Rezaei and Liu~\cite{rezaei2019target} demonstrated the possibility of adversarial manipulation of predictions for transfer learning, without the knowledge of the target domain. Similarly, Mor et al.~\cite{mor2019optimal} study optimal strategies against generative adversarial attacks. A Feature Disruptive Attack was proposed in \cite{ganeshan2019fda} that is targeted at disrupting the internal representation of models for the adversarial samples, instead of simply focusing on altering the prediction.

The literature has also seen attacks to disrupt classifiers for point clouds. For instance,  
Zhou et al.~\cite{zhou2020lg} proposed a label guided GAN-based method for targeted attack on 3D point clouds in real-time. The proposed Label-Guided employs a multi-branch adversarial network for input feature extraction and then embeds the target label information in the features with an encoder. 
Vulnerability of deep 3D point cloud models to isometry transformations has also been exposed \cite{zhao2020isometry}. Other recent examples of attacks on point clouds and 3D data (and their defenses) include~\cite{hamdi2020advpc},  \cite{zhou2019dup}, \cite{wicker2019robustness},
\cite{xiang2019generating},
\cite{zeng2019adversarial}.


We also witness adversarial examples for video classifiers. Due to the additional time dimension, attacks on images can often not be readily translated to video. Hence, specialized attacks for videos are devised.
Zhang et al.~\cite{zhang2020motion} used the movement patterns in the video frames to compute a noise prior that can help in gradient estimation for fooling video classifiers in the context of query-based attacks.  
A spatio-temporal attack is also introduced for embodied agents in \cite{liu2020spatiotemporal}. Liu et al.~\cite{liu2020adversarial} proposed an FGSM-like attack to fool skeleton-based human action recognition models. Their attack also accounts for multiple task-specific constraints in the optimization problem, e.g.~anthropomorphic plausibility of adversarial inputs. Another example of skeleton action recognition attack is \cite{Diao_2021_CVPRBASAR}. Wang et al.~\cite{Wang_2021_CVPRUnderstanding} have also provided an analysis of adversarial robustness of skeleton-based action recognition. 
We also see exploitation of task-specific constraints in developing attacks and defenses against such attacks. For example, Pony et al.~\cite{Pony_2021_CVPR} introduce flickering across the temporal dimension to fool video recognition systems.
Xiao et al.~\cite{xiao2019advit} proposed a defense against attacks on videos that 
detects adversarial inputs by analysing  temporal consistency property of the videos.

We also see examples in the literature that devise attacks for specific types of network architectures. For instance, attacks specifically devised for GCNs are studied in \cite{liu2020adversarial} and \cite{zugner2019adversarial}. Jin et al.~\cite{jin2020certified} also analyze certified robustness of GCNs under topological perturbations.  Similarly, an attack specialized to binarized neural networks can be found in \cite{khalil2018combinatorial}. Another example of attacks on quantized networks can  be found in \cite{lin2019defensive}. We also see other unique ways of rendering inputs adversarial for deep learning models. Alaifari et al.~\cite{alaifari2018adef} deformed image planes to construct adversarial examples. The techniques in~\cite{fan2020sparse} and  \cite{croce2019sparse} aim at perceptibility reduction of adversarial perturbations by directly focusing on $\ell_0$-norm reduction of perturbation vector.
In \cite{liu2019universal}, we witness  an unsupervised universal attack method to compute perturbations that exploit model uncertainty. The method uses a Monte Carlo scheme to activate more neurons to increase model uncertainty during perturbation computation with a stochastic gradient descent technique. It also exploits a textural bias prior. A steganography based universal adversarial perturbation method is proposed in \cite{din2020steganographic} that embeds a secret natural image in another image to render the latter adversarial. In another example of universal attacks, Rampini et al.~\cite{Rampini_2021_CVPR} extended the notion to deformable geometric shapes. They compute the attack in the spectral domain by perturbing eigenvalue sequence of the representation. Recovering shape from spectrum then leads to adversarial samples.

\section{Attacks beyond classification}
\label{sec:BeyondClassification}
In this section, we focus on the contributions that fool deep visual  models for tasks other than classification. Whereas the fundamental tools to generate perturbations for these tasks are the same as those used to fool classifiers, the unique objectives of these tasks results in more specialized attack algorithms.


\subsection{Object detection and tracking}
\label{sec:ODnT}
Object detection and tracking are longstanding computer vision problems. Their wide application in practical deep learning has led to numerous specialized techniques for these tasks. From the adversarial perspective, it has also resulted in specialized attacks.
Interestingly, many of those attacks  foray into the realm of physical world attacks (Section~\ref{sec:RealWorld}) due to the practical nature of these tasks.
For instance, Eykholt et al.~\cite{eykholt2018robust} and Zhong et al.~\cite{zhong2018perception} have analyzed adversarial stickers on stop signs in the context of autonomous driving to fool YOLO~\cite{redmon2017yolo9000} - a popular object detector. Jia et al.~\cite{jia2019fooling} have recently developed a  `tracker hijacking' technique to fool multiple object trackers with adversarial examples computed for object detectors in the perceptual pipeline of autonomous driving. We note that the original tracker used by \cite{jia2019fooling} follows tracking-by-detection paradigm~\cite{sun2019deep}. Adversarial training of detectors for their robustification is discussed in \cite{Chen_2021_CVPR}. The authors also proposed a class-aware adversarial training that uses universal perturbations to eventually compute class-weighted loss for improved robustness.

Yan et al.~\cite{yan2020cooling} developed an attacking technique to deceive single object trackers, in specific SiamRPN++~\cite{li2019siamrpn++}. Their method trains a generator model to construct adversarial frames under a `cooling-shrinking' loss. The loss is designed to cool down the hot target regions and forcing the bounding boxes to shrink during online tracking. 
A Fast-Attack-Network is also developed in \cite{liang2020efficient} to attack the trackers based on Siamese network. In \cite{wiyatno2019physical}, the authors  introduced an adversarial pattern that can be printed on a poster in the physical world. When a target moves in front of that poster, the tracker locks itself to the poster pattern instead of tracking the target.

\begin{figure*}[t!]
    \centering
    \includegraphics[width = 0.97\textwidth]{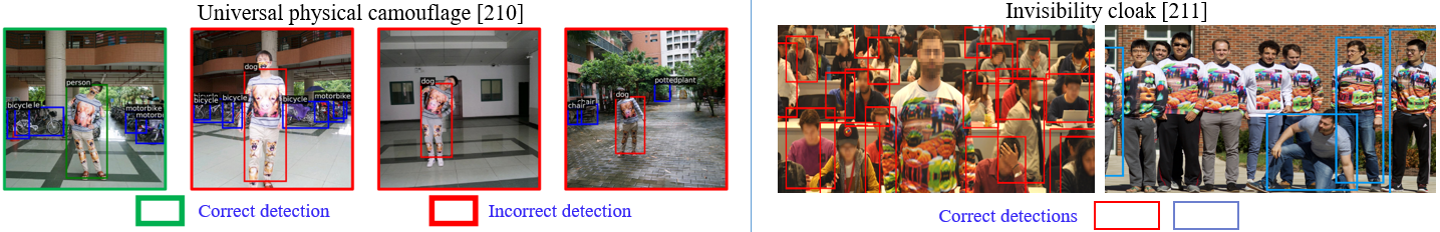}
    \caption{Representative attacks on detectors - randomly selected. \textbf{(Left)} Universal adversarial camouflage \cite{huang2020universal} incorrectly detects the target class at the cost of conspicuous patterns. (\textbf{Right}) Invisibility cloak~\cite{wu2020making} makes target object invisible with high probability. Images are taken from \cite{huang2020universal} and \cite{wu2020making}. }
    \label{fig:detector}
    \vspace{-3mm}
\end{figure*}

Huang et al.~\cite{huang2020universal} studied physical attacks on object detectors in-the-wild by developing a universal camouflage for object categories. The hard objective of the problem resulted  in conspicuous patterns for their attack though, see Fig.~\ref{fig:detector}. Robustness of object detectors are also explicitly studied in \cite{zhang2019towards}. Whereas it appears that object detectors are relatively hard to fool, \cite{zhang2019towards} shows that their robustness can also be improved with adversarial training. 
Another example of deceiving object detector in the real-world can be found in \cite{wu2020making}. Zolfi et al.~\cite{Zolfi_2021_CVPRPatch} develop a physical translucent patch that can be placed on camera lens to deceive detectors operating down the stream.
A one-shot adversarial attack is proposed in \cite{chen2020one} for single object tracking where inserting a patch in the first frame of the video results in losing the target in the subsequent frames.
A spatial-aware attack (SPARK) is proposed in \cite{guo2020spark} to fool online trackers. This method imposes an $L_{1,2}$ regularization constraint over perturbations while computing them incrementally based on previous frames. It is shown that their perturbations are able to fool multiple state-of-the-art trackers. An example of  black-box attack (decision-based) on trackers can be found in \cite{Jia_2021_CVPR} that focuses on IoU reduction by accounting for current and previous frames.

\subsection{Reinforcement learning}
Reinforcement Learning (RL) is a major research direction in AI. Although it is not a mainstream topic in computer vision research, adversarial attacks on RL systems are often inspired by  attacks on visual models. Hence, we find it imperative to briefly touch upon the advances made in adversarial attacks on RL in our literature survey.

Huang et al.~\cite{huang2017adversarial} were among the first to demonstrate that FGSM-like perturbations can also be used to degrade the performance of trained policies in RL. They considered adversaries that can  manipulate raw input of  policies. Their experiments prove success of adversaries even in black-box scenarios. Xiang et al.~\cite{xiang2018pca} developed a PCA-based model for predicting adversarial examples  in the context of Q-learning based path-finding. In another related work, Bai et al.~\cite{bai2018adversarial} also attacked the Deep Q Network (DQN)~\cite{mnih2013playing} for robotic path-finding  in a white-box setup. Similarly, Chen et al.~\cite{chen2018gradient} also explored adversarial attacks for the same problem, and devised a so-called Common Dominant Adversarial Examples Generation Method  for computing adversarial examples for a given map. In light of their threat models, we can categorize \cite{xiang2018pca}, \cite{bai2018adversarial} and  \cite{chen2018gradient} as white-box attacks within the RL context. We can also find early instances of black-box attacks for RL. For example, \cite{behzadan2017vulnerability} showed successful transferability of attacks across different DQN models. Additional early examples of black-box attacks on RL include \cite{lin2017tactics} and \cite{liu2017method}. For the interested readers, we refer to \cite{chen2019adversarial} for a more thorough review of the literature in adversarial attacks and defenses on RL up until 2019.

More recently, Gleave et al.~\cite{gleave2019adversarial} showed the existence of adversarial policies in zero-sum games between robots, especially in high dimensional environments. The victim of their policies are robust opponents trained with self-play. It is claimed that their adversarial policies defeat the victims reliably, while generating apparently random behavior. Rakhsha et al.~\cite{rakhsha2020policy} analysed a training time attack on RL where the adversary can poison the environment of an agent to enforce execution of a target policy in a stealthy manner. Zhang et al.~\cite{zhang2020adaptive} also developed  `adaptive' reward-poisoning attack that allows perturbation to reward at every step to cause learning of adversarial policies.
From the perspective of endowing robustness to deep reinforcement learning from adversarial observations in an agent's environment, Zhang et al.~\cite{zhang2020robust} showed that directly applying robustification methods, e.g.~adversarial training is insufficient. They proposed a State Adversarial Markov Decision Process (SA-MDP) method for regularizing the policies. It is claimed that this method is applicable to a large family of popular deep RL techniques, including DQN.

\subsection{Image Captioning/Description}
\begin{figure}[t]
    \centering
    \includegraphics[width = 0.4\textwidth]{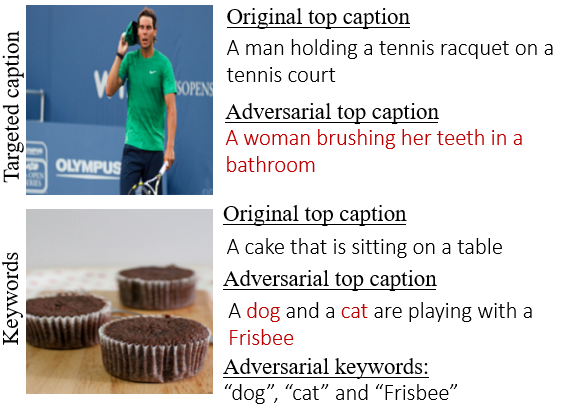}
    \caption{Example of `Show and Fool'~\cite{chen2017attacking} generating a targeted incorrect caption (top) and embedding adversarial keywords in a caption (bottom).}
    \label{fig:ShowAnbTell}
\end{figure}

Image and video captioning/description~\cite{aafaq2019video} is a multi-model task that normally involves a visual model (e.g.~CNN) to extract visual information from the input, followed by a language model (e.g.~RNN). Due to the temporal dependency in captions, attacking such a captioning/description framework is more challenging than  attacking a visual model alone. Nevertheless, we do find recent examples that successfully fool these frameworks. For instance, Xu et al.~\cite{xu2018fooling} fool an image captioning framework by treating the generated sentences as individual labels. Their focus is on fooling the language model (i.e.~RNN) while keeping the CNN embeddings of input image intact. In a related work, Chen et al.~proposed `Show and Fool' method~\cite{chen2017attacking} that fools the `Show and Tell' model~\cite{vinyals2015show}. Their technique can generate a pre-specified target caption for any image, or embed adversarial keywords in the caption, see Fig.~\ref{fig:ShowAnbTell}. Recently, Xu et al.~\cite{xu2019exact} also proposed a targeted partial caption attack that formulates the underlying task of generating adversarial partial captions as a structured output learning task with latent variables. The problem is solved under a generalized expectation maximization method  and structural SVMs with latent variables.
In \cite{xu2020machines}, an adversarial  optimization-based attack is developed for scene text recognition that employs sequential models. However, the method focuses on the related problem of text recognition, not directly on caption generation. Another remotely related work to captioning is FreeLB~\cite{zhu2019freelb} that promotes adversarial robustness in language models with adversarial training. Whereas attacks on captioning are currently not as popular as attacks on other mainstream computer vision tasks, following the trends of other tasks, we can anticipate a gradual rise in the popularity of captioning attack in the future.

\subsection{Face recognition}
\label{sec:FR}
Face recognition is also a long-standing problem in computer vision. Although the task is closely related to classification, due to specific data properties, it is often treated separately from classification. From adversarial perspective, treating face recognition separately is even more meaningful due to the serious implications of adversarial attacks on these systems, which are generally not relevant to general purpose visual classifiers. 


Although  deep learning era has witnessed  highly accurate face recognition models~\cite{zulqarnain2018learning}, \cite{deng2019arcface} these systems are also vulnerable to adversarial attacks. Goswami et al.~\cite{goswami2018unravelling} provided an analyses of face recognition systems' robustness against adversarial attacks. They ascertained the susceptibility of popular model OpenFace~\cite{amos2016openface} and VGG-Face~\cite{BMVC2015_41}. Dong et al.~\cite{dong2019efficient} also reported adversarial vulnerability of of face recognition in black-box setups, specifically to decision-based attacks.  They adapted a popular evolutionary strategy~\cite{hansen2001completely} to perform search over the perturbation in the black-box setup, where the search is guided by the local geometry of the searched directions for efficiency. Zhong et al.~\cite{zhong2020towards} used \textit{transferability} to fool face recognition systems in another black-box scenario. They devised a so-called Drop-out Face Attacking Network (DFANet) that focuses on matching the internal representation of an identify (image) with another identity to confuse the target model between the two. An FGSM-like method,  Penalized Fast Gradient Value Method was introduced in \cite{chatzikyriakidis2019adversarial} to demonstrate fooling of face recognition models. A friend-safe attack on face recognition systems was also introduced in \cite{kwon2019face}, which computes images that are adversarial for `enemy' models, but benign for `friend' models.

\begin{figure}[t]
    \centering
    \includegraphics[width = 0.4\textwidth]{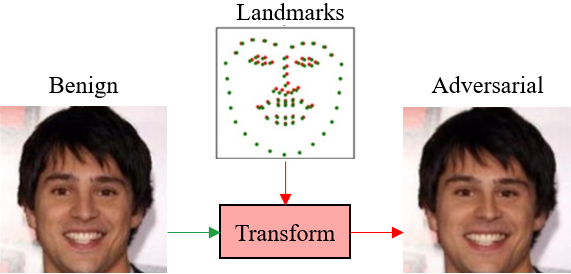}
    \caption{An example of fooling face recognition system by landmark manipulation~\cite{dabouei2019fast}.}
    \label{fig:FR1}
\end{figure}

The above methods mainly compute additive perturbations without explicitly accounting for geometric information of faces.  In contrast, Dabouei et al.~\cite{dabouei2019fast} devised a facial landmark manipulation method to mislead recognition systems. Their technique computes adversarial landmarks to perform spatial distortions in images that result in incorrect recognition, see Fig.~\ref{fig:FR1}. Adversarial patches for faces are also studied for their transferability in \cite{Xiao_2021_CVPR}. In another related example, Yang et al.~\cite{yang2020attacks} proposed an Attentional Adversarial Attack Generative Network ($A^3GN$) for targeted fooling of face recognition models. It is claimed that their network is able to exploit geometric and context information of the target with the help of a conditional VAE and attention modules to achieve this feat. 
Another example of using GAN for deceiving face recognition systems is AdvFaces~\cite{deb2019advfaces} that manipulate geometric features of the face in the image.
Similarly, Li et al.~\cite{Li_2021_CVPR} generate a fake face image by matching the latent representation of the image with its adversarial counterpart that can fool fake image detectors.  Along the line of utilizing GANs for manipulating faces in images and videos, an interesting research direction of DeepFakes is emerging. Interested readers are referred to \cite{tolosana2020deepfakes} for a recent survey of that direction.

\begin{figure}[t]
    \centering
    \includegraphics[width = 0.45\textwidth]{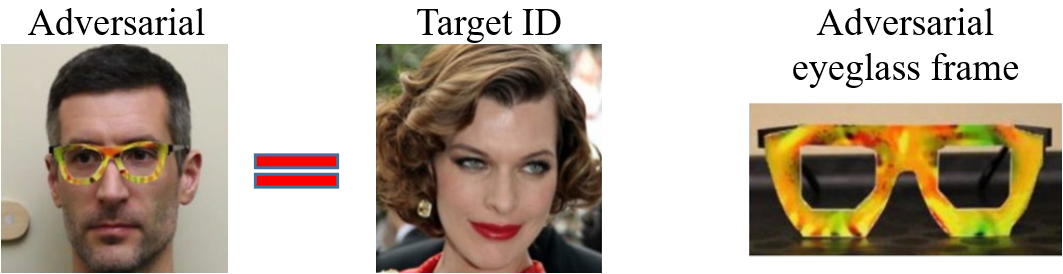}
    \caption{An example of fooling a face recognition model by wearing an adversarial eyeglass fame~\cite{sharif2016accessorize}. }
    \label{fig:FR2}
\end{figure}

Due to the practical nature of face recognition task, the literature has also witnessed fooling attempts through manipulating faces in the physical world. For instance, Sharif et al.~\cite{sharif2016accessorize} demonstrated the possibility of physically realizable attacks to impersonate an identity or evade the face recognition system. They devised an eyeglass frame for fooling the target network, see Fig.~\ref{fig:FR2}. Their technique was further improved in \cite{sharif2019general} for attack robustness. On a similar line, Zhou et al.~\cite{zhou2018invisible} developed a cap that illuminates  face of the person wearing it to fool the recognition system. They compute the adversarial illumination pattern on a image of the identity and use the cap to project that pattern on the face in physical world while presenting the face to the vision system. A related concept of `adversarial light projection' is studied in \cite{nguyen2020adversarial} that projects a rather conspicuous pattern on faces to evade FaceNet model~\cite{schroff2015facenet} in white-box settings. 
Other examples of physical world attack on face recognition systems include AdvHat~\cite{komkov2019advhat} and adversarial patches for faces~\cite{pautov2019adversarial}. Face presentation attacks are also studied in \cite{shao2019multi}. We refer to our first survey~\cite{akhtar2018threat} for the more classic attacks on face recognition.

\subsection{Miscellaneous attacks}
Even for the tasks beyond classification, multiple other attacks exist that do not fall under the categories described above.
For example, Nakka et al.~\cite{nakka2020indirect} devised an attack to demonstrate the vulnerability of semantic segmentation networks against holistic perturbations and localized ones.
Similarly, for the problem of segmentation, a data membership attack is devised in \cite{he2020segmentations}. 
Choi et al.~\cite{choi2019evaluating} also observed that deep
learning models for super-resolution are also vulnerable to adversarial attacks. This is demonstrated by introducing unnoticeable distortions in the low-resolution images, which adversely affect the super resolution results.
Mehra et al.~\cite{Mehra_2021_CVPRHow} proposed a poisoning attacks for reducing the average certified radius of a given class for certified defenses.

Deep neural networks are often successfully applied to predict depth in monocular scenes. Recently, \cite{wong2020targeted} showed that adversarial attacks can be used to manipulate the predicted distance from the camera. 
The method in \cite{wong2020targeted} can match the predicted distance to a different target scene or directly fabricate the depth of specific instances in the scene. 
Targeted attacks on hashing based retrieval are proposed in \cite{bai2020targeted}, \cite{Wang_2021_CVPR}, whereas a  universal perturbation for image retrieval systems is computed in \cite{li2019universal}.
An example of adversarial attack on Graph Matching can be found in \cite{zhang2020adversarial}.
There has also been enhancements and variants of patch attacks for multiple vision tasks. For example, Yang et al.~\cite{yang2020patchattack} improved the patch attack in a blackbox setup by reducing the required number of queries with reinforcement learning. 
Similarly, a universal patch is proposed for face recognition in \cite{yang2019design} and the patch attack is extended to optical flow in \cite{ranjan2019attacking}. It is shown that a patch as small as $1\%$ of the image size can disrupt optical flow networks.
\section{Physical world attacks}
\label{sec:RealWorld}
We already reviewed some of the literature performing physical world attacks in \S~\ref{sec:ODnT} for the tasks of object detection and tracking, and for  face recognition in \S~\ref{sec:FR}. Below, we further  expand  on the literature in this direction by focusing on the practical physical world application of autonomous driving and general purpose object detection and classification attacks.

In the context of autonomous driving, Tu et al.~\cite{tu2020physically} proposed a technique to compute physically realizable adversarial examples using LiDAR data to fool object detectors in simulated autonomous vehicle scenarios. It is claimed that placing an adversarial object (with underlying adversarial mesh computed from their technique) on the rooftop of a target vehicle can make the vehicle undetectable. The mesh surface of the computed object generally remains unnatural though. 
In another study, Cao et al.~\cite{cao2019adversarial} showed that CNN-based object detectors can be fooled in vehicle detectioin scenarios. An adversarial pattern computed by their technique serves as a camouflage to evade detectors in their work.
The notion of camouflage is also explored in the physical world settings in \cite{Wang_2021_CVPRDual}, \cite{zhang2018camou}.

Kong et al.~\cite{kong2020physgan} used a GAN-based setup to generate norm-bounded adversarial images, which when printed, demonstrate resilience to changes in the physical world conditions, e.g.~lighting condition, viewing angle.
Their method, PhysGAN is specifically designed to fool steering models of autonomous vehicles, under a regression-based formulation of the angle prediction problem. PhysGAN computes perturbations for a stream of visual features of driving video while ignoring the scene background. It is claimed that this strategy allows effective perturbations for dynamic scene conditions, nullifying the need of static scene assumption appearing in earlier literature~\cite{eykholt2018robust}.
In \cite{ho2019catastrophic}, it is shown that with camera shake and pose variation while imaging physical world objects, one can acquire images that can easily fool deep learning models. Here, the imperceptibility of the perturbation comes in the form of semantic-imperceptibility i.e.~contextually, the pose or shake appears natural. 

\begin{figure}[t!]
    \centering
    \includegraphics[width = 3in]{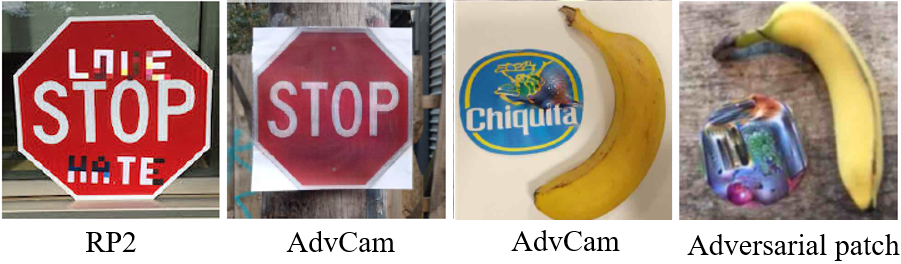}
    \caption{Representative examples of successful physical world attacks to fool recognition systems with AdvCam~\cite{duan2020adversarial}, RP2~\cite{eykholt2018robust} and adversarial patch~\cite{brown2017adversarial}.}
    \label{fig:AdvCam}
    \vspace{-3mm}
\end{figure}

The use of `adversarial patch'~\cite{brown2017adversarial} is another effective method to launch a physical world attack. An adversarial patch is normally a clearly visible, but well localized pattern - i.e.~a patch that can be placed beside an object to cause model fooling, see Fig.~\ref{fig:AdvCam}. 
More recently, Duan et al.~\cite{duan2020adversarial} proposed a neural style transfer \cite{jing2019neural} based technique to compute unrestricted perturbations that can take effect as a physical world attack to camouflage a target object. Their proposed AdvCam is able to compute patterns that are claimed to be more stealthy than earlier related techniques (e.g.~adversarial patch~\cite{brown2017adversarial}, RP2~\cite{eykholt2018robust}) in that the adversarial pattern appears more natural to humans (Fig.~\ref{fig:AdvCam}). To compute the physical world pattern, their technique captures image of the scene with a given camera and estimates the perturbation digitally in a restricted knowledge white box setup. The substitute model used to compute the adversarial pattern has the same architecture as the target model. The pattern is then placed in the physical world and the same camera is again used to capture the adversarial example. The technique requires manual specification of the region to place the adversarial pattern and the target style. In another example related to adversarial patch, Liu et al.~\cite{liu2020bias} constructed a universal patch and used it to deceive automatic checkout models.

\begin{figure}[t!]
    \centering
    \includegraphics[width = 0.47\textwidth]{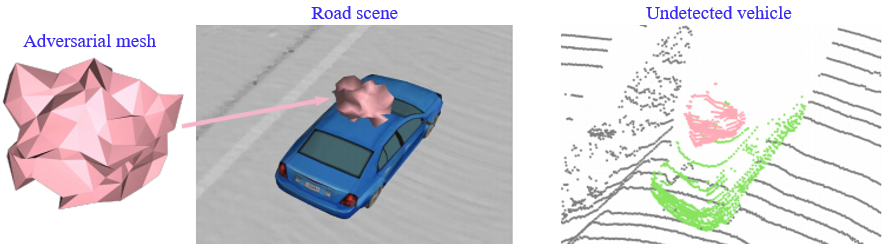}
    \caption{Representative example of fooling object detector with LiDAR data~\cite{tu2020physically}. The adversarial mesh placed on top of vehicle makes it undetectable for visual detector. Image taken from \cite{tu2020physically}.}
    \label{fig:advMesh}
    \vspace{-3mm}
\end{figure}

As also noted in \S~\ref{sec:ODnT}, deceiving object detectors is a particularly interesting problem for the physical world attacks. We are  already witnessing interesting methods in this direction. Recently, Xu et al.~\cite{xu2020adversarial} proposed a technique  to fabricate adversarial T-shirt for evading detectors. Another example of fooling object detectors on similar lines is \cite{wu2020making}. The authors compute a so-mentioned   `invisibility cloak' that contains the patterns causing misdetections for state-of-the-art detectors, see Fig.~\ref{fig:detector}. Considering the implications of this direction of research, a dataset for adversarial attacks on object detectors in the physical world is also introduced in~\cite{braunegg2020apricot}. Whereas currently attacks on object detectors are not as popular as attacks on classifiers, we can anticipate much larger interest of the research community for this problem due to many interesting, and sometimes security-critical applications.

There are also works in the literature that  fool classifiers by distorted illumination in the scene. For instance, Sayles et al.~\cite{Sayles_2021_CVPR} distort leverage radiometric rolling shutter effect for distortions that cause misclassification.    
Similarly, Duan et al.~\cite{Duan_2021_CVPRLaser} proposed  an adversarial laser beam attack, which computes adversarial parameters for a laser that can be used to distort illumination such that the captured image is adversarial. 
We have already seen multiple examples for fooling face recognition systems with adversarial  illumination patterns in Section~\ref{sec:FR}.

\section{Beyond adversarial objective}
\label{sec:BeyondAdvObj}
Although the primary objective of adversarial attacks in the literature is to fool deep learning models, there are also instances where adversarial perturbations are exploited under more constructive objectives of improving model performance, interpreting it, or estimating the performance. Note that, for the former, we are not alluding to the works leveraging adversarial training to robustify models -  explained shortly. 

\subsection{Improving model performance}
\label{sec:ImpPerfo}
Xie et al.~\cite{xie2020adversarial} recently demonstrated that adversarial examples can actually help in performance gain in fully supervised setups for large-scale models, e.g.~ImageNet~\cite{deng2009imagenet}.
To demonstrate that, the authors propose Adversarial Propagation (AdvProp) technique that is applied to EfficientNet-B7~\cite{tan2019efficientnet} to achieve performance gains of 0.7\%, 6.5\%, 7.0\% and 4.8\% for ImageNet, ImageNet-C, ImageNet-A and Stylized ImageNet datasets. Moreover, after enhancing the network to EfficientNet-B8, their method sets the new state-of-the-art of 85.5\% on ImageNet top-1 accuracy without extra training data.
The key insight used by AdvProp is that the underlying distribution of adversarial images is different from natural images. This calls for disentangling the normalisation statistics for the networks in the Batch Normalization (BN) layers. Hence, the authors proposed an auxiliary BN layer that is explicitly used for adversarial examples during training, and dropped during testing. During training, the loss is computed by propagating the clean and adversarial images separately through their respective BN layers.

The AdvProp  is unique in that successfully aims at performance gain for large-scale models on clean images with adversarial examples. This is different from adversarial training, which generally results in sacrificing model accuracy on the clean images to gain robustness to adversarial examples~\cite{akhtar2018threat}, \cite{raghunathan2019adversarial}. There are also other instances that report   performance gain on clean data by accounting for adversarial image in training. For instance, \cite{szegedy2013intriguing}, \cite{li2019inductive} report improved model accuracy for a small dataset (MNIST) under a fully supervised setup. Similarly, \cite{miyato2018virtual} and \cite{qiao2018deep} improve model performance with adversarial examples for large models in a semi-supervised setup.
Ho and Nuno et al.~\cite{ho2020contrastive} also found use of adversarial example in Contrastive Learning for self-supervised learning. They used adversarial examples to augment data for pretext learning of embeddings.

It is also claimed by Salman et al.~\cite{salman2020adversarially} that adversarially trained models, while less accurate than the standard models, often perform better for transfer learning. 
In another study, Gan et al.~\cite{gan2020large} propose VILLA, a representation learning approach based on large-scale adversarial training on vision-and-language data. They perform a task-agnostic adversarial training followed by a task-specific adversarial fine-tuning in the embedding space. This method is claimed to achieve  state-of-the-art performance on a variety of tasks, including Visual Question Answering, Visual Commonsense Reasoning, and Image-Text Retrieval. Using the anti-adversarial directions for weakly supervised models, Lee et al.~\cite{Lee_2021_CVPRAnti} claimed improvement in the semantic segmentation performance.

\subsection{The link between attacks and model interpretation}
\begin{figure*}[t]
    \centering
    \includegraphics[width = 0.95\textwidth]{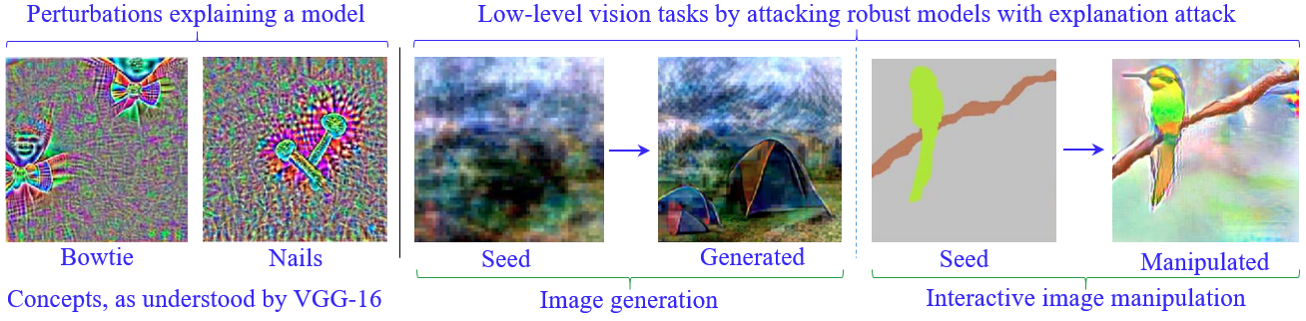}
    \caption{Jalwana et al.~\cite{jalwana2020attack} proposed an `attack to explain' which uses perturbations to visualize human-defined semantic concepts as understood by a model. They also utilized their attack to perform low-level vision tasks by attacking robust classifiers. (\textbf{Left}) Perturbations computed with attack to explain VGG-16. The signals visualize understanding of VGG-16 for human-defined concepts of Bowtie and Nails. (\textbf{Right}) The attack is used to generate an image of `mountain tent' with a random seed image. For interactive image manipulation, the seed image is refined into a bird with the attack.}
    \label{fig:Explanation}
    \vspace{-3mm}
\end{figure*}

Jalwana et al.~\cite{jalwana2020attack} developed a technique to visually reveal the understanding of human-defined semantic concepts by  deep learning perceptual models, see Fig.~\ref{fig:Explanation}. By expanding the domain of the adversarial perturbation and iteratively refining it, the authors demonstrate the presence of human-understandable patterns in the perturbations. A more clear relation between the adversarial and explanation character of their perturbations is later established in \cite{akhtar2021attack}. The authors also utilize their `attack to explain' to perform low-level vision tasks by attacking robust classifiers. This concept builds on \cite{santurkar2019image}.  In a related approach, Augustine et al.~\cite{augustin2020adversarial} associate model explainability to its adversarial robustness, demonstrating generative properties of their adversarially robust model similar to \cite{santurkar2019image}. Elliott et al.~\cite{Elliott_2021_CVPRExplaining} also attempts to bridge the gap between adversarial perturbations and counter-factual explanation of deep models. They localized their perturbations to salient regions of inputs to demonstrate that perceptually regularized counterfactuals provide useful model explanation. 

There is also a line of research that considers interpretability of the induced perturbation patterns themselves. For instance, Xu et al.~\cite{xu2018structured} applied  group sparsity over the perturbation vector and showed that the resulting perturbations are more interpretable. Nevertheless, this method does not offer interpretability of the model itself like \cite{jalwana2020attack}, \cite{akhtar2021attack}.

\subsection{Other applications}
Among other interesting related applications, Elsayed et al.~\cite{elsayed2018adversarial} showed that adversarial perturbations can be used to reprogram a target model. For instance, with embedded perturbations, they successfully converted a classifier into a box-counting machine. Finally,  Sakaguchi et al.~\cite{sakaguchi2020winogrande} recently proposed an algorithm, called AFLITE, to adversarially reduce task- or dataset-specific biases in head distribution, while preserving complexity of the tail. This bias reduction mitigates overestimation of model performance, which is evident by their performance on out-of-distribution and adversarial examples. In a followup work, Le et al.~\cite{le2020adversarial} theoretically studied AFLITE and provided extensive evidence that AFLITE reduces measurable dataset biases. They showed that the models trained on the filtered dataset generalize better to out-of-distribution data.
\section{On the existence of adversarial examples}
\label{sec:On}
The existence of adversarial examples for otherwise highly accurate deep visual models has confounded the research community since the discovery of this phenomenon. The literature has witnessed numerous hypotheses to explain the adversarial vulnerability of deep learning. However, many of those fail to generalise, and the remaining often  conflict with each other. It can be argued that there is still no consensus on the reasons of the existence of adversarial examples. Whereas it was common among the earlier contributions to also hypothesize about  generic causes of the adversarial susceptibility of neural networks, the recent attack methods are more concerned with achieving higher fooling rates and better transferability etc. Nevertheless, the contributions that analyze the causes of adversarial vulnerability as the core topic, are still attractive because the wider impression is that this phenomenon is still not fully understood.
Below, we review contributions and major hypotheses in this direction along the lines of input-specific perturbations, input-agnostic perturbations and other prevailing topics. 

\subsection{On input-specific perturbations}
One of the first popular hypotheses on the existence of adversarial examples for modern deep network was the linearity hypothesis~\cite{goodfellow2014explaining} - see the FGSM attack in \S~\ref{sec:firstGen} for details. However, it was later shown by Tanay and Griffin~\cite{tanay2016boundary} that this hypothesis does not generalize as there are classes that do not suffer from adversarial examples for linear classifiers.  Nevertheless, Kortov and Hopfiled~\cite{krotov2018dense} later provided another evidence based on Dense Associative Memory~\cite{krotov2016dense} that supports the role of linearity in neural model susceptibility to adversarial examples. Similarly, linearity is also blamed for adversarial vulnerability in \cite{taghanaki2019kernelized}.
In \cite{cubuk2017intriguing}, `inherent prediction uncertainty' of neural networks is blamed for their adversarial vulnerability. The claim is corroborated by computing a functional form of the prediction uncertainty which remains independent of the architecture and training of the model. It is also argued that clean image accuracy of models correlates with their adversarial robustness, which resonates with the findings of \cite{rozsa2016accuracy} and other earlier observations \cite{goodfellow2014explaining}, \cite{madry2017towards}. Evolutionary stalling~\cite{rozsa2016towards} is another interesting hypothesis, according to which, the inability of training samples to contribute beyond a certain capacity leaves their representation very close to the model decision boundaries. This allows adversaries to easily nudge those and similar representations out of the correct classification regions.

Exploring the space of adversarial examples, Tabacof and Valle~\cite{tabacof2016exploring} showed that adversarial examples reside in large regions in the pixel space of images. Their findings suggest that weak shallow networks are as susceptible to adversarial examples as the complex deep networks. On a similar note, Tramer et al.~\cite{tramer2017space} claimed that adversarial examples span a contiguous high dimensional space. The high dimensionality of this space and subspaces of different classifiers results in their intersections which causes transferability of the attacks across different models.  More recently, \cite{li2020defense} claimed that model's gradient leakage along the perpendicular to a tangent space of training data manifold contributes to adversarial vulnerability of the models. 

In another work, Jacobsen et al.~\cite{jacobsen2018excessive} claimed that deep neural networks are highly invariant to a variety of task-relevant changes in the input that causes vast input space regions to be vulnerable
to adversarial perturbations. This is in addition to the high sensitivity of the models to the task-irrelevant changes to the input. Along  the lines of analysing the existence of adversarial examples from the robustness perspective, Reddy et al.~\cite{reddy2020biologically} studied the biological visual system of primates. They showed that non-uniform sampling done by  primate retina and the existence of multiple receptive fields (having a range of field sizes) improves the robustness of neural networks to adversarial perturbations.

Pal and Vidal~\cite{pal2020game} proposed a game-theoretic framework for analysing attacks and defenses that exist in equilibrium. They proved that under a locally linear decision boundary model, FGSM and the randomized smoothing~\cite{cohen2019certified} exhibit a Nash Equilibrium~\cite{nash1950equilibrium}. Daniely et al.~\cite{daniely2020most}  provided a theoretical analysis that studies the vulnerability of ReLU networks against adversarial perturbations, concluding that most ReLU networks suffer from $\ell_2$ perturbations.   
We also find a similar but broader claim that adversarial examples are inevitable for certain types of problems in~\cite{shafahi2018adversarial}.

Similar to the linearity hypothesis of Goodfellow et al.~\cite{goodfellow2014explaining}, another  popular concept related to the existence of adversarial examples is `manifold assumption'~\cite{tanay2016boundary}, which argues that adversarial examples tend to leave the clean data manifold. Nevertheless, there is also evidence of on-manifold adversarial examples~\cite{gilmer2018adversarial}, \cite{song2018generative}, \cite{brown2018unrestricted}.
There is also a debate in the literature of associating robustness of neural models to their generalization~\cite{stutz2019disentangling}. For instance, Trsipras et al.~\cite{tsipras2018robustness} provide systematic evidence of clash between adversarial robustness and generalization of a  model. This is also partially supported by the empirical study in \cite{su2018robustness}. However, we also find works in the literature that suggest the opposite \cite{gilmer2018adversarial}, \cite{rozsa2016accuracy}, i.e.~improved generalization results in better robustness.

\subsection{On Input-agnostic perturbations}
Analysing the existence of universal perturbation, Moosavi-Dezfooli et al.~\cite{moosavi2017universal} claimed that these signals leverage the geometric correlations between the decision boundaries of classifiers. The authors theoretically demonstrate the existence of common directions for multiple data points along which a classifier's decision boundaries can be highly curved~\cite{moosavi2017analysis}, \cite{moosavi2018robustness}. It is argued that such directions allow the universal perturbations to effectively fool the classifier across multiple samples. On a similar note, Jetley et al.~\cite{jetley2018friends} demonstrated that the directions (in image space) used by neural networks to achieve higher performance are the same that make them vulnerable to adversarial attacks. Thus, the high accuracy and adversarial vulnerability of neural networks are related phenomena, which allow the existence of universal perturbations. 

Analysing the Pearson correlation between the coefficients of logit vectors of a classifier for clean and adversarial images, Zhang et al.~\cite{zhang2020understanding} showed that for universal perturbations, adversarial examples are strongly correlated with the perturbations. On the other hand, a low correlation is observed between the adversarial and clean images. The leads to the conclusion that universal perturbations hold more dominant features as compared to clean images despite their low power and visual (quasi-)imperceptibility. The authors also leverage this insight to introduce a method to compute universal perturbations using random clean images.

\subsection{Adversarial examples as features \& other sources}
Inline with the findings of Jetley et al.~\cite{jetley2018friends} (discussed above), Ilyas et al.~\cite{ilyas2019adversarial} claimed adversarial examples to be essential data features for neural networks, as opposed to unwarranted bugs. They demonstrated that the existence of adversarial examples can be attributed to non-robust features that are pervasive in datasets and are an effective source of achieving higher accuracy for the neural perception models.  The authors also demonstrate the possibility of disentangling robust and non-robust features and showed that robust features align more to human perception than their non-robust counterparts. This insight is exploited by Santurkar et al.~\cite{santurkar2019image} to use adversarially robust models to perform visually appealing image synthesis. Tsipras et al.~\cite{tsipras2018robustness} also noted the tension between adversarial robustness and classifier accuracy under the idea that adversarial examples are non-robust features used by models to achieve better performance.  

Bubeck et al.~\cite{bubeck2019adversarial} argued that adversarial vulnerability of classifiers in high dimension is ``likely not due to information theoretic limitations, but rather it could be due to computational constraints". They provided evidence to support the hypothesis that ``identifying a robust classifier from limited training data is information theoretically possible but computationally impossible". Interestingly, their evidence  weakens the notion that identifying a robust classifier requires huge amount of training data.   
In a study more focused on ReseNet inspired architectures, Wu et al.~\cite{wu2020skip} identified adversarial vulnerabilities of skip-connections. The authors claimed that the use of skip connection results in more transferable adversarial examples for models. They introduced a Skip Gradient Method (SGM) that relies on the gradient flow from skip connections to  compute more transferable examples for the models that employ skip connections.    
\section{Defense against adversarial attacks}
\label{sec:defense}
Akhtar and Mian \cite{akhtar2018threat} organized defenses against adversarial attacks into three broad categories. They incorporated defenses resulting from (1) modifying the target models for robustification, (2) modifying input for perturbation removal and (3) adding external modules (mainly detectors) to the model. Since 2018, the research direction of adversarial defenses has evolved mainly along the same three lines.
Hence, we first review the recent literature along the same directions. However, there is a subclass of defenses that is gaining rapid popularity in the recent literature, known as `certified defenses'. Although most of the works in this subclass follow (1), we review these methods separately in \S~\ref{sec:certified} due to their unique common objective of providing certificates/guarantees for the developed defenses. In \S~\ref{sec:defOther}, we also provide a bird's-eye view of other recent defense techniques that either combine more than one strategies noted above, or develop specialized defenses, e.g.~for specific tasks or network types. 


\subsection{Model alteration for defense}
The most common framework that modifies the (potentially) targeted model itself for robustification against adversarial attacks is `adversarial training'. Hence, we review techniques focusing on this framework separately, before discussing other recent methods in the category.

\subsubsection{Adversarial training}

The adversarial training framework is considered among the strongest principled defenses against adversarial attacks. It exposes the model to adversarial examples during training to obtain some level of immunity against them. Adversarial training was originally employed  in~\cite{szegedy2013intriguing}, \cite{goodfellow2014explaining}. However, Madry et al.~\cite{madry2017towards} are the first to theoretically study and justify it through the lens of robust optimization for deep learning.
Since \cite{madry2017towards}, adversarial training has attracted significant interest of the research community. 
This also resulted in multiple  contributions highlighting weaknesses of this framework. For instance, Zhang et al.~\cite{zhang2019limitations} showed that adversarialy trained models are still vulnerable to `blind-spot' attacks. Arguments  against the robustness induced by adversarial training can also be found in \cite{schott2018towards}.
It is also claimed that adversarial training is sensitive to the training data distribution in \cite{ding2019sensitivity}. Moreover, poor generalization of adversarial training is also often highlighted in the literature~\cite{song2018improving}, \cite{geirhos2018imagenet}, \cite{zhang2019interpreting}, \cite{Gong_2021_CVPRMaxup}. 

Despite is shortcomings, adversarial training is still favored by the research community due to its principled nature. Over the last few years, multiple variants and enhancements of adversarial training have surfaced. For example, a Missclassification Aware adveRsarial Training (MART) is proposed in \cite{wang2019improving} to incorporate distinctive influence of clean misclassified examples in the training process. Gowal et al.~\cite{gowal2020achieving} improved adversarial training by varying the Style-GAN-based~\cite{karras2019style} disentangled representations of original images. This can be considered a defense against unrestricted perturbations.
A margin-maximization variant of adversarial training was proposed in \cite{ding2019mma} that creates adversarial examples using sample-specific $\eta$ (see Eq.~\ref{eq:conventional}) instead of a fixed $\eta$ value across the training samples. The $\eta$ value corresponds to the ``shortest successful perturbation'' that fools the model. 

Among other variants of adversarial training frameworks, we  have \cite{balunovic2019adversarial} where in each training iteration, the model is verified for robustness using convex relaxation and adversarial examples are computed under that relaxation for training purpose. 
Vivek et al.~\cite{vivek2020single} also proposed a dropout scheduling method to improve the efficacy of adversarial training with single-step methods. 
To improve generalization of adversarially trained models, Song et al.~\cite{song2019robust} proposed Robust Local Features for Adversarial Training (RLFAT) that employs random block shuffle of the input during training. 
Farnia et al.~\cite{farnia2018generalizable} also proposed a spectral normalization based regularization for adversarial training to address the generalization issue.
In \cite{xiao2020one}, enhancement is suggested by using adversarial examples generated by attacking a model other than the model to be defended. 
To make adversarial training more efficient, Zheng et al.~\cite{zheng2020efficient} proposed to use the same adversarial perturbations across multiple epochs during the training. This reduces the number of computations in the overall training process while achieving acceptable performance.  

Naseer et al.~\cite{naseer2020self} proposed  self-supervised adversarial training, whereas adversarial training is independently analyzed for self-supervision by incorporating it in pretraining in \cite{chen2020adversarial}. Similarly, \cite{jang2019adversarial}  used  a generator in adversarial training to generate more diverse adversarial examples.  A Dual Manifold Adversarial Training (DMAT) is proposed in \cite{lin2020dual}, which uses perturbations in the image space as well as latent space of StyleGAN to make training more effective. In another related work, Wang et al.~\cite{wang2019bilateral} proposed a bilateral adversarial training that not only perturbs input images during training, but also their labels. The authors claimed improvements in state-of-the-art adversarial training results with this modification.
It is often argued that  adversarial training leads to the requirement of larger models. Ye et al.~\cite{ye2019adversarial} proposed a concurrent adversarial training and weight
pruning strategy to address this specific issue.

Considering further variants of adversarial training, 
Don et al.~\cite{dong2020adversarial} proposed an adversarial distributional training. Their method also formulates adversarial training as a minimax problem, however, the inner maximization is aimed at learning an adversarial distribution under an entropic regularizer. The outer minimization problem minimizes the loss over the worst-case adversarial distributions.
Madaan et al.~\cite{madaan2020adversarial} proposed a vulnerability suppression loss that minimizes the expected difference between latent features of the network on clean and their corresponding adversarial images. They further learned a pruning mask that explicitly minimizes adversarial loss by pruning features with high distortion.    
In an attempt to address multiple perturbation models for adversarial training, Maini et al.~\cite{maini2020adversarial} proposed to incorporate several perturbations into a single attack by taking the worst-case over the entire steepest directions as an extension to the standard PGD. It is claimed that their approach produces state-of-the-art robust classifiers against $\ell_1, \ell_2,$ and $\ell_\infty$ norm bounded perturbations   simultaneously.

In the literature, we also find methods that conceptually relate to adversarial training closely  without presenting themselves as such. For example, in \cite{zhang2017mixup}, a mixup training of neural networks was introduced. The main concept of the method  is to augment  training data with additional samples that are created as convex linear combination of the already  available samples. The same is done to the labels of the combined samples to provide the label of the resulting samples. It is shown that besides improving accuracy of the original model, this practice also helps in robustness against  adversarial samples~\cite{verma2018manifold}. Pang et al.~\cite{pang2019mixup} takes this notion further by also applying mixup of samples in the inference phase. 
A related adversarial vertex mixup method is adopted in \cite{lee2020adversarial} to achieve better adversarial generalization of the models. 


We also find multiple contributions in the literature that focus on \textit{analyzing} adversarial training instead of devising its variants. For example, Xie et al.~\cite{xie2019intriguing} have reported some interesting properties of adversarial training. The most intriguing ones include an improvement in the adversarial robustness for the process with separate Batch Normalization for clean and adversarial images, and a consistent improvement in the adversarial robustness with even deeper models as compared to the popular depth limits among the visual models. 
Li et al.~\cite{li2019implicit} also analyzed the implicit bias of gradient descent on adversarial training on separable data. Their findings theoretically back the efficacy of adversarial training for robustness. In \cite{shafahi2019adversarially}, it is also demonstrated that transfer learning on adversarially robust models retains (to an extent) the robustness effect for the target domain. Sehwag et al.~\cite{sehwag2020hydra} also devised a method for an adversarial training-aware model pruning in resource constrained environment.

Wong et al.~\cite{wong2020fast} showed that adversarial training with FGSM combined with random initialization is as effective as adversarial training with the first order PGD attack. On their computational setup, they trained a robust CIFAR10 classifier with 45\% robust accuracy in 6 minutes as compared to the 10 hours training of PGD-based counterpart that achieves similar results. Their improvement of adding randomization with FGSM-based adversarial training is, however, contradicted to an extend  by \cite{andriushchenko2020understanding}. Zhao et al.~\cite{zhao2020bridging} study the mode connectivity of loss landscape of adversarially robust and regular models, demonstrating the existence of robustness loss barrier for the former. 
Wu et al.~\cite{wu2020adversarial} showed that many adversarial training improvements appearing in the literature, e.g.~early stopping, new objective functions, or exploiting unlabeled data, implicitly flatten the weight loss landscape (i.e.~loss change w.r.t.~weights). Hence, they proposed an Adversarial Weight Perturbation (AWP) that directly regularizes the flatness of weight loss landscape, and can be used to improve adversarial training.

Even though Madry et al.~\cite{madry2017towards} have justified adversarial training by robust optimization theory, it is still unclear how adversarial training results in low robust training loss. Gau et al.~\cite{gao2019convergence} provide a theoretical analysis of adversarial training to explain its success using Neural Tangent Kernel and tools from online learning. They also prove that more model capacity is required for robust interpolation. However, their approach is limited to networks with  exponential width and run time. Zhang et al.~\cite{zhang2020over} extend their work for situations where the width of the network and its run time is polynomial in input dimension. They also extend the results to ReLU activation function.
Another related method~\cite{pang2020boosting} proposes to boost adversarial training by embedding a hypersphere method in the training process by regularizing features onto a compact manifolds. 


The literature also contains instances in which adversarial training is moulded to specific task requirements. For example, Wu et al.~\cite{wu2019defending} proposed an adversarial training method in which the adversarial samples are generated specifically keeping in view the physical world attacks. It is noted in \cite{wu2019defending}  that commonly used adversarial training and randomized smoothing for the digital attacks do not perform well for the physical world attacks. Hence, the modification was proposed. 
Instead of focusing on robustness against adversarial attacks, Zhu et al.~\cite{zhu2019freelb} employ adversarial training in natural language understanding for achieving higher embedding space invariance by perturbing the word embeddings. This is reported to result in better generalization of language models. This result also  resonates with the observations of \cite{cheng2019robust}.

\subsubsection{Other model modifications}
Besides adversarial training that focuses on modifying model weights through alternate training samples, there are multiple approaches that alter the basic building blocks of the model to incorporate adversarial robustness through regular training data. 
For instance, Pang et al.~\cite{pang2019rethinking} suggested to replace the softmax cross-entropy loss with a new loss, called Max-Mahalanobis center loss to induce adversarial  robustness in the model.
Xiao et al.~\cite{xiao2019enhancing} proposed to alter the ReLU activations with a $k$-winner-takes-all $C^0$ discontinuous to secure models  against the gradient-based attacks. 
There are also works that advocate on modifying  the networks in a holistic manner. For instance, in \cite{sen2020empir}, the authors suggest using quantized models for robustness against  gradient-based attacks. 
Guo et al.~\cite{guo2020meets} also proposed RobNets, designed  with neural architecture search, which are claimed to provide up to 5\% gain in robust accuracy on large datasets, e.g.~ImageNet.
Bui et al.~\cite{bui2020improving} propose an Adversary Divergence Reduction Network that can be used in conjunction with adversarial training for improved robustness. Similarly, a Bayesian neural network is proposed in \cite{liu2018adv} for adversarial robustness.

From the viewpoint of altering internal components of models,
Jeddi et al.~\cite{jeddi2020learn2perturb} proposed perturbation-injection modules in the internal layers of model during training and testing and used an alternating back-propagation scheme to train the network. A 4-7\% improvement in robustness over adversarial training with FGSM and PGD ($\ell_{\infty}$) is claimed by the authors. 
Li et al~\cite{li2020enhancing} introduced image restoration and denoising modules in the network and constrained its classification layer's Lipschitz constant for adversarial robustness.  
In the context of defense against the universal perturbations \cite{moosavi2017universal}, \cite{borkar2020defending} devised  a method to identify adversarially vulnerable convolutional filters in a model and introduces `regeneration units' to generate resilient features for those filters to avoid fooling.  
Wang et al.~\cite{wang2019direct} proposed to model adversarial noise with a generator that is trained jointly with a discriminator classifier and showed its effectiveness against balck-box attacks.
Xie et al.~\cite{xie2019feature} suggested that
adversarial perturbations result in noisy features of the  networks. Hence, they proposed networks containing denoising blocks with  non-local means or other filters.
Building on the idea of injecting noise in the network while training \cite{lecuyer2019certified}, \cite{liu2018towards}, He et al.~\cite{he2019parametric} proposed a trainable Gaussian model for injecting the noise.
A family of CNNs that alternate between the Euclidean convolutions and graph convolutions to leverage the information from the graph of peer samples is proposed in \cite{svoboda2018peernets}.

Another emerging model alteration approach to defend against adversarial attacks is through search for robust architectures. Following this paradigm, Hosseini et al.~\cite{hosseini2021dsrna} propose DSRNA to search for robust architectures via two differentiable metrics for robustness. Moreover,  Cazenavette et al.~\cite{cazenavette2021architectural} proposed a deep pursuit algorithm that formulate the architecture search as a global sparse coding problem that jointly computes all network activations.

We also witness techniques that approach at adversarial robustness from the model regularization perspective. For example, based on the observation that Jacobian of adversarially robust models are more salient and interpretable as compared to their non-robust counterparts~\cite{tsipras2018robustness}, Chan et al.~\cite{chan2019jacobian} proposed a Jacobian-based GAN-like regularization scheme to show improved robustness. 
A joint gradient phase and magnitude regularization was proposed in \cite{dabouei2020exploiting} to improve robustness of ensemble models. 
A concept of biologically inspired post-learning sleep phase of neural networks was introduced in \cite{tadros2019biologically}. The proposed technique allows a trained network to reflect on its statistics in an unsupervised manner and alter the weights to avoid over-fitting to the training data. Addepalli et al.~\cite{addepalli2020towards} proposed to regularize models with Bit-plane consistency as an efficient alternate for adversarial training. 

We note that whereas we discuss the above methods separately from adversarial training to provide a better structure to the literature, the boundary  separating these two lines of research is often abstract. One can understand adversarial training as a more fundamental framework that can generally be combined with other defenses, including those discussed in the  subsequent sections, for improved robustness. Other methods discussed above often demand model modifications that are less generic.

\subsection{Detection for defense}
Instead of proactively inducing a robust model during the training phase, there are also techniques that provide add-on mechanisms and modules for pre-trained models to defend them against adversarial attacks. Mostly, these methods are limited to detecting the presence of adversarial perturbations in the input during inference. Based on our earlier survey~\cite{akhtar2018threat} and recent literature, we can say that this line of research is getting slightly less popular (as compared to its earlier years)  in the leading research sources of computer vision and machine learning. A possible reason for that is their ad-hoc nature as compared to defenses like adversarial training. Nevertheless, we still witness interesting techniques of adversarial detection using add-on mechanisms in the recent literature.

Qin et al.~\cite{qin2019detecting} proposed  a mechanism of  class-conditional reconstruction  of images to detect adversarial examples during test time. The authors also introduced an attack to overcome this defense, demonstrating better robustness of CapsNet~\cite{sabour2017dynamic} over CNNs for their attack. More importantly, their attack shows more visual similarity between the adversarial examples and target object category for CapsNet. In essence, this demonstrates a larger perceptual alignment between CapsNet representation and human visual system as compared to CNNs. For reference, perceptual alignment between deep visual models and human vision is also discussed at length in \cite{akhtar2021attack}, \cite{jalwana2020attack}.  In \cite{Deng_2021_CVPR}, the authors proposed to leverage Lightweight Bayesian neural networks for task agnostic detection of adversarial perturbations in inputs using Bayes principle. The technique replaces last few layers of the attacked model with Bayesian module and performs detection-oriented fine-tuning that allows to maintain original performance while enabling detection.

Li et al.~\cite{li2020connecting} proposed to use context inconsistency of adversarial patterns in images for their detection using an external mechanism. For face recognition,  Tao et al.~\cite{tao2018attacks} proposed a method to identify internal neurons corresponding to critical facial attributes. By amplifying activation of these neurons, they construct an attribute-steered model. Later, they detect adversarial examples by identifying inconsistencies between the original and the attribute-steered models.  
In \cite{qiu2019adversarial}, the authors proposed a mechanism to  trace the activation paths of clean and adversarial images and detect adversarial perturbations based on the different characteristics of these paths. Liu et al.~\cite{liu2019detection} proposed to detect adversarial examples by analysing inputs from steganography point of view.  Their method estimates the probability of modification to images keeping in view adversarial perturbations.
Yin et al.~\cite{yin2019gat} introduced a so-called generative adversarial training method that learns an adversarial example detector. To robustify the detector against adaptive attacks, the authors employed asymmetric adversarial training.

\subsection{Input transformations for defense}
Instead of focusing on model robustness to `adversarial' inputs, transformation based methods aim at cleaning inputs to make them benign for the target model. For instance, JPEG-based compression of input has been studied for removing adversarial perturbations from images~\cite{liu2019feature}, \cite{das2017keeping}, \cite{guo2017countering}. Compressed adversarial images have been found to significantly loose their fooling abilities. Generally, input transformation provides the benefit that it can be easily used in conjunction with other defense mechanisms, e.g.~with adversarialy trained models. In some cases, different input transformations are also combined to improve their collective strength. For example, in \cite{raff2019barrage}, Raff et al.~proposed to stochastically combine multiple input transformations to also secure their defense against adaptive attacks. However, it is also observed in \cite{raff2019barrage} that more transformations undesirably lead to significant reduction  in model performance on clean images.
Similarly, \cite{taran2019defending} also proposed to utilize a set of random input transformations as an adversarial defense. The main idea behind this  method is that the `key' controlling the  randomization of transformations is assumed to be kept secret during test time. This mitigates the risk of potential adaptive attacks on their defense.

Instead of directly using standard image compression, learnable compression methods that use neural models are also  proposed in the literature for adversarial defense  \cite{jia2019comdefend}, \cite{theagarajan2019shieldnets}. In \cite{theagarajan2019shieldnets}, an external defender module for a deployed model is learned that projects inputs to a so-called adversarial-free data zone for the target model. We can also categorize learning-based compression techniques as defense mechanisms altering the models by appending add-on modules to them.
In another related work, Sun et al.~\cite{sun2019adversarial} transformed an input image using convolutional sparse coding. Their method use a `Sparse Transformation Layer' to project input to a quasi-natural space that is claimed to be less sensitive to adversarial manipulation.

In \cite{samangouei2018defense}, Samangouei et al.~presented one of the first examples of input transformation using GANs. Their method, Defense-GAN learns the
distribution of clean images. For inference, it computes an output close to the input image, which does not contain the potential adversarial perturbation.
A desnoising based defense is proposed by \cite{gupta2019ciidefence} that selectively denoises high attention regions of an image to recover the correct label. 
Kuo et al.~\cite{kou2019enhancing} noted that when input transformation is employed as a defense technique~\cite{guo2017countering}, the softmax distribution characteristics can be used to improve the clean image accuracy of the classifier with the help of an external lightweight classifier trained on the softmax distribution of clean images. 

In \cite{yuan2020ensemble}, an ensemble generative cleaning with feedback loop is  proposed to clean the image from adversarial patterns. Their method also relies on external generative modules to denoise adversarial images. Cohen et al.~\cite{cohen2020detecting} developed an external detector mechanism for adversarial samples using a so-called influence function.
This function measures the impact of all training samples on the validation data to provide sample influence scores. Supportive training instances for validation samples are identified with their scores. A k-nearest neighbor (k-NN) model is also fitted on the models activations to compute a ranking of the supportive training samples. Supportive samples are claimed to be highly correlated with the nearest neighbors of clean test sample, while
the correlation is found to be weaker for adversarial inputs.

\subsection{Certified defenses}
\label{sec:certified}
Although the literature is witnessing multiple defense techniques, it is shown that stronger attacks can be formed to defeat the existing defense methods~\cite{athalye2018obfuscated}, \cite{carlini2017adversarial}, see \S~\ref{sec:break} for more examples. Even adversarial training has its problems  despite being widely considered a reliable defense strategy. For instance, adversarially trained models with $\ell_{\infty}$-norm bounded perturbations are still found vulnerable to $\ell_p$-norm perturbations, where $p \neq \infty$~\cite{schott2018towards}, \cite{croce2019scaling}.
Certified defenses attempt to provide guarantee that the target model can not be fooled within an $\ell_p$-ball of the clean image. This guarantee is either achieved by computing the minimal $\ell_p$-norm of the perturbation to break the provided defense~\cite{katz2017reluplex}, \cite{tjeng2017evaluating}; or by providing a lower bound on the norm \cite{hein2017formal}, \cite{raghunathan2018certified}, \cite{wong2018provable}. There are also other methods that aim to both enhance network robustness and produce models that are more amenable to robustness verification techniques~\cite{mirman2018differentiable}, \cite{xiao2018training}.
Nevertheless, most of the certified defenses are able to prove their robustness against only one kind of bound on the perturbation, e.g.~$\ell_2$, $\ell_{\infty}$, struggling to provide generic bounds for multiple  $\ell_p$-norms simultaneously~\cite{croce2020provable}, with a few exceptions \cite{schott2018towards}, \cite{tramer2019adversarial}.

Corce and Hein~\cite{croce2020provable} recently proposed a regularization scheme for ReLU networks to enforce robustness against $\ell_1$ and $\ell_{\infty}$ attacks and showed that it results in provable robust models for any $\ell_p$ norm, where $p \geq 1$. As opposed to providing certified robustness for top-1 predictions, Jia et al.~\cite{jia2019certified} derived tight robustness in $\ell_2$-norm using Gaussian randomized smoothing for top-k predictions. 
Their method builds on the notion of randomized smoothing introduced in \cite{cao2017mitigating} and \cite{cohen2019certified}. Zhai et al.~\cite{zhai2020macer} also built on the insights of \cite{cohen2019certified} to develop a method for MAximizing the CErtified Radius (MACER) of the models that is claimed to be scalable to large models.

Fischer et al.~\cite{fischer2020certified} also extended the notion of randomized smoothing to incorporate parameterized transformations (e.g., translations, rotations) and certified the robustness of models in parameter space (e.g., rotation angle).
Another example of using randomised smoothing for a certifiable defense can  be found in \cite{levine2020randomized}. This defense is aimed at patch attacks, and it provides certificate against given image and patch size. For patch attacks, more certified defenses are studied in~\cite{brown2017adversarial},  \cite{chiang2020certified}, \cite{awasthi2020adversarial}. 
Zhang et al.~\cite{zhang2020black} extended the Gaussian smoothing noise in randomized classifiers to non-Gaussian noise. They designed a family of non-Gaussian smoothing distributions that works more efficiently against $\ell_1$, $\ell_2$, and $\ell_\infty$ attacks.

As noted earlier, the direction of certified defenses is gradually becoming quite popular in adversarial machine learning literature. Incidentally, the  problem is attracting more interest of machine learning community as compared to the compute vision community. Nevertheless, inspirations for the techniques developed along this line of research are coming from different directions. For instance, Rahnama et al.~\cite{rahnama2020robust} treat the networks from a control theory perspective to  provide tight bounds on any layer's response to adversarial examples.
Generally, randomised smoothing is the most commonly utilised tool be certified defense tools, which relates to adversarial in essence. 
Further recent examples of certified defenses can be found in \cite{saralajew2020fast}, \cite{zhang2020tightness}, \cite{goldwasser2020beyond}, \cite{salman2020denoised}, \cite{Awasthi_2021_CVPR}. 


\subsection{Miscellaneous methods}
\label{sec:defOther}
Among defenses, there are numerous works  that either propose methods for specialized tasks, networks or attack types. There are also techniques that mainly  focus on improving the defense strength by combining multiple defense strategies discussed above. This section provides a summary of such works in the recent literature.

Cemgil et al.~\cite{cemgil2019adversarially} analyzed the susceptibility of  Variational Auto-Encoders (VAEs) to adversarial examples. They identified `evidence lower bound' as one of its major  causes, which is addressed by a data augmentation strategy during training in their work. 
He et al.~\cite{he2020defending} specifically proposed a binarization aware training method to defend against the Bit Flip Attack~\cite{rakin2019bit}.
Robustness of Bayesian networks to gradient-based attacks is studied in \cite{carbone2020robustness}. Similarly, inherent robustness of spiking neural networks is the main topic of discussion in \cite{sharmin2020inherent}. 
Defending Graph Neural Networks (GNNs) is studied in \cite{zhang2020gnnguard}. 
Among other specialized defenses, differential privacy is used in \cite{du2019robust} to detect poisoning samples for backdoor attacks. 
There are also methods that focus entirely on defending neural models against the universal adversarial perturbations~\cite{mummadi2019defending}, \cite{akhtar2018defense}
Cost sensitive adversarial robustness is studied in \cite{zhang2018cost}, whereas a so-called `guided complement entropy' loss is proposed in  \cite{chen2019improving}, claiming better robustness over the standard cross entropy loss.

There are also examples providing specialized defenses for computer vision tasks other than standard classification, e.g.~tracking~\cite{jia2020robust}, open-set recognition~\cite{shao2020open}, face recognition~\cite{zhou2020manifold}.
Goldblum et al.~\cite{goldblum2020adversarially} proposed a method to infer  robust models for few-shot classification tasks based on adversarially robust meta-learners.
A prediction poisoning attack is adopted as a defense against the model stealing attacks in \cite{orekondy2019prediction}. The technique systematically alters the prediction of a target model to maintain the original performance but poison any model trained to steal the target model. A similar approach is taken in \cite{kariyappa2020defending} by selectively making incorrect prediction for out-of-distribution queries to avoid model stealing.

Methods analyzing defense mechanisms and robustness instead of proposing new specialized defenses are also found for this category of defenses. For instance, 
\cite{sulam2020adversarial} anaylzes the robustness of sparse coding to adverarial examples.
It is observed in \cite{yang2020multitask} that multitask learning generally results in improving adversarial robustness of the models. Kim et al.~\cite{kim2020modeling} claim that by leveraging sparsity and other perceptual biological mechanisms, adversarial robustness of models can be improved.
Wang et al.~\cite{wang2020once} studied how to  calibrate a trained model in-situ, in order to analyze the achievable trade-offs between the standard and robust accuracy of the model. Trade-off between the backdoor and adversarial robustness of models is studied in \cite{weng2020trade}. Chen et al.~\cite{chen2020anti} proposed to use Neural Architecture Search to find adversarially robust architectures. Adversarial robustness in the more practical scenario of long-tailed data distribution is analyzed in \cite{Wu_2021_CVPR}. The authors combine adversarial training with the existing recognition methods for imbalanced and long-tailed data to highlight interesting properties of models. For instance, it is shown that unreliable evaluation can easily give  fake robustness gain impression for these models. 
\section{Discussion}
\label{sec:Disc}
Since its advent in 2013, the problem of adversarial attacks and lack of defenses for deep learning has intrigued the computer vision community considerably. Currently, this research direction  is more active than ever. We found an ever-increasing number of papers appearing in the leading research sources of computer vision. The mainstream venues of machine learning research are also publishing papers with almost the same frequency as the computer vision venues. Interestingly, we found that most works in the direction of adversarial attacks and defenses appearing in machine learning sources still use `visual models' as their test-bed. Nevertheless, we find a particular interest of the machine learning community in robustification of the models instead of devising new methods of fooling them. Of 400+ papers identified among the top six computer vision and machine learning venues in the last three years, we find around $74\%$ papers dealing with defense techniques  in machine learning venues. In contrast, only $40\%$ of the papers in computer vision venues make adversarial defense as their central topic. 

Among many interesting sub-problems in this area, the problem of `black-box' attacks under better transferability and query-based setup is gaining significant popularity in computer vision research sources. In parallel, the topics of `adversarial training' and `provable/certified' defenses currently stand out in the literature appearing in machine learning sources. Below we summarize a few general trends that we observed in the literature. We intentionally keep the discussion at a higher-level of abstraction while covering the broader direction. The reader is encouraged  to visit the related sections of the article to observe these trends with specific instances.  

\subsection{General Trends and Challenges}
\paragraph{Adversarial Attacks}
Whereas the first generation of attacks explored new core tools to fool deep visual classifiers, the more recent attacks are concerned with utilizing those tools for more specific fooling objectives. Gradient ascend over the loss surface of the model is arguably the most common (and effective) tool for adversarial attacks in the literature. An overwhelming majority of the existing white-box attacks and transfer-based black-box attacks use this tool in some form to compute the additive adversarial perturbations. Model gradients are sometimes also utilized to satisfy linearization assumptions used by the attacks that aim at exploiting the geometry of the classification regions of the model. The observation that model gradients are the central tool for adversarial attacks resounds with the fact that deep models are, in the end, differentiable programs. Nevertheless, there are also other tools and heuristics, e.g.~evolutionary algorithms, color-space search, that have been shown to find effective adversarial examples. As compared to model-gradients, such techniques are found to be more ad-hoc though.

The more recent core attack methods often aim at making the attacks more threatening by further reducing the norm of the perturbations and amplifying the transferability of the adversarial examples in black-box setups. Although  universal perturbations can be considered a more serious threat from a practical viewpoint, the vast majority of the existing literature ($>95\%$) is concerned with image-specific attacks. A major reason for that is, from the defense perspective, securing models against (stronger) image-specific attacks already provides some robustness against the universal attacks, because the adversarial objective of the latter is already more challenging than that of the former. Nevertheless, we still find active investigations related to specifically  securing models against the universal perturbations.

Since black-box attacks are gaining considerable popularity in the recent literature, it is worth summarizing some of the trends specifically in this direction. For the transfer-based black-box attacks, currently an accuracy reduction (of the target model) in the range $40-50\%$ with $\ell_{\infty}$ perturbation norm of $15/255$ is generally considered a good achievement in the recent literature for ImageNet models. This is true only for untargeted fooling though. The norm-bound is often considerably relaxed for the targeted black-box fooling (e.g.~up to $32/255$) without achieving fooling ratios at par with untargeted fooling with half the perturbation norm. We also observer that black-box attacks are reported to transfer better between the  models with architectural similarity. For instance, one can expect to see an accuracy reduction of $\sim50\%$ when transferring perturbations computed on inception-V3 to inception-V4. This number is expected to be $\sim25\%$ when those perturbations are transferred to a ResNet-50 model. These numbers are not hypothetical. We provide them by observing multiple contributions. However, since each attack method has its own specific algorithm, the exact transfer rates may vary. We intentionally do not associate these numbers to specific methods, and only provide a rough estimate as a general guide to the readers.

One surprising trend we observed in the literature is about the evaluation of  transfer-based black-box attacks. The term  `black-box' is understood by the community as a setup where the attacker does not have `any' information about the target model (except its output in query-based setup). However, the existing methods generally report the attack transfer rates on `ImageNet' models while also computing the perturbations on the `ImageNet' models. In essence, this setup entails  complete knowledge of the training data of the target model, which violates the definition of `black-box' setup. Strictly speaking, the target models  must be trained on unseen data, and should have, e.g.~unknown number of output labels. We suggest that the research community considers this aspect in evaluating the transfer-based black-box attacks.     

Among the query-based black-box attacks, the boundary attacks are more popular -  outweighing their score-based counterpart by $\sim 5$ to $1$. Generally, the query-based attacks optimize for two contradictory objectives of (a) achieving high fooling rates with stronger perturbations that use minimum number of queries, (b) keeping the perturbations imperceptible by restricting their norm. The most widely used strategy is to first query the black-box model with large perturbations, and then reduce the perturbation norm with a refinement mechanism while maintaining the incorrect prediction. We witness a large variation in the achieved fooling ratios and the number of queries utilized by different methods in this direction. It is clear from the reported results that these values depend rather strongly on image size. For ImageNet sized images, current literature considers 20K to 100K+ queries per image to still be reasonable to achieve imperceptible perturbations. 
This number drastically reduces to $\sim 1$K for image sizes of $32\times 32$. 

From the perspective of threat of adversarial attacks in the physical world, we do not find research to be as active as in the digital domain. One reason for that is the processes involved in physically realizing the computed adversarial patterns are often cumbersome and time taking. This does not mix well with the extremely fast pace of this research direction in the digital domain. Hence, even the attacks devised for the physical world applications are often just evaluated in simulated environments, e.g.~camouflage cars for autonomous driving. Whereas we did not find any convincing argument that could suggest that physical world attacks are not a real concern to vision systems, we do find the adversarial samples for the physical world to be more conspicuous. Generally, such samples can be marked by obvious unnatural/irrelevant geometry or texture. Such a  compromise over the stealthiness of the attack directly comes from the fact that visual sensors digitize only the 'visible' information. Hence, the adversarial patterns have to be visible to the sensor. Hiding the adversarial patterns from humans by semantically blending them in the scene environment is then the obvious choice for imperceptibility of the attacks in the physical world.             

One interesting emerging utility of adversarial attacks is in explaining deep visual models. Considering that model gradients are utilized by both attack methods and popular model explanation methods, e.g., Grad-CAM~\cite{selvaraju2017grad}, CAMERAS~\cite{jalwana2021cameras}, it is not surprising that this overlap is emerging. Since deep learning models are differentiable programs, one can expect adversarial perturbations (which are a processed form of gradient information) can carry a signature of that program. From another perspective, perturbations can also be expected to focus (in some form) more on the salient regions of object to take model's attention away from those regions. This notion also resounds with image saliency. Hence, researchers are also getting interested in explainability of the perturbation itself.  

Beyond the above-mentioned trends and challenges, the recent attacks are gradually getting more and more specialized to specific vision tasks and data modalities. Nevertheless, since the discovery of adversarial perturbations, their theoretical understanding has always been a topic of debate. Though a number of hypotheses exist in the literature on the susceptibility of deep learning to adversarial attacks, there is no single theory to fully explain all the observed phenomena in this direction. The adversarial vulnerability of deep models seem to emerge from a number of processes, and the debate on its existence and theory to explain all its aspect can be expected to stay as a long-standing problem for this research direction.    










\paragraph{Adversarial defenses}
Whereas a large number of defenses against adversarial attacks are appearing in the literature, arguably the most promising stream of works still concerns itself with `adversarial training'. Interestingly, the concept of adversarial training was presented simultaneously with adversarial perturbations in the original work of Szegedy et al.~\cite{szegedy2013intriguing}. Most of the later literature significantly digressed from this original idea of robustifying the models. However, the later defense strategies mostly rely on ad-hoc rules and heuristics. Many of those are also shown to be broken with stronger attacks or different attack conditions~\cite{Yu_2021_CVPRLAFEAT}. 
In fact, recently, Tramer et al.~\cite{tramer2020adaptive} also show that thirteen different defenses that actually account for adaptive attack strategies can also be broken.  
From the defense perspective, the research community (especially machine learning community) is focusing more on adversarial training and certified defenses due to their principled nature. Nevertheless, reduction in the accuracy of the robust models on clean images is a major challenge at this front. It is easy to observe in the literature that methods withstanding stronger attacks have proportionally low accuracy on clean images.




\paragraph{Future outlook}
Considering an ever-increasing influx of research papers in adversarial attacks (and defenses) since the advent of this direction, we can easily predict high research activity in this direction in the near-future. From the attacks perspective, whereas white-box attacks are likely to keep building around the tools used by the first-generation attack methods, a variety of new techniques for black-box setup can be anticipated. This is especially true for query-based attacks that is gaining increasing interest of the research community. Naturally, we can also anticipate the attacks to soon  circumscribe Transformer models in vision, which are gaining popularity in computer vision community~\cite{khan2021transformers}. 

Based on the existing literature, we can argue that topics like understanding the existence of adversarial examples, intrinsically robust models, robustness-accuracy trade-off, adversarial training, certified defenses; are gradually adapting into long-standing problems of this direction. Hence, we also expect a multitude of works directed to address these problems in the future. We are likely to see mergence of  adversarial perturbation techniques with other related directions, e.g.~deepFakes~\cite{tolosana2020deepfakes}, backdoor attacks~\cite{liu2017trojaning}. Specifically, a potential interesting scenario is adopting the adversarial objective of perturbations to independently fool the detectors of deepFakes and backdoor attacks. We are also likely to witness more activity in terms of expansion of adversarial attacks through visual models to multi-model tasks, e.g.~image/video captioning~\cite{aafaq2019video} which combines visual models with language models, providing the opportunity to control the latter by attacking the former.

Since adversarial examples question the core utility of deep learning of making `reliable' automated decisions, the research direction of adversarial attacks (and their defenses) seems to be here to stay with deep learning research. Just like deep learning is finding utilities in all applications, adversarial attacks are gradually adapting to those applications as its nemesis. From the viewpoint of this research direction, this arm race is promising, but not so much for deep learning in practice.          
\section{Conclusion}
\label{sec:Conc}
In this article, we reviewed the research direction of adversarial attacks and defenses for deep learning models, focusing on the visual models. Since its advent in 2013, this direction has particularly intrigued the computer vision community, which has led to a large influx of papers in the recent years. To ensure the authenticity and quality of the discussed contributions, the survey mainly focused on the papers published in the top-ranked sources of computer vision and machine learning research. For standardising  technical  terminologies in this relatively new research direction, the survey also provided a list of definitions of the frequently used terms in the related literature. It also presented a detailed discussion on the early contributions in adversarial attacks to provide a historical account of the overall direction. The presented review builds on the first-ever peer-reviewed survey in this direction~\cite{akhtar2018threat} - co-authored by the authors of this survey - as a legacy sequel. In~\cite{akhtar2018threat}, literature until 2018 is covered thoroughly. Hence, this article focused on the more recent literature, published after 2018. The covered literature is divided into attacks and defenses methods, which are further broken down into sub-topics by clustering the papers. This provided a clear indication of the current and emerging trends in the literature, that we discussed and reflected upon explicitly after reviewing the literature.

\section*{Acknowledgment}
This material is based upon work supported by the Defense Advanced Research Projects Agency (DARPA) under Agreement No. HR00112090095, and the Australian Research Council Discovery Grant DP190102443. Dr.~Naveed Akhtar is the recipient of an Office of National Intelligence
National Intelligence Postdoctoral Grant funded by
the Australian Government.


\ifCLASSOPTIONcaptionsoff
  \newpage
\fi

\bibliographystyle{IEEEtran}
\bibliography{ForArxiv}

\begin{thebibliography}{100}
\providecommand{\url}[1]{#1}
\csname url@samestyle\endcsname
\providecommand{\newblock}{\relax}
\providecommand{\bibinfo}[2]{#2}
\providecommand{\BIBentrySTDinterwordspacing}{\spaceskip=0pt\relax}
\providecommand{\BIBentryALTinterwordstretchfactor}{4}
\providecommand{\BIBentryALTinterwordspacing}{\spaceskip=\fontdimen2\font plus
\BIBentryALTinterwordstretchfactor\fontdimen3\font minus
  \fontdimen4\font\relax}
\providecommand{\BIBforeignlanguage}[2]{{%
\expandafter\ifx\csname l@#1\endcsname\relax
\typeout{** WARNING: IEEEtran.bst: No hyphenation pattern has been}%
\typeout{** loaded for the language `#1'. Using the pattern for}%
\typeout{** the default language instead.}%
\else
\language=\csname l@#1\endcsname
\fi
#2}}
\providecommand{\BIBdecl}{\relax}
\BIBdecl

\bibitem{szegedy2013intriguing}
C.~Szegedy, W.~Zaremba, I.~Sutskever, J.~Bruna, D.~Erhan, I.~Goodfellow, and
  R.~Fergus, ``Intriguing properties of neural networks,'' \emph{arXiv preprint
  arXiv:1312.6199}, 2013.

\bibitem{akhtar2018threat}
N.~Akhtar and A.~Mian, ``Threat of adversarial attacks on deep learning in
  computer vision: A survey,'' \emph{IEEE Access}, vol.~6, pp.
  14\,410--14\,430, 2018.

\bibitem{lecun2015deep}
Y.~LeCun, Y.~Bengio, and G.~Hinton, ``Deep learning,'' \emph{nature}, vol. 521,
  no. 7553, pp. 436--444, 2015.

\bibitem{xiong2015human}
H.~Y. Xiong, B.~Alipanahi, L.~J. Lee, H.~Bretschneider, D.~Merico, R.~K. Yuen,
  Y.~Hua, S.~Gueroussov, H.~S. Najafabadi, T.~R. Hughes \emph{et~al.}, ``The
  human splicing code reveals new insights into the genetic determinants of
  disease,'' \emph{Science}, vol. 347, no. 6218, 2015.

\bibitem{helmstaedter2013connectomic}
M.~Helmstaedter, K.~L. Briggman, S.~C. Turaga, V.~Jain, H.~S. Seung, and
  W.~Denk, ``Connectomic reconstruction of the inner plexiform layer in the
  mouse retina,'' \emph{Nature}, vol. 500, no. 7461, pp. 168--174, 2013.

\bibitem{amodio2019exploring}
M.~Amodio, D.~Van~Dijk, K.~Srinivasan, W.~S. Chen, H.~Mohsen, K.~R. Moon,
  A.~Campbell, Y.~Zhao, X.~Wang, M.~Venkataswamy \emph{et~al.}, ``Exploring
  single-cell data with deep multitasking neural networks,'' \emph{Nature
  methods}, pp. 1--7, 2019.

\bibitem{hickok2007cortical}
G.~Hickok and D.~Poeppel, ``The cortical organization of speech processing,''
  \emph{Nature reviews neuroscience}, vol.~8, no.~5, pp. 393--402, 2007.

\bibitem{manning1999foundations}
C.~Manning and H.~Schutze, \emph{Foundations of statistical natural language
  processing}.\hskip 1em plus 0.5em minus 0.4em\relax MIT press, 1999.

\bibitem{krizhevsky2012imagenet}
A.~Krizhevsky, I.~Sutskever, and G.~E. Hinton, ``Imagenet classification with
  deep convolutional neural networks,'' in \emph{Advances in neural information
  processing systems}, 2012, pp. 1097--1105.

\bibitem{deng2009imagenet}
J.~Deng, W.~Dong, R.~Socher, L.-J. Li, K.~Li, and L.~Fei-Fei, ``Imagenet: A
  large-scale hierarchical image database,'' in \emph{2009 IEEE conference on
  computer vision and pattern recognition}.\hskip 1em plus 0.5em minus
  0.4em\relax Ieee, 2009, pp. 248--255.

\bibitem{lecun1989backpropagation}
Y.~LeCun, B.~Boser, J.~S. Denker, D.~Henderson, R.~E. Howard, W.~Hubbard, and
  L.~D. Jackel, ``Backpropagation applied to handwritten zip code
  recognition,'' \emph{Neural computation}, vol.~1, no.~4, pp. 541--551, 1989.

\bibitem{he2016deep}
K.~He, X.~Zhang, S.~Ren, and J.~Sun, ``Deep residual learning for image
  recognition,'' in \emph{Proceedings of the IEEE conference on computer vision
  and pattern recognition}, 2016, pp. 770--778.

\bibitem{huang2017densely}
G.~Huang, Z.~Liu, L.~Van Der~Maaten, and K.~Q. Weinberger, ``Densely connected
  convolutional networks,'' in \emph{Proceedings of the IEEE conference on
  computer vision and pattern recognition}, 2017, pp. 4700--4708.

\bibitem{szegedy2016inception}
C.~Szegedy, S.~Ioffe, V.~Vanhoucke, and A.~Alemi, ``Inception-v4,
  inception-resnet and the impact of residual connections on learning,''
  \emph{arXiv preprint arXiv:1602.07261}, 2016.

\bibitem{silver2017mastering}
D.~Silver, J.~Schrittwieser, K.~Simonyan, I.~Antonoglou, A.~Huang, A.~Guez,
  T.~Hubert, L.~Baker, M.~Lai, A.~Bolton \emph{et~al.}, ``Mastering the game of
  go without human knowledge,'' \emph{nature}, vol. 550, no. 7676, pp.
  354--359, 2017.

\bibitem{szegedy2015going}
C.~Szegedy, W.~Liu, Y.~Jia, P.~Sermanet, S.~Reed, D.~Anguelov, D.~Erhan,
  V.~Vanhoucke, and A.~Rabinovich, ``Going deeper with convolutions,'' in
  \emph{Proceedings of the IEEE conference on computer vision and pattern
  recognition}, 2015, pp. 1--9.

\bibitem{goodfellow2014explaining}
I.~J. Goodfellow, J.~Shlens, and C.~Szegedy, ``Explaining and harnessing
  adversarial examples,'' \emph{arXiv preprint arXiv:1412.6572}, 2014.

\bibitem{Tesla:2020}
Tesla, ``Future of driving,'' 2020 (accessed August 25, 2020),
  \url{https://www.tesla.com/en_AU/autopilot}.

\bibitem{ATM:2020}
C.~Middlehurst, ``China unveils world’s first facial recognition atm,'' 2020
  (accessed August 25, 2020),
  \url{https://www.telegraph.co.uk/news/worldnews/asia/china/11643314/China-unveils-worlds-first-facial-recognition-ATM.html}.

\bibitem{FaceID:2020}
Apple, ``About face id advanced technology,'' 2020 (accessed August 25, 2020),
  \url{, https://support.apple.com/en-au/HT208108}.

\bibitem{grigorescu2020survey}
S.~Grigorescu, B.~Trasnea, T.~Cocias, and G.~Macesanu, ``A survey of deep
  learning techniques for autonomous driving,'' \emph{Journal of Field
  Robotics}, vol.~37, no.~3, pp. 362--386, 2020.

\bibitem{zulqarnain2018learning}
S.~Zulqarnain~Gilani and A.~Mian, ``Learning from millions of 3d scans for
  large-scale 3d face recognition,'' in \emph{Proceedings of the IEEE
  Conference on Computer Vision and Pattern Recognition}, 2018, pp. 1896--1905.

\bibitem{masi2018deep}
I.~Masi, Y.~Wu, T.~Hassner, and P.~Natarajan, ``Deep face recognition: A
  survey,'' in \emph{2018 31st SIBGRAPI conference on graphics, patterns and
  images (SIBGRAPI)}.\hskip 1em plus 0.5em minus 0.4em\relax IEEE, 2018, pp.
  471--478.

\bibitem{sunderhauf2018limits}
N.~S{\"u}nderhauf, O.~Brock, W.~Scheirer, R.~Hadsell, D.~Fox, J.~Leitner,
  B.~Upcroft, P.~Abbeel, W.~Burgard, M.~Milford \emph{et~al.}, ``The limits and
  potentials of deep learning for robotics,'' \emph{The International Journal
  of Robotics Research}, vol.~37, no. 4-5, pp. 405--420, 2018.

\bibitem{najafabadi2015deep}
M.~M. Najafabadi, F.~Villanustre, T.~M. Khoshgoftaar, N.~Seliya, R.~Wald, and
  E.~Muharemagic, ``Deep learning applications and challenges in big data
  analytics,'' \emph{Journal of Big Data}, vol.~2, no.~1, p.~1, 2015.

\bibitem{Carlini:20}
\BIBentryALTinterwordspacing
N.~Carlini, \emph{A Complete List of All (arXiv) Adversarial Example Papers},
  2020 (accessed October 1, 2020). [Online]. Available:
  \url{https://nicholas.carlini.com/writing/2019/all-adversarial-example-papers.html}
\BIBentrySTDinterwordspacing

\bibitem{arnab2018robustness}
A.~Arnab, O.~Miksik, and P.~H. Torr, ``On the robustness of semantic
  segmentation models to adversarial attacks,'' in \emph{Proceedings of the
  IEEE Conference on Computer Vision and Pattern Recognition}, 2018, pp.
  888--897.

\bibitem{he2019segmentations}
Y.~He, S.~Rahimian, B.~Schiele, and M.~Fritz, ``Segmentations-leak: Membership
  inference attacks and defenses in semantic image segmentation,'' \emph{arXiv
  preprint arXiv:1912.09685}, 2019.

\bibitem{tu2020physically}
J.~Tu, M.~Ren, S.~Manivasagam, M.~Liang, B.~Yang, R.~Du, F.~Cheng, and
  R.~Urtasun, ``Physically realizable adversarial examples for lidar object
  detection,'' in \emph{Proceedings of the IEEE/CVF Conference on Computer
  Vision and Pattern Recognition}, 2020, pp. 13\,716--13\,725.

\bibitem{zhang2019towards}
H.~Zhang and J.~Wang, ``Towards adversarially robust object detection,'' in
  \emph{Proceedings of the IEEE International Conference on Computer Vision},
  2019, pp. 421--430.

\bibitem{jia2019fooling}
Y.~Jia, Y.~Lu, J.~Shen, Q.~A. Chen, H.~Chen, Z.~Zhong, and T.~Wei, ``Fooling
  detection alone is not enough: Adversarial attack against multiple object
  tracking,'' in \emph{International Conference on Learning Representations},
  2019.

\bibitem{chen2020one}
X.~Chen, X.~Yan, F.~Zheng, Y.~Jiang, S.-T. Xia, Y.~Zhao, and R.~Ji, ``One-shot
  adversarial attacks on visual tracking with dual attention,'' in
  \emph{Proceedings of the IEEE/CVF Conference on Computer Vision and Pattern
  Recognition}, 2020, pp. 10\,176--10\,185.

\bibitem{zheng2020efficient}
H.~Zheng, Z.~Zhang, J.~Gu, H.~Lee, and A.~Prakash, ``Efficient adversarial
  training with transferable adversarial examples,'' in \emph{Proceedings of
  the IEEE/CVF Conference on Computer Vision and Pattern Recognition}, 2020,
  pp. 1181--1190.

\bibitem{zhou2018transferable}
W.~Zhou, X.~Hou, Y.~Chen, M.~Tang, X.~Huang, X.~Gan, and Y.~Yang,
  ``Transferable adversarial perturbations,'' in \emph{Proceedings of the
  European Conference on Computer Vision (ECCV)}, 2018, pp. 452--467.

\bibitem{moosavi2017universal}
S.-M. Moosavi-Dezfooli, A.~Fawzi, O.~Fawzi, and P.~Frossard, ``Universal
  adversarial perturbations,'' in \emph{Proceedings of the IEEE conference on
  computer vision and pattern recognition}, 2017, pp. 1765--1773.

\bibitem{akhtar2019label}
N.~Akhtar, M.~A. Jalwana, M.~Bennamoun, and A.~Mian, ``Label universal targeted
  attack,'' \emph{arXiv preprint arXiv:1905.11544}, 2019.

\bibitem{vinyals2019grandmaster}
O.~Vinyals, I.~Babuschkin, W.~M. Czarnecki, M.~Mathieu, A.~Dudzik, J.~Chung,
  D.~H. Choi, R.~Powell, T.~Ewalds, P.~Georgiev \emph{et~al.}, ``Grandmaster
  level in starcraft ii using multi-agent reinforcement learning,''
  \emph{Nature}, vol. 575, no. 7782, pp. 350--354, 2019.

\bibitem{yuan2019adversarial}
X.~Yuan, P.~He, Q.~Zhu, and X.~Li, ``Adversarial examples: Attacks and defenses
  for deep learning,'' \emph{IEEE transactions on neural networks and learning
  systems}, vol.~30, no.~9, pp. 2805--2824, 2019.

\bibitem{hao2020adversarial}
H.~X. Y.~M. Hao-Chen, L.~D. Deb, H.~L. J.-L.~T. Anil, and K.~Jain,
  ``Adversarial attacks and defenses in images, graphs and text: A review,''
  \emph{International Journal of Automation and Computing}, vol.~17, no.~2, pp.
  151--178, 2020.

\bibitem{ozdag2018adversarial}
M.~Ozdag, ``Adversarial attacks and defenses against deep neural networks: a
  survey,'' \emph{Procedia Computer Science}, vol. 140, pp. 152--161, 2018.

\bibitem{zhou2019adversarial}
Y.~Zhou, M.~Han, L.~Liu, J.~He, and X.~Gao, ``The adversarial attacks threats
  on computer vision: A survey,'' in \emph{2019 IEEE 16th International
  Conference on Mobile Ad Hoc and Sensor Systems Workshops (MASSW)}.\hskip 1em
  plus 0.5em minus 0.4em\relax IEEE, 2019, pp. 25--30.

\bibitem{vakhshiteh2020threat}
F.~Vakhshiteh, R.~Ramachandra, and A.~Nickabadi, ``Threat of adversarial
  attacks on face recognition: A comprehensive survey,'' \emph{arXiv preprint
  arXiv:2007.11709}, 2020.

\bibitem{brown2018unrestricted}
T.~B. Brown, N.~Carlini, C.~Zhang, C.~Olsson, P.~Christiano, and I.~Goodfellow,
  ``Unrestricted adversarial examples,'' \emph{arXiv preprint
  arXiv:1809.08352}, 2018.

\bibitem{song2018constructing}
Y.~Song, R.~Shu, N.~Kushman, and S.~Ermon, ``Constructing unrestricted
  adversarial examples with generative models,'' in \emph{Advances in Neural
  Information Processing Systems}, 2018, pp. 8312--8323.

\bibitem{brown2017adversarial}
T.~B. Brown, D.~Man{\'e}, A.~Roy, M.~Abadi, and J.~Gilmer, ``Adversarial
  patch,'' \emph{arXiv preprint arXiv:1712.09665}, 2017.

\bibitem{fletcher2013practical}
R.~Fletcher, \emph{Practical methods of optimization}.\hskip 1em plus 0.5em
  minus 0.4em\relax John Wiley \& Sons, 2013.

\bibitem{rozsa2016adversarial}
A.~Rozsa, E.~M. Rudd, and T.~E. Boult, ``Adversarial diversity and hard
  positive generation,'' in \emph{Proceedings of the IEEE Conference on
  Computer Vision and Pattern Recognition Workshops}, 2016, pp. 25--32.

\bibitem{miyato2018virtual}
T.~Miyato, S.-i. Maeda, M.~Koyama, and S.~Ishii, ``Virtual adversarial
  training: a regularization method for supervised and semi-supervised
  learning,'' \emph{IEEE transactions on pattern analysis and machine
  intelligence}, vol.~41, no.~8, pp. 1979--1993, 2018.

\bibitem{kurakin2016adversarial}
A.~Kurakin, I.~Goodfellow, and S.~Bengio, ``Adversarial examples in the
  physical world,'' \emph{arXiv preprint arXiv:1607.02533}, 2016.

\bibitem{dong2018boosting}
Y.~Dong, F.~Liao, T.~Pang, H.~Su, J.~Zhu, X.~Hu, and J.~Li, ``Boosting
  adversarial attacks with momentum,'' in \emph{Proceedings of the IEEE
  conference on computer vision and pattern recognition}, 2018, pp. 9185--9193.

\bibitem{xie2019improving}
C.~Xie, Z.~Zhang, Y.~Zhou, S.~Bai, J.~Wang, Z.~Ren, and A.~L. Yuille,
  ``Improving transferability of adversarial examples with input diversity,''
  in \emph{Proceedings of the IEEE Conference on Computer Vision and Pattern
  Recognition}, 2019, pp. 2730--2739.

\bibitem{szegedy2016rethinking}
C.~Szegedy, V.~Vanhoucke, S.~Ioffe, J.~Shlens, and Z.~Wojna, ``Rethinking the
  inception architecture for computer vision,'' in \emph{Proceedings of the
  IEEE conference on computer vision and pattern recognition}, 2016, pp.
  2818--2826.

\bibitem{kurakin2016adversarial_scale}
A.~Kurakin, I.~Goodfellow, and S.~Bengio, ``Adversarial machine learning at
  scale,'' \emph{arXiv preprint arXiv:1611.01236}, 2016.

\bibitem{madry2017towards}
A.~Madry, A.~Makelov, L.~Schmidt, D.~Tsipras, and A.~Vladu, ``Towards deep
  learning models resistant to adversarial attacks,'' \emph{arXiv preprint
  arXiv:1706.06083}, 2017.

\bibitem{papernot2016limitations}
N.~Papernot, P.~McDaniel, S.~Jha, M.~Fredrikson, Z.~B. Celik, and A.~Swami,
  ``The limitations of deep learning in adversarial settings,'' in \emph{2016
  IEEE European symposium on security and privacy (EuroS\&P)}.\hskip 1em plus
  0.5em minus 0.4em\relax IEEE, 2016, pp. 372--387.

\bibitem{su2019one}
J.~Su, D.~V. Vargas, and K.~Sakurai, ``One pixel attack for fooling deep neural
  networks,'' \emph{IEEE Transactions on Evolutionary Computation}, vol.~23,
  no.~5, pp. 828--841, 2019.

\bibitem{simonyan2013deep}
K.~Simonyan, A.~Vedaldi, and A.~Zisserman, ``Deep inside convolutional
  networks: Visualising image classification models and saliency maps,''
  \emph{arXiv preprint arXiv:1312.6034}, 2013.

\bibitem{das2010differential}
S.~Das and P.~N. Suganthan, ``Differential evolution: A survey of the
  state-of-the-art,'' \emph{IEEE transactions on evolutionary computation},
  vol.~15, no.~1, pp. 4--31, 2010.

\bibitem{moosavi2016deepfool}
S.-M. Moosavi-Dezfooli, A.~Fawzi, and P.~Frossard, ``Deepfool: a simple and
  accurate method to fool deep neural networks,'' in \emph{Proceedings of the
  IEEE conference on computer vision and pattern recognition}, 2016, pp.
  2574--2582.

\bibitem{papernot2016distillation}
N.~Papernot, P.~McDaniel, X.~Wu, S.~Jha, and A.~Swami, ``Distillation as a
  defense to adversarial perturbations against deep neural networks,'' in
  \emph{2016 IEEE Symposium on Security and Privacy (SP)}.\hskip 1em plus 0.5em
  minus 0.4em\relax IEEE, 2016, pp. 582--597.

\bibitem{hinton2015distilling}
G.~Hinton, O.~Vinyals, and J.~Dean, ``Distilling the knowledge in a neural
  network,'' \emph{arXiv preprint arXiv:1503.02531}, 2015.

\bibitem{carlini2017towards}
N.~Carlini and D.~Wagner, ``Towards evaluating the robustness of neural
  networks,'' in \emph{2017 ieee symposium on security and privacy (sp)}.\hskip
  1em plus 0.5em minus 0.4em\relax IEEE, 2017, pp. 39--57.

\bibitem{carlini2017adversarial}
------, ``Adversarial examples are not easily detected: Bypassing ten detection
  methods,'' in \emph{Proceedings of the 10th ACM workshop on artificial
  intelligence and security}, 2017, pp. 3--14.

\bibitem{dong2020robust}
X.~Dong, J.~Han, D.~Chen, J.~Liu, H.~Bian, Z.~Ma, H.~Li, X.~Wang, W.~Zhang, and
  N.~Yu, ``Robust superpixel-guided attentional adversarial attack,'' in
  \emph{Proceedings of the IEEE/CVF Conference on Computer Vision and Pattern
  Recognition}, 2020, pp. 12\,895--12\,904.

\bibitem{guo2020backpropagating}
Y.~Guo, Q.~Li, and H.~Chen, ``Backpropagating linearly improves transferability
  of adversarial examples,'' \emph{NeurIPS}, 2020.

\bibitem{dong2020greedyfool}
X.~Dong, D.~Chen, J.~Bao, C.~Qin, L.~Yuan, W.~Zhang, N.~Yu, and D.~Chen,
  ``Greedyfool: Distortion-aware sparse adversarial attack,'' \emph{arXiv
  preprint arXiv:2010.13773}, 2020.

\bibitem{sriramanan2020guided}
G.~Sriramanan, S.~Addepalli, A.~Baburaj, and R.~V. Babu, ``Guided adversarial
  attack for evaluating and enhancing adversarial defenses,'' \emph{NeurIPS},
  2020.

\bibitem{tashiro2020diversity}
Y.~Tashiro, Y.~Song, and S.~Ermon, ``Diversity can be transferred: Output
  diversification for white-and black-box attacks,'' \emph{Advances in Neural
  Information Processing Systems}, vol.~33, 2020.

\bibitem{rony2019decoupling}
J.~Rony, L.~G. Hafemann, L.~S. Oliveira, I.~B. Ayed, R.~Sabourin, and
  E.~Granger, ``Decoupling direction and norm for efficient gradient-based l2
  adversarial attacks and defenses,'' in \emph{Proceedings of the IEEE/CVF
  Conference on Computer Vision and Pattern Recognition}, 2019, pp. 4322--4330.

\bibitem{yao2019trust}
Z.~Yao, A.~Gholami, P.~Xu, K.~Keutzer, and M.~W. Mahoney, ``Trust region based
  adversarial attack on neural networks,'' in \emph{Proceedings of the IEEE/CVF
  Conference on Computer Vision and Pattern Recognition}, 2019, pp.
  11\,350--11\,359.

\bibitem{conn2000trust}
A.~R. Conn, N.~I. Gould, and P.~L. Toint, \emph{Trust region methods}.\hskip
  1em plus 0.5em minus 0.4em\relax SIAM, 2000.

\bibitem{Phan_2021_CVPR}
B.~Phan, F.~Mannan, and F.~Heide, ``Adversarial imaging pipelines,'' in
  \emph{Proceedings of the IEEE/CVF Conference on Computer Vision and Pattern
  Recognition (CVPR)}, June 2021, pp. 16\,051--16\,061.

\bibitem{rahmati2020geoda}
A.~Rahmati, S.-M. Moosavi-Dezfooli, P.~Frossard, and H.~Dai, ``Geoda: a
  geometric framework for black-box adversarial attacks,'' in \emph{Proceedings
  of the IEEE/CVF Conference on Computer Vision and Pattern Recognition}, 2020,
  pp. 8446--8455.

\bibitem{brendel2017decision}
W.~Brendel, J.~Rauber, and M.~Bethge, ``Decision-based adversarial attacks:
  Reliable attacks against black-box machine learning models,'' \emph{ICLR},
  2018.

\bibitem{chen2020hopskipjumpattack}
J.~Chen, M.~I. Jordan, and M.~J. Wainwright, ``Hopskipjumpattack: A
  query-efficient decision-based attack,'' in \emph{2020 ieee symposium on
  security and privacy (sp)}.\hskip 1em plus 0.5em minus 0.4em\relax IEEE,
  2020, pp. 1277--1294.

\bibitem{liu2019geometry}
Y.~Liu, S.-M. Moosavi-Dezfooli, and P.~Frossard, ``A geometry-inspired
  decision-based attack,'' in \emph{Proceedings of the IEEE International
  Conference on Computer Vision}, 2019, pp. 4890--4898.

\bibitem{shi2020polishing}
Y.~Shi, Y.~Han, and Q.~Tian, ``Polishing decision-based adversarial noise with
  a customized sampling,'' in \emph{Proceedings of the IEEE/CVF Conference on
  Computer Vision and Pattern Recognition}, 2020, pp. 1030--1038.

\bibitem{du2019query}
J.~Du, H.~Zhang, J.~T. Zhou, Y.~Yang, and J.~Feng, ``Query-efficient meta
  attack to deep neural networks,'' \emph{ICLR}, 2020.

\bibitem{li2020projection}
J.~Li, R.~Ji, H.~Liu, J.~Liu, B.~Zhong, C.~Deng, and Q.~Tian, ``Projection \&
  probability-driven black-box attack,'' in \emph{Proceedings of the IEEE/CVF
  Conference on Computer Vision and Pattern Recognition}, 2020, pp. 362--371.

\bibitem{li2020qeba}
H.~Li, X.~Xu, X.~Zhang, S.~Yang, and B.~Li, ``Qeba: Query-efficient
  boundary-based blackbox attack,'' in \emph{Proceedings of the IEEE/CVF
  Conference on Computer Vision and Pattern Recognition}, 2020, pp. 1221--1230.

\bibitem{ru2019bayesopt}
B.~Ru, A.~Cobb, A.~Blaas, and Y.~Gal, ``Bayesopt adversarial attack,'' in
  \emph{International Conference on Learning Representations}, 2019.

\bibitem{akhtar2018hyperspectral}
N.~Akhtar and A.~Mian, ``Hyperspectral recovery from rgb images using gaussian
  processes,'' \emph{IEEE transactions on pattern analysis and machine
  intelligence}, vol.~42, no.~1, pp. 100--113, 2018.

\bibitem{cheng2019sign}
M.~Cheng, S.~Singh, P.~Chen, P.-Y. Chen, S.~Liu, and C.-J. Hsieh, ``Sign-opt: A
  query-efficient hard-label adversarial attack,'' \emph{ICLR}, 2020.

\bibitem{cheng2018query}
M.~Cheng, T.~Le, P.-Y. Chen, J.~Yi, H.~Zhang, and C.-J. Hsieh,
  ``Query-efficient hard-label black-box attack: An optimization-based
  approach,'' \emph{ICLR}, 2019.

\bibitem{cheng2019improving}
S.~Cheng, Y.~Dong, T.~Pang, H.~Su, and J.~Zhu, ``Improving black-box
  adversarial attacks with a transfer-based prior,'' \emph{arXiv preprint
  arXiv:1906.06919}, 2019.

\bibitem{huang2019black}
Z.~Huang and T.~Zhang, ``Black-box adversarial attack with transferable
  model-based embedding,'' \emph{ICLR}, 2020.

\bibitem{al2019sign}
A.~Al-Dujaili and U.-M. O'Reilly, ``Sign bits are all you need for black-box
  attacks,'' in \emph{International Conference on Learning Representations},
  2019.

\bibitem{ilyas2018prior}
A.~Ilyas, L.~Engstrom, and A.~Madry, ``Prior convictions: Black-box adversarial
  attacks with bandits and priors,'' \emph{ICLR}, 2019.

\bibitem{zhao2019design}
P.~Zhao, S.~Liu, P.-Y. Chen, N.~Hoang, K.~Xu, B.~Kailkhura, and X.~Lin, ``On
  the design of black-box adversarial examples by leveraging gradient-free
  optimization and operator splitting method,'' in \emph{Proceedings of the
  IEEE/CVF International Conference on Computer Vision}, 2019, pp. 121--130.

\bibitem{andriushchenko2020square}
M.~Andriushchenko, F.~Croce, N.~Flammarion, and M.~Hein, ``Square attack: a
  query-efficient black-box adversarial attack via random search,'' in
  \emph{European Conference on Computer Vision}.\hskip 1em plus 0.5em minus
  0.4em\relax Springer, 2020, pp. 484--501.

\bibitem{chen2020boosting}
W.~Chen, Z.~Zhang, X.~Hu, and B.~Wu, ``Boosting decision-based black-box
  adversarial attacks with random sign flip,'' in \emph{European Conference on
  Computer Vision}.\hskip 1em plus 0.5em minus 0.4em\relax Springer, 2020, pp.
  276--293.

\bibitem{xia2020improving}
S.-T. Xia and W.~Guo, ``Improving query efficiency of black-box adversarial
  attack,'' in \emph{European Conference on Computer Vision}.\hskip 1em plus
  0.5em minus 0.4em\relax Springer, 2020.

\bibitem{Wang_2021_CVPRDelving}
W.~Wang, B.~Yin, T.~Yao, L.~Zhang, Y.~Fu, S.~Ding, J.~Li, F.~Huang, and X.~Xue,
  ``Delving into data: Effectively substitute training for black-box attack,''
  in \emph{Proceedings of the IEEE/CVF Conference on Computer Vision and
  Pattern Recognition (CVPR)}, June 2021, pp. 4761--4770.

\bibitem{brunner2019guessing}
T.~Brunner, F.~Diehl, M.~T. Le, and A.~Knoll, ``Guessing smart: Biased sampling
  for efficient black-box adversarial attacks,'' in \emph{Proceedings of the
  IEEE/CVF International Conference on Computer Vision}, 2019, pp. 4958--4966.

\bibitem{tolias2019targeted}
G.~Tolias, F.~Radenovic, and O.~Chum, ``Targeted mismatch adversarial attack:
  Query with a flower to retrieve the tower,'' in \emph{Proceedings of the
  IEEE/CVF International Conference on Computer Vision}, 2019, pp. 5037--5046.

\bibitem{dolatabadi2020advflow}
H.~M. Dolatabadi, S.~Erfani, and C.~Leckie, ``Advflow: Inconspicuous black-box
  adversarial attacks using normalizing flows,'' \emph{NeruIPS}, 2020.

\bibitem{Maho_2021_CVPR}
T.~Maho, T.~Furon, and E.~Le~Merrer, ``Surfree: A fast surrogate-free black-box
  attack,'' in \emph{Proceedings of the IEEE/CVF Conference on Computer Vision
  and Pattern Recognition (CVPR)}, June 2021, pp. 10\,430--10\,439.

\bibitem{Li_2021_CVPRQAIR}
X.~Li, J.~Li, Y.~Chen, S.~Ye, Y.~He, S.~Wang, H.~Su, and H.~Xue, ``Qair:
  Practical query-efficient black-box attacks for image retrieval,'' in
  \emph{Proceedings of the IEEE/CVF Conference on Computer Vision and Pattern
  Recognition (CVPR)}, June 2021, pp. 3330--3339.

\bibitem{yang2020learning}
J.~Yang, Y.~Jiang, X.~Huang, B.~Ni, and C.~Zhao, ``Learning black-box attackers
  with transferable priors and query feedback,'' \emph{Advances in Neural
  Information Processing Systems}, vol.~33, 2020.

\bibitem{Ma_2021_CVPRSimulating}
C.~Ma, L.~Chen, and J.-H. Yong, ``Simulating unknown target models for
  query-efficient black-box attacks,'' in \emph{Proceedings of the IEEE/CVF
  Conference on Computer Vision and Pattern Recognition (CVPR)}, June 2021, pp.
  11\,835--11\,844.

\bibitem{wu2020boosting}
W.~Wu, Y.~Su, X.~Chen, S.~Zhao, I.~King, M.~R. Lyu, and Y.-W. Tai, ``Boosting
  the transferability of adversarial samples via attention,'' in
  \emph{Proceedings of the IEEE/CVF Conference on Computer Vision and Pattern
  Recognition}, 2020, pp. 1161--1170.

\bibitem{selvaraju2017grad}
R.~R. Selvaraju, M.~Cogswell, A.~Das, R.~Vedantam, D.~Parikh, and D.~Batra,
  ``Grad-cam: Visual explanations from deep networks via gradient-based
  localization,'' in \emph{Proceedings of the IEEE international conference on
  computer vision}, 2017, pp. 618--626.

\bibitem{li2020yet}
Q.~Li, Y.~Guo, and H.~Chen, ``Yet another intermediate-level attack,'' in
  \emph{European Conference on Computer Vision}.\hskip 1em plus 0.5em minus
  0.4em\relax Springer, 2020, pp. 241--257.

\bibitem{huang2019enhancing}
Q.~Huang, I.~Katsman, H.~He, Z.~Gu, S.~Belongie, and S.-N. Lim, ``Enhancing
  adversarial example transferability with an intermediate level attack,'' in
  \emph{Proceedings of the IEEE/CVF International Conference on Computer
  Vision}, 2019, pp. 4733--4742.

\bibitem{bose2020adversarial}
A.~J. Bose, G.~Gidel, H.~Berrard, A.~Cianflone, P.~Vincent, S.~Lacoste-Julien,
  and W.~L. Hamilton, ``Adversarial example games,'' \emph{arXiv preprint
  arXiv:2007.00720}, 2020.

\bibitem{li2020practical}
Q.~Li, Y.~Guo, and H.~Chen, ``Practical no-box adversarial attacks against
  dnns,'' \emph{NeurIPS}, 2020.

\bibitem{lin2019nesterov}
J.~Lin, C.~Song, K.~He, L.~Wang, and J.~E. Hopcroft, ``Nesterov accelerated
  gradient and scale invariance for adversarial attacks,'' in
  \emph{International Conference on Learning Representations}, 2019.

\bibitem{li2020towards}
M.~Li, C.~Deng, T.~Li, J.~Yan, X.~Gao, and H.~Huang, ``Towards transferable
  targeted attack,'' in \emph{Proceedings of the IEEE/CVF Conference on
  Computer Vision and Pattern Recognition}, 2020, pp. 641--649.

\bibitem{nesterov1983method}
Y.~Nesterov, ``A method for unconstrained convex minimization problem with the
  rate of convergence o (1/k\^{} 2),'' in \emph{Doklady an ussr}, vol. 269,
  1983, pp. 543--547.

\bibitem{lu2020enhancing}
Y.~Lu, Y.~Jia, J.~Wang, B.~Li, W.~Chai, L.~Carin, and S.~Velipasalar,
  ``Enhancing cross-task black-box transferability of adversarial examples with
  dispersion reduction,'' in \emph{Proceedings of the IEEE/CVF Conference on
  Computer Vision and Pattern Recognition}, 2020, pp. 940--949.

\bibitem{inkawhich2019feature}
N.~Inkawhich, W.~Wen, H.~H. Li, and Y.~Chen, ``Feature space perturbations
  yield more transferable adversarial examples,'' in \emph{Proceedings of the
  IEEE/CVF Conference on Computer Vision and Pattern Recognition}, 2019, pp.
  7066--7074.

\bibitem{inkawhich2020transferable}
N.~Inkawhich, K.~J. Liang, L.~Carin, and Y.~Chen, ``Transferable perturbations
  of deep feature distributions,'' \emph{ICLR}, 2020.

\bibitem{wang2020transferable}
H.~Wang, G.~Wang, Y.~Li, D.~Zhang, and L.~Lin, ``Transferable, controllable,
  and inconspicuous adversarial attacks on person re-identification with deep
  mis-ranking,'' in \emph{Proceedings of the IEEE/CVF Conference on Computer
  Vision and Pattern Recognition}, 2020, pp. 342--351.

\bibitem{li2019regional}
Y.~Li, S.~Bai, C.~Xie, Z.~Liao, X.~Shen, and A.~L. Yuille, ``Regional
  homogeneity: Towards learning transferable universal adversarial
  perturbations against defenses,'' \emph{ECCV}, 2020.

\bibitem{zhou2020dast}
M.~Zhou, J.~Wu, Y.~Liu, S.~Liu, and C.~Zhu, ``Dast: Data-free substitute
  training for adversarial attacks,'' in \emph{Proceedings of the IEEE/CVF
  Conference on Computer Vision and Pattern Recognition}, 2020, pp. 234--243.

\bibitem{zou2020improving}
J.~Zou, Z.~Pan, J.~Qiu, X.~Liu, T.~Rui, and W.~Li, ``Improving the
  transferability of adversarial examples with resized-diverse-inputs,
  diversity-ensemble and region fitting,'' in \emph{European Conference on
  Computer Vision}.\hskip 1em plus 0.5em minus 0.4em\relax Springer, 2020, pp.
  563--579.

\bibitem{dong2019evading}
Y.~Dong, T.~Pang, H.~Su, and J.~Zhu, ``Evading defenses to transferable
  adversarial examples by translation-invariant attacks,'' in \emph{Proceedings
  of the IEEE/CVF Conference on Computer Vision and Pattern Recognition}, 2019,
  pp. 4312--4321.

\bibitem{shi2019curls}
Y.~Shi, S.~Wang, and Y.~Han, ``Curls \& whey: Boosting black-box adversarial
  attacks,'' in \emph{Proceedings of the IEEE/CVF Conference on Computer Vision
  and Pattern Recognition}, 2019, pp. 6519--6527.

\bibitem{Wang_2021_CVPR}
X.~Wang and K.~He, ``Enhancing the transferability of adversarial attacks
  through variance tuning,'' in \emph{Proceedings of the IEEE/CVF Conference on
  Computer Vision and Pattern Recognition (CVPR)}, June 2021, pp. 1924--1933.

\bibitem{Wu_2021_CVPRImproving}
W.~Wu, Y.~Su, M.~R. Lyu, and I.~King, ``Improving the transferability of
  adversarial samples with adversarial transformations,'' in \emph{Proceedings
  of the IEEE/CVF Conference on Computer Vision and Pattern Recognition
  (CVPR)}, June 2021, pp. 9024--9033.

\bibitem{sharif2018suitability}
M.~Sharif, L.~Bauer, and M.~K. Reiter, ``On the suitability of lp-norms for
  creating and preventing adversarial examples,'' in \emph{Proceedings of the
  IEEE Conference on Computer Vision and Pattern Recognition Workshops}, 2018,
  pp. 1605--1613.

\bibitem{hosseini2018semantic}
H.~Hosseini and R.~Poovendran, ``Semantic adversarial examples,'' in
  \emph{Proceedings of the IEEE Conference on Computer Vision and Pattern
  Recognition Workshops}, 2018, pp. 1614--1619.

\bibitem{eykholt2018robust}
K.~Eykholt, I.~Evtimov, E.~Fernandes, B.~Li, A.~Rahmati, C.~Xiao, A.~Prakash,
  T.~Kohno, and D.~Song, ``Robust physical-world attacks on deep learning
  visual classification,'' in \emph{Proceedings of the IEEE Conference on
  Computer Vision and Pattern Recognition}, 2018, pp. 1625--1634.

\bibitem{joshi2019semantic}
A.~Joshi, A.~Mukherjee, S.~Sarkar, and C.~Hegde, ``Semantic adversarial
  attacks: Parametric transformations that fool deep classifiers,'' in
  \emph{Proceedings of the IEEE International Conference on Computer Vision},
  2019, pp. 4773--4783.

\bibitem{sharif2019general}
M.~Sharif, S.~Bhagavatula, L.~Bauer, and M.~K. Reiter, ``A general framework
  for adversarial examples with objectives,'' \emph{ACM Transactions on Privacy
  and Security (TOPS)}, vol.~22, no.~3, pp. 1--30, 2019.

\bibitem{croce2019sparse}
F.~Croce and M.~Hein, ``Sparse and imperceivable adversarial attacks,'' in
  \emph{Proceedings of the IEEE International Conference on Computer Vision},
  2019, pp. 4724--4732.

\bibitem{wong2019wasserstein}
E.~Wong, F.~R. Schmidt, and J.~Z. Kolter, ``Wasserstein adversarial examples
  via projected sinkhorn iterations,'' \emph{arXiv preprint arXiv:1902.07906},
  2019.

\bibitem{bhattad2019unrestricted}
A.~Bhattad, M.~J. Chong, K.~Liang, B.~Li, and D.~A. Forsyth, ``Unrestricted
  adversarial examples via semantic manipulation,'' \emph{ICLR}, 2020.

\bibitem{shamsabadi2020colorfool}
A.~S. Shamsabadi, R.~Sanchez-Matilla, and A.~Cavallaro, ``Colorfool: Semantic
  adversarial colorization,'' in \emph{Proceedings of the IEEE/CVF Conference
  on Computer Vision and Pattern Recognition}, 2020, pp. 1151--1160.

\bibitem{ruderman1998statistics}
D.~L. Ruderman, T.~W. Cronin, and C.-C. Chiao, ``Statistics of cone responses
  to natural images: implications for visual coding,'' \emph{JOSA A}, vol.~15,
  no.~8, pp. 2036--2045, 1998.

\bibitem{zhao2020towards}
Z.~Zhao, Z.~Liu, and M.~Larson, ``Towards large yet imperceptible adversarial
  image perturbations with perceptual color distance,'' in \emph{Proceedings of
  the IEEE/CVF Conference on Computer Vision and Pattern Recognition}, 2020,
  pp. 1039--1048.

\bibitem{luo2001development}
M.~R. Luo, G.~Cui, and B.~Rigg, ``The development of the cie 2000
  colour-difference formula: Ciede2000,'' \emph{Color Research \& Application:
  Endorsed by Inter-Society Color Council, The Colour Group (Great Britain),
  Canadian Society for Color, Color Science Association of Japan, Dutch Society
  for the Study of Color, The Swedish Colour Centre Foundation, Colour Society
  of Australia, Centre Fran{\c{c}}ais de la Couleur}, vol.~26, no.~5, pp.
  340--350, 2001.

\bibitem{qiu2020semanticadv}
H.~Qiu, C.~Xiao, L.~Yang, X.~Yan, H.~Lee, and B.~Li, ``Semanticadv: Generating
  adversarial examples via attribute-conditioned image editing,'' in
  \emph{European Conference on Computer Vision}.\hskip 1em plus 0.5em minus
  0.4em\relax Springer, 2020, pp. 19--37.

\bibitem{korshunov2018deepfakes}
P.~Korshunov and S.~Marcel, ``Deepfakes: a new threat to face recognition?
  assessment and detection,'' \emph{arXiv preprint arXiv:1812.08685}, 2018.

\bibitem{Chen_2021_CVPRMagDR}
Z.~Chen, L.~Xie, S.~Pang, Y.~He, and B.~Zhang, ``Magdr: Mask-guided detection
  and reconstruction for defending deepfakes,'' in \emph{Proceedings of the
  IEEE/CVF Conference on Computer Vision and Pattern Recognition (CVPR)}, June
  2021, pp. 9014--9023.

\bibitem{Hendrycks_2021_CVPR}
D.~Hendrycks, K.~Zhao, S.~Basart, J.~Steinhardt, and D.~Song, ``Natural
  adversarial examples,'' in \emph{Proceedings of the IEEE/CVF Conference on
  Computer Vision and Pattern Recognition (CVPR)}, June 2021, pp.
  15\,262--15\,271.

\bibitem{li2020backdoor}
Y.~Li, B.~Wu, Y.~Jiang, Z.~Li, and S.-T. Xia, ``Backdoor learning: A survey,''
  \emph{arXiv preprint arXiv:2007.08745}, 2020.

\bibitem{liu2020survey}
Y.~Liu, A.~Mondal, A.~Chakraborty, M.~Zuzak, N.~Jacobsen, D.~Xing, and
  A.~Srivastava, ``A survey on neural trojans,'' in \emph{2020 21st
  International Symposium on Quality Electronic Design (ISQED)}.\hskip 1em plus
  0.5em minus 0.4em\relax IEEE, 2020, pp. 33--39.

\bibitem{liu2020reflection}
Y.~Liu, X.~Ma, J.~Bailey, and F.~Lu, ``Reflection backdoor: A natural backdoor
  attack on deep neural networks,'' in \emph{European Conference on Computer
  Vision}.\hskip 1em plus 0.5em minus 0.4em\relax Springer, 2020, pp. 182--199.

\bibitem{nguyen2020input}
A.~Nguyen and A.~Tran, ``Input-aware dynamic backdoor attack,'' \emph{arXiv
  preprint arXiv:2010.08138}, 2020.

\bibitem{xie2019dba}
C.~Xie, K.~Huang, P.-Y. Chen, and B.~Li, ``Dba: Distributed backdoor attacks
  against federated learning,'' in \emph{International Conference on Learning
  Representations}, 2019.

\bibitem{smith2017federated}
V.~Smith, C.-K. Chiang, M.~Sanjabi, and A.~S. Talwalkar, ``Federated multi-task
  learning,'' in \emph{Advances in neural information processing systems},
  2017, pp. 4424--4434.

\bibitem{bhagoji2019analyzing}
A.~N. Bhagoji, S.~Chakraborty, P.~Mittal, and S.~Calo, ``Analyzing federated
  learning through an adversarial lens,'' in \emph{International Conference on
  Machine Learning}.\hskip 1em plus 0.5em minus 0.4em\relax PMLR, 2019, pp.
  634--643.

\bibitem{bagdasaryan2020backdoor}
E.~Bagdasaryan, A.~Veit, Y.~Hua, D.~Estrin, and V.~Shmatikov, ``How to backdoor
  federated learning,'' in \emph{International Conference on Artificial
  Intelligence and Statistics}.\hskip 1em plus 0.5em minus 0.4em\relax PMLR,
  2020, pp. 2938--2948.

\bibitem{guo2020practical}
J.~Guo and C.~Liu, ``Practical poisoning attacks on neural networks,'' in
  \emph{European Conference on Computer Vision}.\hskip 1em plus 0.5em minus
  0.4em\relax Springer, 2020, pp. 142--158.

\bibitem{rakin2020tbt}
A.~S. Rakin, Z.~He, and D.~Fan, ``Tbt: Targeted neural network attack with bit
  trojan,'' in \emph{Proceedings of the IEEE/CVF Conference on Computer Vision
  and Pattern Recognition}, 2020, pp. 13\,198--13\,207.

\bibitem{zhao2020clean}
S.~Zhao, X.~Ma, X.~Zheng, J.~Bailey, J.~Chen, and Y.-G. Jiang, ``Clean-label
  backdoor attacks on video recognition models,'' in \emph{Proceedings of the
  IEEE/CVF Conference on Computer Vision and Pattern Recognition}, 2020, pp.
  14\,443--14\,452.

\bibitem{kolouri2020universal}
S.~Kolouri, A.~Saha, H.~Pirsiavash, and H.~Hoffmann, ``Universal litmus
  patterns: Revealing backdoor attacks in cnns,'' in \emph{Proceedings of the
  IEEE/CVF Conference on Computer Vision and Pattern Recognition}, 2020, pp.
  301--310.

\bibitem{wang2020attack}
H.~Wang, K.~Sreenivasan, S.~Rajput, H.~Vishwakarma, S.~Agarwal, J.-y. Sohn,
  K.~Lee, and D.~Papailiopoulos, ``Attack of the tails: Yes, you really can
  backdoor federated learning,'' \emph{arXiv preprint arXiv:2007.05084}, 2020.

\bibitem{fredrikson2014privacy}
M.~Fredrikson, E.~Lantz, S.~Jha, S.~Lin, D.~Page, and T.~Ristenpart, ``Privacy
  in pharmacogenetics: An end-to-end case study of personalized warfarin
  dosing,'' in \emph{23rd $\{$USENIX$\}$ Security Symposium ($\{$USENIX$\}$
  Security 14)}, 2014, pp. 17--32.

\bibitem{wu2016methodology}
X.~Wu, M.~Fredrikson, S.~Jha, and J.~F. Naughton, ``A methodology for
  formalizing model-inversion attacks,'' in \emph{2016 IEEE 29th Computer
  Security Foundations Symposium (CSF)}.\hskip 1em plus 0.5em minus 0.4em\relax
  IEEE, 2016, pp. 355--370.

\bibitem{yeom2018privacy}
S.~Yeom, I.~Giacomelli, M.~Fredrikson, and S.~Jha, ``Privacy risk in machine
  learning: Analyzing the connection to overfitting,'' in \emph{2018 IEEE 31st
  Computer Security Foundations Symposium (CSF)}.\hskip 1em plus 0.5em minus
  0.4em\relax IEEE, 2018, pp. 268--282.

\bibitem{nguyen2016synthesizing}
A.~Nguyen, A.~Dosovitskiy, J.~Yosinski, T.~Brox, and J.~Clune, ``Synthesizing
  the preferred inputs for neurons in neural networks via deep generator
  networks,'' \emph{Advances in neural information processing systems},
  vol.~29, pp. 3387--3395, 2016.

\bibitem{akhtar2021attack}
N.~Akhtar, M.~Jalwana, M.~Bennamoun, and A.~S. Mian, ``Attack to fool and
  explain deep networks,'' \emph{IEEE Transactions on Pattern Analysis and
  Machine Intelligence}, 2021.

\bibitem{zhang2020secret}
Y.~Zhang, R.~Jia, H.~Pei, W.~Wang, B.~Li, and D.~Song, ``The secret revealer:
  generative model-inversion attacks against deep neural networks,'' in
  \emph{Proceedings of the IEEE/CVF Conference on Computer Vision and Pattern
  Recognition}, 2020, pp. 253--261.

\bibitem{athalye2018obfuscated}
A.~Athalye, N.~Carlini, and D.~Wagner, ``Obfuscated gradients give a false
  sense of security: Circumventing defenses to adversarial examples,''
  \emph{ICML}, 2018.

\bibitem{tramer2020adaptive}
F.~Tramer, N.~Carlini, W.~Brendel, and A.~Madry, ``On adaptive attacks to
  adversarial example defenses,'' \emph{NeurIPS}, 2020.

\bibitem{xiao2019enhancing}
C.~Xiao, P.~Zhong, and C.~Zheng, ``Enhancing adversarial defense by
  k-winners-take-all,'' \emph{arXiv preprint arXiv:1905.10510}, 2020.

\bibitem{roth2019odds}
K.~Roth, Y.~Kilcher, and T.~Hofmann, ``The odds are odd: A statistical test for
  detecting adversarial examples,'' in \emph{International Conference on
  Machine Learning}.\hskip 1em plus 0.5em minus 0.4em\relax PMLR, 2019, pp.
  5498--5507.

\bibitem{li2019generative}
Y.~Li, J.~Bradshaw, and Y.~Sharma, ``Are generative classifiers more robust to
  adversarial attacks?'' in \emph{International Conference on Machine
  Learning}.\hskip 1em plus 0.5em minus 0.4em\relax PMLR, 2019, pp. 3804--3814.

\bibitem{bafna2018thwarting}
M.~Bafna, J.~Murtagh, and N.~Vyas, ``Thwarting adversarial examples: An $ l\_0
  $-robustsparse fourier transform,'' \emph{NeurIPS}, 2018.

\bibitem{pang2019rethinking}
T.~Pang, K.~Xu, Y.~Dong, C.~Du, N.~Chen, and J.~Zhu, ``Rethinking softmax
  cross-entropy loss for adversarial robustness,'' \emph{ICLR}, 2020.

\bibitem{verma2019error}
G.~Verma and A.~Swami, ``Error correcting output codes improve probability
  estimation and adversarial robustness of deep neural networks,''
  \emph{Advances in Neural Information Processing Systems}, vol.~32, pp.
  8646--8656, 2019.

\bibitem{pang2019improving}
T.~Pang, K.~Xu, C.~Du, N.~Chen, and J.~Zhu, ``Improving adversarial robustness
  via promoting ensemble diversity,'' in \emph{International Conference on
  Machine Learning}.\hskip 1em plus 0.5em minus 0.4em\relax PMLR, 2019, pp.
  4970--4979.

\bibitem{sen2020empir}
S.~Sen, B.~Ravindran, and A.~Raghunathan, ``Empir: Ensembles of mixed precision
  deep networks for increased robustness against adversarial attacks,''
  \emph{ICLR}, 2020.

\bibitem{yang2018characterizing}
Z.~Yang, B.~Li, P.-Y. Chen, and D.~Song, ``Characterizing audio adversarial
  examples using temporal dependency,'' \emph{ICLR}, 2019.

\bibitem{pang2019mixup}
T.~Pang, K.~Xu, and J.~Zhu, ``Mixup inference: Better exploiting mixup to
  defend adversarial attacks,'' \emph{ICLR}, 2020.

\bibitem{yang2019me}
Y.~Yang, G.~Zhang, D.~Katabi, and Z.~Xu, ``Me-net: Towards effective
  adversarial robustness with matrix estimation,'' \emph{ICLR}, 2019.

\bibitem{yin2019adversarial}
X.~Yin, S.~Kolouri, and G.~K. Rohde, ``Adversarial example detection and
  classification with asymmetrical adversarial training,'' \emph{ICLR}, 2020.

\bibitem{yu2019new}
T.~Yu, S.~Hu, C.~Guo, W.-L. Chao, and K.~Q. Weinberger, ``A new defense against
  adversarial images: Turning a weakness into a strength,'' \emph{NeurIPS},
  2019.

\bibitem{ghiasi2020breaking}
A.~Ghiasi, A.~Shafahi, and T.~Goldstein, ``Breaking certified defenses:
  Semantic adversarial examples with spoofed robustness certificates,''
  \emph{ICLR}, 2020.

\bibitem{zhou2020adversarial}
M.~Zhou, Z.~Niu, L.~Wang, Q.~Zhang, and G.~Hua, ``Adversarial ranking attack
  and defense,'' \emph{ECCV}, 2020.

\bibitem{razavi2016flip}
K.~Razavi, B.~Gras, E.~Bosman, B.~Preneel, C.~Giuffrida, and H.~Bos, ``Flip
  feng shui: Hammering a needle in the software stack,'' in \emph{25th
  $\{$USENIX$\}$ Security Symposium ($\{$USENIX$\}$ Security 16)}, 2016, pp.
  1--18.

\bibitem{hong2019terminal}
S.~Hong, P.~Frigo, Y.~Kaya, C.~Giuffrida, and T.~Dumitraș, ``Terminal brain
  damage: Exposing the graceless degradation in deep neural networks under
  hardware fault attacks,'' in \emph{28th $\{$USENIX$\}$ Security Symposium
  ($\{$USENIX$\}$ Security 19)}, 2019, pp. 497--514.

\bibitem{rakin2019bit}
A.~S. Rakin, Z.~He, and D.~Fan, ``Bit-flip attack: Crushing neural network with
  progressive bit search,'' in \emph{Proceedings of the IEEE International
  Conference on Computer Vision}, 2019, pp. 1211--1220.

\bibitem{rezaei2019target}
S.~Rezaei and X.~Liu, ``A target-agnostic attack on deep models: Exploiting
  security vulnerabilities of transfer learning,'' \emph{ICLR}, 2020.

\bibitem{mor2019optimal}
R.~Mor, E.~Peterfreund, M.~Gavish, and A.~Globerson, ``Optimal strategies
  against generative attacks,'' in \emph{International Conference on Learning
  Representations}, 2019.

\bibitem{ganeshan2019fda}
A.~Ganeshan and R.~V. Babu, ``Fda: Feature disruptive attack,'' in
  \emph{Proceedings of the IEEE/CVF International Conference on Computer
  Vision}, 2019, pp. 8069--8079.

\bibitem{zhou2020lg}
H.~Zhou, D.~Chen, J.~Liao, K.~Chen, X.~Dong, K.~Liu, W.~Zhang, G.~Hua, and
  N.~Yu, ``Lg-gan: Label guided adversarial network for flexible targeted
  attack of point cloud based deep networks,'' in \emph{Proceedings of the
  IEEE/CVF Conference on Computer Vision and Pattern Recognition}, 2020, pp.
  10\,356--10\,365.

\bibitem{zhao2020isometry}
Y.~Zhao, Y.~Wu, C.~Chen, and A.~Lim, ``On isometry robustness of deep 3d point
  cloud models under adversarial attacks,'' in \emph{Proceedings of the
  IEEE/CVF Conference on Computer Vision and Pattern Recognition}, 2020, pp.
  1201--1210.

\bibitem{hamdi2020advpc}
A.~Hamdi, S.~Rojas, A.~Thabet, and B.~Ghanem, ``Advpc: Transferable adversarial
  perturbations on 3d point clouds,'' in \emph{European Conference on Computer
  Vision}.\hskip 1em plus 0.5em minus 0.4em\relax Springer, 2020, pp. 241--257.

\bibitem{zhou2019dup}
H.~Zhou, K.~Chen, W.~Zhang, H.~Fang, W.~Zhou, and N.~Yu, ``Dup-net: Denoiser
  and upsampler network for 3d adversarial point clouds defense,'' in
  \emph{Proceedings of the IEEE/CVF International Conference on Computer
  Vision}, 2019, pp. 1961--1970.

\bibitem{wicker2019robustness}
M.~Wicker and M.~Kwiatkowska, ``Robustness of 3d deep learning in an
  adversarial setting,'' in \emph{Proceedings of the IEEE Conference on
  Computer Vision and Pattern Recognition}, 2019, pp. 11\,767--11\,775.

\bibitem{xiang2019generating}
C.~Xiang, C.~R. Qi, and B.~Li, ``Generating 3d adversarial point clouds,'' in
  \emph{Proceedings of the IEEE Conference on Computer Vision and Pattern
  Recognition}, 2019, pp. 9136--9144.

\bibitem{zeng2019adversarial}
X.~Zeng, C.~Liu, Y.-S. Wang, W.~Qiu, L.~Xie, Y.-W. Tai, C.-K. Tang, and A.~L.
  Yuille, ``Adversarial attacks beyond the image space,'' in \emph{Proceedings
  of the IEEE Conference on Computer Vision and Pattern Recognition}, 2019, pp.
  4302--4311.

\bibitem{zhang2020motion}
H.~Zhang, L.~Zhu, Y.~Zhu, and Y.~Yang, ``Motion-excited sampler: Video
  adversarial attack with sparked prior,'' \emph{ECCV}, 2020.

\bibitem{liu2020spatiotemporal}
A.~Liu, T.~Huang, X.~Liu, Y.~Xu, Y.~Ma, X.~Chen, S.~J. Maybank, and D.~Tao,
  ``Spatiotemporal attacks for embodied agents,'' in \emph{European Conference
  on Computer Vision}.\hskip 1em plus 0.5em minus 0.4em\relax Springer, 2020,
  pp. 122--138.

\bibitem{liu2020adversarial}
J.~Liu, N.~Akhtar, and A.~Mian, ``Adversarial attack on skeleton-based human
  action recognition,'' \emph{IEEE Transactions on Neural Networks and Learning
  Systems}, 2020.

\bibitem{Diao_2021_CVPRBASAR}
Y.~Diao, T.~Shao, Y.-L. Yang, K.~Zhou, and H.~Wang, ``Basar:black-box attack on
  skeletal action recognition,'' in \emph{Proceedings of the IEEE/CVF
  Conference on Computer Vision and Pattern Recognition (CVPR)}, June 2021, pp.
  7597--7607.

\bibitem{Wang_2021_CVPRUnderstanding}
H.~Wang, F.~He, Z.~Peng, T.~Shao, Y.-L. Yang, K.~Zhou, and D.~Hogg,
  ``Understanding the robustness of skeleton-based action recognition under
  adversarial attack,'' in \emph{Proceedings of the IEEE/CVF Conference on
  Computer Vision and Pattern Recognition (CVPR)}, June 2021, pp.
  14\,656--14\,665.

\bibitem{Pony_2021_CVPR}
R.~Pony, I.~Naeh, and S.~Mannor, ``Over-the-air adversarial flickering attacks
  against video recognition networks,'' in \emph{Proceedings of the IEEE/CVF
  Conference on Computer Vision and Pattern Recognition (CVPR)}, June 2021, pp.
  515--524.

\bibitem{xiao2019advit}
C.~Xiao, R.~Deng, B.~Li, T.~Lee, B.~Edwards, J.~Yi, D.~Song, M.~Liu, and
  I.~Molloy, ``Advit: Adversarial frames identifier based on temporal
  consistency in videos,'' in \emph{Proceedings of the IEEE/CVF International
  Conference on Computer Vision}, 2019, pp. 3968--3977.

\bibitem{zugner2019adversarial}
D.~Z{\"u}gner and S.~G{\"u}nnemann, ``Adversarial attacks on graph neural
  networks via meta learning,'' \emph{ICLR}, 2019.

\bibitem{jin2020certified}
H.~Jin, Z.~Shi, V.~J. S.~A. Peruri, and X.~Zhang, ``Certified robustness of
  graph convolution networks for graph classification under topological
  attacks,'' \emph{Advances in Neural Information Processing Systems}, vol.~33,
  2020.

\bibitem{khalil2018combinatorial}
E.~B. Khalil, A.~Gupta, and B.~Dilkina, ``Combinatorial attacks on binarized
  neural networks,'' \emph{arXiv preprint arXiv:1810.03538}, 2018.

\bibitem{lin2019defensive}
J.~Lin, C.~Gan, and S.~Han, ``Defensive quantization: When efficiency meets
  robustness,'' \emph{arXiv preprint arXiv:1904.08444}, 2019.

\bibitem{alaifari2018adef}
R.~Alaifari, G.~S. Alberti, and T.~Gauksson, ``Adef: an iterative algorithm to
  construct adversarial deformations,'' \emph{arXiv preprint arXiv:1804.07729},
  2018.

\bibitem{fan2020sparse}
Y.~Fan, B.~Wu, T.~Li, Y.~Zhang, M.~Li, Z.~Li, and Y.~Yang, ``Sparse adversarial
  attack via perturbation factorization,'' in \emph{Proceedings of European
  Conference on Computer Vision}, 2020.

\bibitem{liu2019universal}
H.~Liu, R.~Ji, J.~Li, B.~Zhang, Y.~Gao, Y.~Wu, and F.~Huang, ``Universal
  adversarial perturbation via prior driven uncertainty approximation,'' in
  \emph{Proceedings of the IEEE/CVF International Conference on Computer
  Vision}, 2019, pp. 2941--2949.

\bibitem{din2020steganographic}
S.~U. Din, N.~Akhtar, S.~Younis, F.~Shafait, A.~Mansoor, and M.~Shafique,
  ``Steganographic universal adversarial perturbations,'' \emph{Pattern
  Recognition Letters}, vol. 135, pp. 146--152, 2020.

\bibitem{Rampini_2021_CVPR}
A.~Rampini, F.~Pestarini, L.~Cosmo, S.~Melzi, and E.~Rodola, ``Universal
  spectral adversarial attacks for deformable shapes,'' in \emph{Proceedings of
  the IEEE/CVF Conference on Computer Vision and Pattern Recognition (CVPR)},
  June 2021, pp. 3216--3226.

\bibitem{zhong2018perception}
Z.~Zhong, W.~Xu, Y.~Jia, and T.~Wei, ``Perception deception: Physical
  adversarial attack challenges and tactics for dnn-based object detection,''
  \emph{Black Hat Europe}, 2018.

\bibitem{redmon2017yolo9000}
J.~Redmon and A.~Farhadi, ``Yolo9000: better, faster, stronger,'' in
  \emph{Proceedings of the IEEE conference on computer vision and pattern
  recognition}, 2017, pp. 7263--7271.

\bibitem{sun2019deep}
S.~Sun, N.~Akhtar, H.~Song, A.~Mian, and M.~Shah, ``Deep affinity network for
  multiple object tracking,'' \emph{IEEE transactions on pattern analysis and
  machine intelligence}, vol.~43, no.~1, pp. 104--119, 2019.

\bibitem{Chen_2021_CVPR}
P.-C. Chen, B.-H. Kung, and J.-C. Chen, ``Class-aware robust adversarial
  training for object detection,'' in \emph{Proceedings of the IEEE/CVF
  Conference on Computer Vision and Pattern Recognition (CVPR)}, June 2021, pp.
  10\,420--10\,429.

\bibitem{yan2020cooling}
B.~Yan, D.~Wang, H.~Lu, and X.~Yang, ``Cooling-shrinking attack: Blinding the
  tracker with imperceptible noises,'' in \emph{Proceedings of the IEEE/CVF
  Conference on Computer Vision and Pattern Recognition}, 2020, pp. 990--999.

\bibitem{li2019siamrpn++}
B.~Li, W.~Wu, Q.~Wang, F.~Zhang, J.~Xing, and J.~Yan, ``Siamrpn++: Evolution of
  siamese visual tracking with very deep networks,'' in \emph{Proceedings of
  the IEEE Conference on Computer Vision and Pattern Recognition}, 2019, pp.
  4282--4291.

\bibitem{liang2020efficient}
S.~Liang, X.~Wei, S.~Yao, and X.~Cao, ``Efficient adversarial attacks for
  visual object tracking,'' in \emph{European Conference on Computer
  Vision}.\hskip 1em plus 0.5em minus 0.4em\relax Springer, 2020, pp. 34--50.

\bibitem{wiyatno2019physical}
R.~R. Wiyatno and A.~Xu, ``Physical adversarial textures that fool visual
  object tracking,'' in \emph{Proceedings of the IEEE/CVF International
  Conference on Computer Vision}, 2019, pp. 4822--4831.

\bibitem{huang2020universal}
L.~Huang, C.~Gao, Y.~Zhou, C.~Xie, A.~L. Yuille, C.~Zou, and N.~Liu,
  ``Universal physical camouflage attacks on object detectors,'' in
  \emph{Proceedings of the IEEE/CVF Conference on Computer Vision and Pattern
  Recognition}, 2020, pp. 720--729.

\bibitem{wu2020making}
Z.~Wu, S.-N. Lim, L.~S. Davis, and T.~Goldstein, ``Making an invisibility
  cloak: Real world adversarial attacks on object detectors,'' in
  \emph{European Conference on Computer Vision}.\hskip 1em plus 0.5em minus
  0.4em\relax Springer, 2020, pp. 1--17.

\bibitem{Zolfi_2021_CVPRPatch}
A.~Zolfi, M.~Kravchik, Y.~Elovici, and A.~Shabtai, ``The translucent patch: A
  physical and universal attack on object detectors,'' in \emph{Proceedings of
  the IEEE/CVF Conference on Computer Vision and Pattern Recognition (CVPR)},
  June 2021, pp. 15\,232--15\,241.

\bibitem{guo2020spark}
Q.~Guo, X.~Xie, F.~Juefei-Xu, L.~Ma, Z.~Li, W.~Xue, W.~Feng, and Y.~Liu,
  ``Spark: Spatial-aware online incremental attack against visual tracking,''
  in \emph{Proceedings of the European Conference on Computer Vision (ECCV)},
  vol.~2.\hskip 1em plus 0.5em minus 0.4em\relax Springer, 2020.

\bibitem{Jia_2021_CVPR}
S.~Jia, Y.~Song, C.~Ma, and X.~Yang, ``Iou attack: Towards temporally coherent
  black-box adversarial attack for visual object tracking,'' in
  \emph{Proceedings of the IEEE/CVF Conference on Computer Vision and Pattern
  Recognition (CVPR)}, June 2021, pp. 6709--6718.

\bibitem{huang2017adversarial}
S.~Huang, N.~Papernot, I.~Goodfellow, Y.~Duan, and P.~Abbeel, ``Adversarial
  attacks on neural network policies,'' \emph{arXiv preprint arXiv:1702.02284},
  2017.

\bibitem{xiang2018pca}
Y.~Xiang, W.~Niu, J.~Liu, T.~Chen, and Z.~Han, ``A pca-based model to predict
  adversarial examples on q-learning of path finding,'' in \emph{2018 IEEE
  Third International Conference on Data Science in Cyberspace (DSC)}.\hskip
  1em plus 0.5em minus 0.4em\relax IEEE, 2018, pp. 773--780.

\bibitem{bai2018adversarial}
X.~Bai, W.~Niu, J.~Liu, X.~Gao, Y.~Xiang, and J.~Liu, ``Adversarial examples
  construction towards white-box q table variation in dqn pathfinding
  training,'' in \emph{2018 IEEE Third International Conference on Data Science
  in Cyberspace (DSC)}.\hskip 1em plus 0.5em minus 0.4em\relax IEEE, 2018, pp.
  781--787.

\bibitem{mnih2013playing}
V.~Mnih, K.~Kavukcuoglu, D.~Silver, A.~Graves, I.~Antonoglou, D.~Wierstra, and
  M.~Riedmiller, ``Playing atari with deep reinforcement learning,''
  \emph{arXiv preprint arXiv:1312.5602}, 2013.

\bibitem{chen2018gradient}
T.~Chen, W.~Niu, Y.~Xiang, X.~Bai, J.~Liu, Z.~Han, and G.~Li, ``Gradient
  band-based adversarial training for generalized attack immunity of a3c path
  finding,'' \emph{arXiv preprint arXiv:1807.06752}, 2018.

\bibitem{behzadan2017vulnerability}
V.~Behzadan and A.~Munir, ``Vulnerability of deep reinforcement learning to
  policy induction attacks,'' in \emph{International Conference on Machine
  Learning and Data Mining in Pattern Recognition}.\hskip 1em plus 0.5em minus
  0.4em\relax Springer, 2017, pp. 262--275.

\bibitem{lin2017tactics}
Y.-C. Lin, Z.-W. Hong, Y.-H. Liao, M.-L. Shih, M.-Y. Liu, and M.~Sun, ``Tactics
  of adversarial attack on deep reinforcement learning agents,'' \emph{arXiv
  preprint arXiv:1703.06748}, 2017.

\bibitem{liu2017method}
J.~Liu, W.~Niu, J.~Liu, J.~Zhao, T.~Chen, Y.~Yang, Y.~Xiang, and L.~Han, ``A
  method to effectively detect vulnerabilities on path planning of vin,'' in
  \emph{International Conference on Information and Communications
  Security}.\hskip 1em plus 0.5em minus 0.4em\relax Springer, 2017, pp.
  374--384.

\bibitem{chen2019adversarial}
T.~Chen, J.~Liu, Y.~Xiang, W.~Niu, E.~Tong, and Z.~Han, ``Adversarial attack
  and defense in reinforcement learning-from ai security view,''
  \emph{Cybersecurity}, vol.~2, no.~1, pp. 1--22, 2019.

\bibitem{gleave2019adversarial}
A.~Gleave, M.~Dennis, C.~Wild, N.~Kant, S.~Levine, and S.~Russell,
  ``Adversarial policies: Attacking deep reinforcement learning,'' \emph{ICLR},
  2020.

\bibitem{rakhsha2020policy}
A.~Rakhsha, G.~Radanovic, R.~Devidze, X.~Zhu, and A.~Singla, ``Policy teaching
  via environment poisoning: Training-time adversarial attacks against
  reinforcement learning,'' in \emph{International Conference on Machine
  Learning}.\hskip 1em plus 0.5em minus 0.4em\relax PMLR, 2020, pp. 7974--7984.

\bibitem{zhang2020adaptive}
X.~Zhang, Y.~Ma, A.~Singla, and X.~Zhu, ``Adaptive reward-poisoning attacks
  against reinforcement learning,'' in \emph{International Conference on
  Machine Learning}.\hskip 1em plus 0.5em minus 0.4em\relax PMLR, 2020, pp.
  11\,225--11\,234.

\bibitem{zhang2020robust}
H.~Zhang, H.~Chen, C.~Xiao, B.~Li, M.~Liu, D.~Boning, and C.-J. Hsieh, ``Robust
  deep reinforcement learning against adversarial perturbations on state
  observations,'' \emph{NeurIPS}, 2020.

\bibitem{aafaq2019video}
N.~Aafaq, A.~Mian, W.~Liu, S.~Z. Gilani, and M.~Shah, ``Video description: A
  survey of methods, datasets, and evaluation metrics,'' \emph{ACM Computing
  Surveys (CSUR)}, vol.~52, no.~6, pp. 1--37, 2019.

\bibitem{xu2018fooling}
X.~Xu, X.~Chen, C.~Liu, A.~Rohrbach, T.~Darrell, and D.~Song, ``Fooling vision
  and language models despite localization and attention mechanism,'' in
  \emph{Proceedings of the IEEE Conference on Computer Vision and Pattern
  Recognition}, 2018, pp. 4951--4961.

\bibitem{chen2017attacking}
H.~Chen, H.~Zhang, P.-Y. Chen, J.~Yi, and C.-J. Hsieh, ``Attacking visual
  language grounding with adversarial examples: A case study on neural image
  captioning,'' \emph{arXiv preprint arXiv:1712.02051}, 2017.

\bibitem{vinyals2015show}
O.~Vinyals, A.~Toshev, S.~Bengio, and D.~Erhan, ``Show and tell: A neural image
  caption generator,'' in \emph{Proceedings of the IEEE conference on computer
  vision and pattern recognition}, 2015, pp. 3156--3164.

\bibitem{xu2019exact}
Y.~Xu, B.~Wu, F.~Shen, Y.~Fan, Y.~Zhang, H.~T. Shen, and W.~Liu, ``Exact
  adversarial attack to image captioning via structured output learning with
  latent variables,'' in \emph{Proceedings of the IEEE/CVF Conference on
  Computer Vision and Pattern Recognition}, 2019, pp. 4135--4144.

\bibitem{xu2020machines}
X.~Xu, J.~Chen, J.~Xiao, L.~Gao, F.~Shen, and H.~T. Shen, ``What machines see
  is not what they get: Fooling scene text recognition models with adversarial
  text images,'' in \emph{Proceedings of the IEEE/CVF Conference on Computer
  Vision and Pattern Recognition}, 2020, pp. 12\,304--12\,314.

\bibitem{zhu2019freelb}
C.~Zhu, Y.~Cheng, Z.~Gan, S.~Sun, T.~Goldstein, and J.~Liu, ``Freelb: Enhanced
  adversarial training for natural language understanding,'' in
  \emph{International Conference on Learning Representations}, 2019.

\bibitem{deng2019arcface}
J.~Deng, J.~Guo, N.~Xue, and S.~Zafeiriou, ``Arcface: Additive angular margin
  loss for deep face recognition,'' in \emph{Proceedings of the IEEE/CVF
  Conference on Computer Vision and Pattern Recognition}, 2019, pp. 4690--4699.

\bibitem{goswami2018unravelling}
G.~Goswami, N.~Ratha, A.~Agarwal, R.~Singh, and M.~Vatsa, ``Unravelling
  robustness of deep learning based face recognition against adversarial
  attacks,'' in \emph{Proceedings of the AAAI Conference on Artificial
  Intelligence}, vol.~32, no.~1, 2018.

\bibitem{amos2016openface}
B.~Amos, B.~Ludwiczuk, J.~Harkes, P.~Pillai, K.~Elgazzar, and
  M.~Satyanarayanan, ``Openface: Face recognition with deep neural networks,''
  in \emph{IEEE Winter Conference on Applications of Computer Vision}, vol.~1,
  no.~2, 2016, p.~6.

\bibitem{BMVC2015_41}
O.~M. Parkhi, A.~Vedaldi, and A.~Zisserman, ``Deep face recognition,'' in
  \emph{Proceedings of the British Machine Vision Conference (BMVC)}.\hskip 1em
  plus 0.5em minus 0.4em\relax BMVA Press, September 2015, pp. 41.1--41.12.

\bibitem{dong2019efficient}
Y.~Dong, H.~Su, B.~Wu, Z.~Li, W.~Liu, T.~Zhang, and J.~Zhu, ``Efficient
  decision-based black-box adversarial attacks on face recognition,'' in
  \emph{Proceedings of the IEEE/CVF Conference on Computer Vision and Pattern
  Recognition}, 2019, pp. 7714--7722.

\bibitem{hansen2001completely}
N.~Hansen and A.~Ostermeier, ``Completely derandomized self-adaptation in
  evolution strategies,'' \emph{Evolutionary computation}, vol.~9, no.~2, pp.
  159--195, 2001.

\bibitem{zhong2020towards}
Y.~Zhong and W.~Deng, ``Towards transferable adversarial attack against deep
  face recognition,'' \emph{IEEE Transactions on Information Forensics and
  Security}, vol.~16, pp. 1452--1466, 2020.

\bibitem{chatzikyriakidis2019adversarial}
E.~Chatzikyriakidis, C.~Papaioannidis, and I.~Pitas, ``Adversarial face
  de-identification,'' in \emph{2019 IEEE International Conference on Image
  Processing (ICIP)}.\hskip 1em plus 0.5em minus 0.4em\relax IEEE, 2019, pp.
  684--688.

\bibitem{kwon2019face}
H.~Kwon, O.~Kwon, H.~Yoon, and K.-W. Park, ``Face friend-safe adversarial
  example on face recognition system,'' in \emph{2019 Eleventh International
  Conference on Ubiquitous and Future Networks (ICUFN)}.\hskip 1em plus 0.5em
  minus 0.4em\relax IEEE, 2019, pp. 547--551.

\bibitem{dabouei2019fast}
A.~Dabouei, S.~Soleymani, J.~Dawson, and N.~Nasrabadi, ``Fast
  geometrically-perturbed adversarial faces,'' in \emph{2019 IEEE Winter
  Conference on Applications of Computer Vision (WACV)}.\hskip 1em plus 0.5em
  minus 0.4em\relax IEEE, 2019, pp. 1979--1988.

\bibitem{Xiao_2021_CVPR}
Z.~Xiao, X.~Gao, C.~Fu, Y.~Dong, W.~Gao, X.~Zhang, J.~Zhou, and J.~Zhu,
  ``Improving transferability of adversarial patches on face recognition with
  generative models,'' in \emph{Proceedings of the IEEE/CVF Conference on
  Computer Vision and Pattern Recognition (CVPR)}, June 2021, pp.
  11\,845--11\,854.

\bibitem{yang2020attacks}
L.~Yang, Q.~Song, and Y.~Wu, ``Attacks on state-of-the-art face recognition
  using attentional adversarial attack generative network,'' \emph{Multimedia
  Tools and Applications}, pp. 1--21, 2020.

\bibitem{deb2019advfaces}
D.~Deb, J.~Zhang, and A.~K. Jain, ``Advfaces: Adversarial face synthesis,'' in
  \emph{2020 IEEE International Joint Conference on Biometrics (IJCB)}.\hskip
  1em plus 0.5em minus 0.4em\relax IEEE, 2019, pp. 1--10.

\bibitem{Li_2021_CVPR}
D.~Li, W.~Wang, H.~Fan, and J.~Dong, ``Exploring adversarial fake images on
  face manifold,'' in \emph{Proceedings of the IEEE/CVF Conference on Computer
  Vision and Pattern Recognition (CVPR)}, June 2021, pp. 5789--5798.

\bibitem{tolosana2020deepfakes}
R.~Tolosana, R.~Vera-Rodriguez, J.~Fierrez, A.~Morales, and J.~Ortega-Garcia,
  ``Deepfakes and beyond: A survey of face manipulation and fake detection,''
  \emph{Information Fusion}, vol.~64, pp. 131--148, 2020.

\bibitem{sharif2016accessorize}
M.~Sharif, S.~Bhagavatula, L.~Bauer, and M.~K. Reiter, ``Accessorize to a
  crime: Real and stealthy attacks on state-of-the-art face recognition,'' in
  \emph{Proceedings of the 2016 acm sigsac conference on computer and
  communications security}, 2016, pp. 1528--1540.

\bibitem{zhou2018invisible}
Z.~Zhou, D.~Tang, X.~Wang, W.~Han, X.~Liu, and K.~Zhang, ``Invisible mask:
  Practical attacks on face recognition with infrared,'' \emph{arXiv preprint
  arXiv:1803.04683}, 2018.

\bibitem{nguyen2020adversarial}
D.-L. Nguyen, S.~S. Arora, Y.~Wu, and H.~Yang, ``Adversarial light projection
  attacks on face recognition systems: A feasibility study,'' in
  \emph{Proceedings of the IEEE/CVF Conference on Computer Vision and Pattern
  Recognition Workshops}, 2020, pp. 814--815.

\bibitem{schroff2015facenet}
F.~Schroff, D.~Kalenichenko, and J.~Philbin, ``Facenet: A unified embedding for
  face recognition and clustering,'' in \emph{Proceedings of the IEEE
  conference on computer vision and pattern recognition}, 2015, pp. 815--823.

\bibitem{komkov2019advhat}
S.~Komkov and A.~Petiushko, ``Advhat: Real-world adversarial attack on arcface
  face id system,'' \emph{arXiv preprint arXiv:1908.08705}, 2019.

\bibitem{pautov2019adversarial}
M.~Pautov, G.~Melnikov, E.~Kaziakhmedov, K.~Kireev, and A.~Petiushko, ``On
  adversarial patches: real-world attack on arcface-100 face recognition
  system,'' in \emph{2019 International Multi-Conference on Engineering,
  Computer and Information Sciences (SIBIRCON)}.\hskip 1em plus 0.5em minus
  0.4em\relax IEEE, 2019, pp. 0391--0396.

\bibitem{shao2019multi}
R.~Shao, X.~Lan, J.~Li, and P.~C. Yuen, ``Multi-adversarial discriminative deep
  domain generalization for face presentation attack detection,'' in
  \emph{Proceedings of the IEEE/CVF Conference on Computer Vision and Pattern
  Recognition}, 2019, pp. 10\,023--10\,031.

\bibitem{nakka2020indirect}
K.~K. Nakka and M.~Salzmann, ``Indirect local attacks for context-aware
  semantic segmentation networks,'' in \emph{European Conference on Computer
  Vision}.\hskip 1em plus 0.5em minus 0.4em\relax Springer, 2020, pp. 611--628.

\bibitem{he2020segmentations}
Y.~He, S.~Rahimian, B.~Schiele, and M.~Fritz, ``Segmentations-leak: Membership
  inference attacks and defenses in semantic image segmentation,'' in
  \emph{European Conference on Computer Vision}.\hskip 1em plus 0.5em minus
  0.4em\relax Springer, 2020, pp. 519--535.

\bibitem{choi2019evaluating}
J.-H. Choi, H.~Zhang, J.-H. Kim, C.-J. Hsieh, and J.-S. Lee, ``Evaluating
  robustness of deep image super-resolution against adversarial attacks,'' in
  \emph{Proceedings of the IEEE/CVF International Conference on Computer
  Vision}, 2019, pp. 303--311.

\bibitem{Mehra_2021_CVPRHow}
A.~Mehra, B.~Kailkhura, P.-Y. Chen, and J.~Hamm, ``How robust are randomized
  smoothing based defenses to data poisoning?'' in \emph{Proceedings of the
  IEEE/CVF Conference on Computer Vision and Pattern Recognition (CVPR)}, June
  2021, pp. 13\,244--13\,253.

\bibitem{wong2020targeted}
A.~Wong, S.~Cicek, and S.~Soatto, ``Targeted adversarial perturbations for
  monocular depth prediction,'' \emph{arXiv preprint arXiv:2006.08602}, 2020.

\bibitem{bai2020targeted}
J.~Bai, B.~Chen, Y.~Li, D.~Wu, W.~Guo, S.-t. Xia, and E.-h. Yang, ``Targeted
  attack for deep hashing based retrieval,'' in \emph{European Conference on
  Computer Vision}.\hskip 1em plus 0.5em minus 0.4em\relax Springer, 2020, pp.
  618--634.

\bibitem{li2019universal}
J.~Li, R.~Ji, H.~Liu, X.~Hong, Y.~Gao, and Q.~Tian, ``Universal perturbation
  attack against image retrieval,'' in \emph{Proceedings of the IEEE/CVF
  International Conference on Computer Vision}, 2019, pp. 4899--4908.

\bibitem{zhang2020adversarial}
Z.~Zhang, Z.~Zhang, Y.~Zhou, Y.~Shen, R.~Jin, and D.~Dou, ``Adversarial attacks
  on deep graph matching,'' \emph{Advances in Neural Information Processing
  Systems}, vol.~33, 2020.

\bibitem{yang2020patchattack}
C.~Yang, A.~Kortylewski, C.~Xie, Y.~Cao, and A.~Yuille, ``Patchattack: A
  black-box texture-based attack with reinforcement learning,'' in
  \emph{European Conference on Computer Vision}.\hskip 1em plus 0.5em minus
  0.4em\relax Springer, 2020, pp. 681--698.

\bibitem{yang2019design}
X.~Yang, F.~Wei, H.~Zhang, X.~Ming, and J.~Zhu, ``Design and interpretation of
  universal adversarial patches in face detection,'' \emph{ECCV}, 2020.

\bibitem{ranjan2019attacking}
A.~Ranjan, J.~Janai, A.~Geiger, and M.~J. Black, ``Attacking optical flow,'' in
  \emph{Proceedings of the IEEE/CVF International Conference on Computer
  Vision}, 2019, pp. 2404--2413.

\bibitem{cao2019adversarial}
Y.~Cao, C.~Xiao, B.~Cyr, Y.~Zhou, W.~Park, S.~Rampazzi, Q.~A. Chen, K.~Fu, and
  Z.~M. Mao, ``Adversarial sensor attack on lidar-based perception in
  autonomous driving,'' in \emph{Proceedings of the 2019 ACM SIGSAC Conference
  on Computer and Communications Security}, 2019, pp. 2267--2281.

\bibitem{Wang_2021_CVPRDual}
J.~Wang, A.~Liu, Z.~Yin, S.~Liu, S.~Tang, and X.~Liu, ``Dual attention
  suppression attack: Generate adversarial camouflage in physical world,'' in
  \emph{Proceedings of the IEEE/CVF Conference on Computer Vision and Pattern
  Recognition (CVPR)}, June 2021, pp. 8565--8574.

\bibitem{zhang2018camou}
Y.~Zhang, H.~Foroosh, P.~David, and B.~Gong, ``Camou: Learning physical vehicle
  camouflages to adversarially attack detectors in the wild,'' in
  \emph{International Conference on Learning Representations}, 2019.

\bibitem{kong2020physgan}
Z.~Kong, J.~Guo, A.~Li, and C.~Liu, ``Physgan: Generating
  physical-world-resilient adversarial examples for autonomous driving,'' in
  \emph{Proceedings of the IEEE/CVF Conference on Computer Vision and Pattern
  Recognition}, 2020, pp. 14\,254--14\,263.

\bibitem{ho2019catastrophic}
C.-H. Ho, B.~Leung, E.~Sandstrom, Y.~Chang, and N.~Vasconcelos, ``Catastrophic
  child's play: easy to perform, hard to defend adversarial attacks,'' in
  \emph{Proceedings of the IEEE/CVF Conference on Computer Vision and Pattern
  Recognition}, 2019, pp. 9229--9237.

\bibitem{duan2020adversarial}
R.~Duan, X.~Ma, Y.~Wang, J.~Bailey, A.~K. Qin, and Y.~Yang, ``Adversarial
  camouflage: Hiding physical-world attacks with natural styles,'' in
  \emph{Proceedings of the IEEE/CVF Conference on Computer Vision and Pattern
  Recognition}, 2020, pp. 1000--1008.

\bibitem{jing2019neural}
Y.~Jing, Y.~Yang, Z.~Feng, J.~Ye, Y.~Yu, and M.~Song, ``Neural style transfer:
  A review,'' \emph{IEEE transactions on visualization and computer graphics},
  2019.

\bibitem{liu2020bias}
A.~Liu, J.~Wang, X.~Liu, B.~Cao, C.~Zhang, and H.~Yu, ``Bias-based universal
  adversarial patch attack for automatic check-out,'' in \emph{Proc. Eur. Conf.
  Comput. Vis.}\hskip 1em plus 0.5em minus 0.4em\relax Springer, 2020, pp.
  395--410.

\bibitem{xu2020adversarial}
K.~Xu, G.~Zhang, S.~Liu, Q.~Fan, M.~Sun, H.~Chen, P.-Y. Chen, Y.~Wang, and
  X.~Lin, ``Adversarial t-shirt! evading person detectors in a physical
  world,'' in \emph{European Conference on Computer Vision}.\hskip 1em plus
  0.5em minus 0.4em\relax Springer, 2020, pp. 665--681.

\bibitem{braunegg2020apricot}
A.~Braunegg, A.~Chakraborty, M.~Krumdick, N.~Lape, S.~Leary, K.~Manville,
  E.~Merkhofer, L.~Strickhart, and M.~Walmer, ``Apricot: A dataset of physical
  adversarial attacks on object detection,'' in \emph{European Conference on
  Computer Vision}.\hskip 1em plus 0.5em minus 0.4em\relax Springer, 2020, pp.
  35--50.

\bibitem{Sayles_2021_CVPR}
A.~Sayles, A.~Hooda, M.~Gupta, R.~Chatterjee, and E.~Fernandes, ``Invisible
  perturbations: Physical adversarial examples exploiting the rolling shutter
  effect,'' in \emph{Proceedings of the IEEE/CVF Conference on Computer Vision
  and Pattern Recognition (CVPR)}, June 2021, pp. 14\,666--14\,675.

\bibitem{Duan_2021_CVPRLaser}
R.~Duan, X.~Mao, A.~K. Qin, Y.~Chen, S.~Ye, Y.~He, and Y.~Yang, ``Adversarial
  laser beam: Effective physical-world attack to dnns in a blink,'' in
  \emph{Proceedings of the IEEE/CVF Conference on Computer Vision and Pattern
  Recognition (CVPR)}, June 2021, pp. 16\,062--16\,071.

\bibitem{xie2020adversarial}
C.~Xie, M.~Tan, B.~Gong, J.~Wang, A.~L. Yuille, and Q.~V. Le, ``Adversarial
  examples improve image recognition,'' in \emph{Proceedings of the IEEE/CVF
  Conference on Computer Vision and Pattern Recognition}, 2020, pp. 819--828.

\bibitem{tan2019efficientnet}
M.~Tan and Q.~V. Le, ``Efficientnet: Rethinking model scaling for convolutional
  neural networks,'' \emph{arXiv preprint arXiv:1905.11946}, 2019.

\bibitem{raghunathan2019adversarial}
A.~Raghunathan, S.~M. Xie, F.~Yang, J.~C. Duchi, and P.~Liang, ``Adversarial
  training can hurt generalization,'' \emph{arXiv preprint arXiv:1906.06032},
  2019.

\bibitem{li2019inductive}
Y.~Li, E.~X. Fang, H.~Xu, and T.~Zhao, ``Inductive bias of gradient descent
  based adversarial training on separable data,'' \emph{arXiv preprint
  arXiv:1906.02931}, 2019.

\bibitem{qiao2018deep}
S.~Qiao, W.~Shen, Z.~Zhang, B.~Wang, and A.~Yuille, ``Deep co-training for
  semi-supervised image recognition,'' in \emph{Proceedings of the european
  conference on computer vision (eccv)}, 2018, pp. 135--152.

\bibitem{ho2020contrastive}
C.-H. Ho and N.~Vasconcelos, ``Contrastive learning with adversarial
  examples,'' \emph{NeurIPS}.

\bibitem{salman2020adversarially}
H.~Salman, A.~Ilyas, L.~Engstrom, A.~Kapoor, and A.~Madry, ``Do adversarially
  robust imagenet models transfer better?'' \emph{arXiv preprint
  arXiv:2007.08489}, 2020.

\bibitem{gan2020large}
Z.~Gan, Y.-C. Chen, L.~Li, C.~Zhu, Y.~Cheng, and J.~Liu, ``Large-scale
  adversarial training for vision-and-language representation learning,''
  \emph{arXiv preprint arXiv:2006.06195}, 2020.

\bibitem{Lee_2021_CVPRAnti}
J.~Lee, E.~Kim, and S.~Yoon, ``Anti-adversarially manipulated attributions for
  weakly and semi-supervised semantic segmentation,'' in \emph{Proceedings of
  the IEEE/CVF Conference on Computer Vision and Pattern Recognition (CVPR)},
  June 2021, pp. 4071--4080.

\bibitem{jalwana2020attack}
M.~A. Jalwana, N.~Akhtar, M.~Bennamoun, and A.~Mian, ``Attack to explain deep
  representation,'' in \emph{Proceedings of the IEEE/CVF Conference on Computer
  Vision and Pattern Recognition}, 2020, pp. 9543--9552.

\bibitem{santurkar2019image}
S.~Santurkar, A.~Ilyas, D.~Tsipras, L.~Engstrom, B.~Tran, and A.~Madry, ``Image
  synthesis with a single (robust) classifier,'' in \emph{Advances in Neural
  Information Processing Systems}, 2019, pp. 1262--1273.

\bibitem{augustin2020adversarial}
M.~Augustin, A.~Meinke, and M.~Hein, ``Adversarial robustness on in-and
  out-distribution improves explainability,'' in \emph{European Conference on
  Computer Vision}.\hskip 1em plus 0.5em minus 0.4em\relax Springer, 2020, pp.
  228--245.

\bibitem{Elliott_2021_CVPRExplaining}
A.~Elliott, S.~Law, and C.~Russell, ``Explaining classifiers using adversarial
  perturbations on the perceptual ball,'' in \emph{Proceedings of the IEEE/CVF
  Conference on Computer Vision and Pattern Recognition (CVPR)}, June 2021, pp.
  10\,693--10\,702.

\bibitem{xu2018structured}
K.~Xu, S.~Liu, P.~Zhao, P.-Y. Chen, H.~Zhang, Q.~Fan, D.~Erdogmus, Y.~Wang, and
  X.~Lin, ``Structured adversarial attack: Towards general implementation and
  better interpretability,'' \emph{ICLR}, 2019.

\bibitem{elsayed2018adversarial}
G.~F. Elsayed, I.~Goodfellow, and J.~Sohl-Dickstein, ``Adversarial
  reprogramming of neural networks,'' \emph{arXiv preprint arXiv:1806.11146},
  2018.

\bibitem{sakaguchi2020winogrande}
K.~Sakaguchi, R.~Le~Bras, C.~Bhagavatula, and Y.~Choi, ``Winogrande: An
  adversarial winograd schema challenge at scale,'' in \emph{Proceedings of the
  AAAI Conference on Artificial Intelligence}, vol.~34, no.~05, 2020, pp.
  8732--8740.

\bibitem{le2020adversarial}
R.~Le~Bras, S.~Swayamdipta, C.~Bhagavatula, R.~Zellers, M.~Peters,
  A.~Sabharwal, and Y.~Choi, ``Adversarial filters of dataset biases,'' in
  \emph{International Conference on Machine Learning}.\hskip 1em plus 0.5em
  minus 0.4em\relax PMLR, 2020, pp. 1078--1088.

\bibitem{tanay2016boundary}
T.~Tanay and L.~Griffin, ``A boundary tilting persepective on the phenomenon of
  adversarial examples,'' \emph{arXiv preprint arXiv:1608.07690}, 2016.

\bibitem{krotov2018dense}
D.~Krotov and J.~Hopfield, ``Dense associative memory is robust to adversarial
  inputs,'' \emph{Neural computation}, vol.~30, no.~12, pp. 3151--3167, 2018.

\bibitem{krotov2016dense}
D.~Krotov and J.~J. Hopfield, ``Dense associative memory for pattern
  recognition,'' \emph{Advances in neural information processing systems},
  vol.~29, pp. 1172--1180, 2016.

\bibitem{taghanaki2019kernelized}
S.~A. Taghanaki, K.~Abhishek, S.~Azizi, and G.~Hamarneh, ``A kernelized
  manifold mapping to diminish the effect of adversarial perturbations,'' in
  \emph{Proceedings of the IEEE/CVF Conference on Computer Vision and Pattern
  Recognition}, 2019, pp. 11\,340--11\,349.

\bibitem{cubuk2017intriguing}
E.~D. Cubuk, B.~Zoph, S.~S. Schoenholz, and Q.~V. Le, ``Intriguing properties
  of adversarial examples,'' \emph{ICLR}, 2018.

\bibitem{rozsa2016accuracy}
A.~Rozsa, M.~G{\"u}nther, and T.~E. Boult, ``Are accuracy and robustness
  correlated,'' in \emph{2016 15th IEEE international conference on machine
  learning and applications (ICMLA)}.\hskip 1em plus 0.5em minus 0.4em\relax
  IEEE, 2016, pp. 227--232.

\bibitem{rozsa2016towards}
A.~Rozsa, M.~Gunther, and T.~E. Boult, ``Towards robust deep neural networks
  with bang,'' \emph{WACV}, 2018.

\bibitem{tabacof2016exploring}
P.~Tabacof and E.~Valle, ``Exploring the space of adversarial images,'' in
  \emph{2016 International Joint Conference on Neural Networks (IJCNN)}.\hskip
  1em plus 0.5em minus 0.4em\relax IEEE, 2016, pp. 426--433.

\bibitem{tramer2017space}
F.~Tram{\`e}r, N.~Papernot, I.~Goodfellow, D.~Boneh, and P.~McDaniel, ``The
  space of transferable adversarial examples,'' \emph{arXiv preprint
  arXiv:1704.03453}, 2017.

\bibitem{li2020defense}
Y.~Li, S.~Cheng, H.~Su, and J.~Zhu, ``Defense against adversarial attacks via
  controlling gradient leaking on embedded manifolds,'' in \emph{European
  Conference on Computer Vision}.\hskip 1em plus 0.5em minus 0.4em\relax
  Springer, 2020, pp. 753--769.

\bibitem{jacobsen2018excessive}
J.-H. Jacobsen, J.~Behrmann, R.~Zemel, and M.~Bethge, ``Excessive invariance
  causes adversarial vulnerability,'' \emph{ICLR}, 2019.

\bibitem{reddy2020biologically}
M.~V. Reddy, A.~Banburski, N.~Pant, and T.~Poggio, ``Biologically inspired
  mechanisms for adversarial robustness,'' \emph{arXiv preprint
  arXiv:2006.16427}, 2020.

\bibitem{pal2020game}
A.~Pal and R.~Vidal, ``A game theoretic analysis of additive adversarial
  attacks and defenses,'' \emph{NeurIPS}, 2020.

\bibitem{cohen2019certified}
J.~M. Cohen, E.~Rosenfeld, and J.~Z. Kolter, ``Certified adversarial robustness
  via randomized smoothing,'' \emph{ICML}, 2019.

\bibitem{nash1950equilibrium}
J.~F. Nash \emph{et~al.}, ``Equilibrium points in n-person games,''
  \emph{Proceedings of the national academy of sciences}, vol.~36, no.~1, pp.
  48--49, 1950.

\bibitem{daniely2020most}
A.~Daniely and H.~Schacham, ``Most relu networks suffer from l2 adversarial
  perturbations,'' \emph{NeurIPS}, 2020.

\bibitem{shafahi2018adversarial}
A.~Shafahi, W.~R. Huang, C.~Studer, S.~Feizi, and T.~Goldstein, ``Are
  adversarial examples inevitable?'' \emph{arXiv preprint arXiv:1809.02104},
  2018.

\bibitem{gilmer2018adversarial}
J.~Gilmer, L.~Metz, F.~Faghri, S.~S. Schoenholz, M.~Raghu, M.~Wattenberg, and
  I.~Goodfellow, ``Adversarial spheres,'' \emph{arXiv preprint
  arXiv:1801.02774}, 2018.

\bibitem{song2018generative}
Y.~Song, R.~Shu, N.~Kushman, and S.~Ermon, ``Generative adversarial examples,''
  \emph{arXiv preprint arXiv:1805.07894}, 2018.

\bibitem{stutz2019disentangling}
D.~Stutz, M.~Hein, and B.~Schiele, ``Disentangling adversarial robustness and
  generalization,'' in \emph{Proceedings of the IEEE/CVF Conference on Computer
  Vision and Pattern Recognition}, 2019, pp. 6976--6987.

\bibitem{tsipras2018robustness}
D.~Tsipras, S.~Santurkar, L.~Engstrom, A.~Turner, and A.~Madry, ``Robustness
  may be at odds with accuracy,'' \emph{ICLR}, 2019.

\bibitem{su2018robustness}
D.~Su, H.~Zhang, H.~Chen, J.~Yi, P.-Y. Chen, and Y.~Gao, ``Is robustness the
  cost of accuracy?--a comprehensive study on the robustness of 18 deep image
  classification models,'' in \emph{Proceedings of the European Conference on
  Computer Vision (ECCV)}, 2018, pp. 631--648.

\bibitem{moosavi2017analysis}
S.-M. Moosavi-Dezfooli, A.~Fawzi, O.~Fawzi, P.~Frossard, and S.~Soatto,
  ``Analysis of universal adversarial perturbations,'' \emph{arXiv preprint
  arXiv:1705.09554}, 2017.

\bibitem{moosavi2018robustness}
------, ``Robustness of classifiers to universal perturbations: A geometric
  perspective,'' in \emph{International Conference on Learning
  Representations}, 2018.

\bibitem{jetley2018friends}
S.~Jetley, N.~Lord, and P.~Torr, ``With friends like these, who needs
  adversaries?'' in \emph{Advances in neural information processing systems},
  2018, pp. 10\,749--10\,759.

\bibitem{zhang2020understanding}
C.~Zhang, P.~Benz, T.~Imtiaz, and I.~S. Kweon, ``Understanding adversarial
  examples from the mutual influence of images and perturbations,'' in
  \emph{Proceedings of the IEEE/CVF Conference on Computer Vision and Pattern
  Recognition}, 2020, pp. 14\,521--14\,530.

\bibitem{ilyas2019adversarial}
A.~Ilyas, S.~Santurkar, D.~Tsipras, L.~Engstrom, B.~Tran, and A.~Madry,
  ``Adversarial examples are not bugs, they are features,'' in \emph{Advances
  in Neural Information Processing Systems}, 2019, pp. 125--136.

\bibitem{bubeck2019adversarial}
S.~Bubeck, Y.~T. Lee, E.~Price, and I.~Razenshteyn, ``Adversarial examples from
  computational constraints,'' in \emph{International Conference on Machine
  Learning}.\hskip 1em plus 0.5em minus 0.4em\relax PMLR, 2019, pp. 831--840.

\bibitem{wu2020skip}
D.~Wu, Y.~Wang, S.-T. Xia, J.~Bailey, and X.~Ma, ``Skip connections matter: On
  the transferability of adversarial examples generated with resnets,''
  \emph{arXiv preprint arXiv:2002.05990}, 2020.

\bibitem{zhang2019limitations}
H.~Zhang, H.~Chen, Z.~Song, D.~Boning, I.~S. Dhillon, and C.-J. Hsieh, ``The
  limitations of adversarial training and the blind-spot attack,'' \emph{ICLR},
  2020.

\bibitem{schott2018towards}
L.~Schott, J.~Rauber, M.~Bethge, and W.~Brendel, ``Towards the first
  adversarially robust neural network model on mnist,'' \emph{ICLR}, 2019.

\bibitem{ding2019sensitivity}
G.~W. Ding, K.~Y.~C. Lui, X.~Jin, L.~Wang, and R.~Huang, ``On the sensitivity
  of adversarial robustness to input data distributions.'' in \emph{ICLR
  (Poster)}, 2019.

\bibitem{song2018improving}
C.~Song, K.~He, L.~Wang, and J.~E. Hopcroft, ``Improving the generalization of
  adversarial training with domain adaptation,'' \emph{arXiv preprint
  arXiv:1810.00740}, 2018.

\bibitem{geirhos2018imagenet}
R.~Geirhos, P.~Rubisch, C.~Michaelis, M.~Bethge, F.~A. Wichmann, and
  W.~Brendel, ``Imagenet-trained cnns are biased towards texture; increasing
  shape bias improves accuracy and robustness,'' \emph{ICLR}, 2019.

\bibitem{zhang2019interpreting}
T.~Zhang and Z.~Zhu, ``Interpreting adversarially trained convolutional neural
  networks,'' \emph{ICML}, 2019.

\bibitem{Gong_2021_CVPRMaxup}
C.~Gong, T.~Ren, M.~Ye, and Q.~Liu, ``Maxup: Lightweight adversarial training
  with data augmentation improves neural network training,'' in
  \emph{Proceedings of the IEEE/CVF Conference on Computer Vision and Pattern
  Recognition (CVPR)}, June 2021, pp. 2474--2483.

\bibitem{wang2019improving}
Y.~Wang, D.~Zou, J.~Yi, J.~Bailey, X.~Ma, and Q.~Gu, ``Improving adversarial
  robustness requires revisiting misclassified examples,'' in
  \emph{International Conference on Learning Representations}, 2019.

\bibitem{gowal2020achieving}
S.~Gowal, C.~Qin, P.-S. Huang, T.~Cemgil, K.~Dvijotham, T.~Mann, and P.~Kohli,
  ``Achieving robustness in the wild via adversarial mixing with disentangled
  representations,'' in \emph{Proceedings of the IEEE/CVF Conference on
  Computer Vision and Pattern Recognition}, 2020, pp. 1211--1220.

\bibitem{karras2019style}
T.~Karras, S.~Laine, and T.~Aila, ``A style-based generator architecture for
  generative adversarial networks,'' in \emph{Proceedings of the IEEE
  conference on computer vision and pattern recognition}, 2019, pp. 4401--4410.

\bibitem{ding2019mma}
G.~W. Ding, Y.~Sharma, K.~Y.~C. Lui, and R.~Huang, ``Mma training: Direct input
  space margin maximization through adversarial training,'' in
  \emph{International Conference on Learning Representations}, 2019.

\bibitem{balunovic2019adversarial}
M.~Balunovic and M.~Vechev, ``Adversarial training and provable defenses:
  Bridging the gap,'' in \emph{International Conference on Learning
  Representations}, 2019.

\bibitem{vivek2020single}
B.~Vivek and R.~V. Babu, ``Single-step adversarial training with dropout
  scheduling,'' in \emph{2020 IEEE/CVF Conference on Computer Vision and
  Pattern Recognition (CVPR)}.\hskip 1em plus 0.5em minus 0.4em\relax IEEE,
  2020, pp. 947--956.

\bibitem{song2019robust}
C.~Song, K.~He, J.~Lin, L.~Wang, and J.~E. Hopcroft, ``Robust local features
  for improving the generalization of adversarial training,'' \emph{ICLR},
  2020.

\bibitem{farnia2018generalizable}
F.~Farnia, J.~M. Zhang, and D.~Tse, ``Generalizable adversarial training via
  spectral normalization,'' \emph{ICLR}, 2019.

\bibitem{xiao2020one}
C.~Xiao and C.~Zheng, ``One man's trash is another man's treasure: Resisting
  adversarial examples by adversarial examples,'' in \emph{Proceedings of the
  IEEE/CVF Conference on Computer Vision and Pattern Recognition}, 2020, pp.
  412--421.

\bibitem{naseer2020self}
M.~Naseer, S.~Khan, M.~Hayat, F.~S. Khan, and F.~Porikli, ``A self-supervised
  approach for adversarial robustness,'' in \emph{Proceedings of the IEEE/CVF
  Conference on Computer Vision and Pattern Recognition}, 2020, pp. 262--271.

\bibitem{chen2020adversarial}
T.~Chen, S.~Liu, S.~Chang, Y.~Cheng, L.~Amini, and Z.~Wang, ``Adversarial
  robustness: From self-supervised pre-training to fine-tuning,'' in
  \emph{Proceedings of the IEEE/CVF Conference on Computer Vision and Pattern
  Recognition}, 2020, pp. 699--708.

\bibitem{jang2019adversarial}
Y.~Jang, T.~Zhao, S.~Hong, and H.~Lee, ``Adversarial defense via learning to
  generate diverse attacks,'' in \emph{Proceedings of the IEEE/CVF
  International Conference on Computer Vision}, 2019, pp. 2740--2749.

\bibitem{lin2020dual}
W.-A. Lin, C.~P. Lau, A.~Levine, R.~Chellappa, and S.~Feizi, ``Dual manifold
  adversarial robustness: Defense against lp and non-lp adversarial attacks,''
  \emph{arXiv preprint arXiv:2009.02470}, 2020.

\bibitem{wang2019bilateral}
J.~Wang and H.~Zhang, ``Bilateral adversarial training: Towards fast training
  of more robust models against adversarial attacks,'' in \emph{Proceedings of
  the IEEE/CVF International Conference on Computer Vision}, 2019, pp.
  6629--6638.

\bibitem{ye2019adversarial}
S.~Ye, K.~Xu, S.~Liu, H.~Cheng, J.-H. Lambrechts, H.~Zhang, A.~Zhou, K.~Ma,
  Y.~Wang, and X.~Lin, ``Adversarial robustness vs. model compression, or
  both?'' in \emph{Proceedings of the IEEE/CVF International Conference on
  Computer Vision}, 2019, pp. 111--120.

\bibitem{dong2020adversarial}
Y.~Dong, Z.~Deng, T.~Pang, H.~Su, and J.~Zhu, ``Adversarial distributional
  training for robust deep learning,'' \emph{NeurIPS}, 2020.

\bibitem{madaan2020adversarial}
D.~Madaan, J.~Shin, and S.~J. Hwang, ``Adversarial neural pruning with latent
  vulnerability suppression,'' in \emph{International Conference on Machine
  Learning}.\hskip 1em plus 0.5em minus 0.4em\relax PMLR, 2020, pp. 6575--6585.

\bibitem{maini2020adversarial}
P.~Maini, E.~Wong, and Z.~Kolter, ``Adversarial robustness against the union of
  multiple perturbation models,'' in \emph{International Conference on Machine
  Learning}.\hskip 1em plus 0.5em minus 0.4em\relax PMLR, 2020, pp. 6640--6650.

\bibitem{zhang2017mixup}
H.~Zhang, M.~Cisse, Y.~N. Dauphin, and D.~Lopez-Paz, ``mixup: Beyond empirical
  risk minimization,'' \emph{arXiv preprint arXiv:1710.09412}, 2017.

\bibitem{verma2018manifold}
V.~Verma, A.~Lamb, C.~Beckham, A.~Courville, I.~Mitliagkis, and Y.~Bengio,
  ``Manifold mixup: Encouraging meaningful on-manifold interpolation as a
  regularizer,'' \emph{ICML}, 2019.

\bibitem{lee2020adversarial}
S.~Lee, H.~Lee, and S.~Yoon, ``Adversarial vertex mixup: Toward better
  adversarially robust generalization,'' in \emph{Proceedings of the IEEE/CVF
  Conference on Computer Vision and Pattern Recognition}, 2020, pp. 272--281.

\bibitem{xie2019intriguing}
C.~Xie and A.~Yuille, ``Intriguing properties of adversarial training at
  scale,'' \emph{ICLR}, 2020.

\bibitem{li2019implicit}
Y.~Li, E.~X. Fang, H.~Xu, and T.~Zhao, ``Implicit bias of gradient descent
  based adversarial training on separable data,'' in \emph{International
  Conference on Learning Representations}, 2019.

\bibitem{shafahi2019adversarially}
A.~Shafahi, P.~Saadatpanah, C.~Zhu, A.~Ghiasi, C.~Studer, D.~Jacobs, and
  T.~Goldstein, ``Adversarially robust transfer learning,'' \emph{arXiv
  preprint arXiv:1905.08232}, 2019.

\bibitem{sehwag2020hydra}
V.~Sehwag, S.~Wang, P.~Mittal, and S.~Jana, ``Hydra: Pruning adversarially
  robust neural networks,'' \emph{Advances in Neural Information Processing
  Systems (NeurIPS)}, vol.~7, 2020.

\bibitem{wong2020fast}
E.~Wong, L.~Rice, and J.~Z. Kolter, ``Fast is better than free: Revisiting
  adversarial training,'' \emph{ICLR}, 2020.

\bibitem{andriushchenko2020understanding}
M.~Andriushchenko and N.~Flammarion, ``Understanding and improving fast
  adversarial training,'' \emph{NeurIPS}, 2020.

\bibitem{zhao2020bridging}
P.~Zhao, P.-Y. Chen, P.~Das, K.~N. Ramamurthy, and X.~Lin, ``Bridging mode
  connectivity in loss landscapes and adversarial robustness,'' \emph{ICLR},
  2020.

\bibitem{wu2020adversarial}
D.~Wu, S.-T. Xia, and Y.~Wang, ``Adversarial weight perturbation helps robust
  generalization,'' \emph{Advances in Neural Information Processing Systems},
  vol.~33, 2020.

\bibitem{gao2019convergence}
R.~Gao, T.~Cai, H.~Li, C.-J. Hsieh, L.~Wang, and J.~D. Lee, ``Convergence of
  adversarial training in overparametrized neural networks,'' \emph{Advances in
  Neural Information Processing Systems}, vol.~32, pp. 13\,029--13\,040, 2019.

\bibitem{zhang2020over}
Y.~Zhang, O.~Plevrakis, S.~S. Du, X.~Li, Z.~Song, and S.~Arora,
  ``Over-parameterized adversarial training: An analysis overcoming the curse
  of dimensionality,'' \emph{arXiv preprint arXiv:2002.06668}, 2020.

\bibitem{pang2020boosting}
T.~Pang, X.~Yang, Y.~Dong, K.~Xu, J.~Zhu, and H.~Su, ``Boosting adversarial
  training with hypersphere embedding,'' \emph{arXiv preprint
  arXiv:2002.08619}, 2020.

\bibitem{wu2019defending}
T.~Wu, L.~Tong, and Y.~Vorobeychik, ``Defending against physically realizable
  attacks on image classification,'' \emph{ICLR}, 2020.

\bibitem{cheng2019robust}
Y.~Cheng, L.~Jiang, and W.~Macherey, ``Robust neural machine translation with
  doubly adversarial inputs,'' \emph{arXiv preprint arXiv:1906.02443}, 2019.

\bibitem{guo2020meets}
M.~Guo, Y.~Yang, R.~Xu, Z.~Liu, and D.~Lin, ``When nas meets robustness: In
  search of robust architectures against adversarial attacks,'' in
  \emph{Proceedings of the IEEE/CVF Conference on Computer Vision and Pattern
  Recognition}, 2020, pp. 631--640.

\bibitem{bui2020improving}
A.~Bui, T.~Le, H.~Zhao, P.~Montague, O.~deVel, T.~Abraham, and D.~Phung,
  ``Improving adversarial robustness by enforcing local and global
  compactness,'' \emph{ECCV}, 2020.

\bibitem{liu2018adv}
X.~Liu, Y.~Li, C.~Wu, and C.-J. Hsieh, ``Adv-bnn: Improved adversarial defense
  through robust bayesian neural network,'' \emph{ICLR}, 2019.

\bibitem{jeddi2020learn2perturb}
A.~Jeddi, M.~J. Shafiee, M.~Karg, C.~Scharfenberger, and A.~Wong,
  ``Learn2perturb: an end-to-end feature perturbation learning to improve
  adversarial robustness,'' in \emph{Proceedings of the IEEE/CVF Conference on
  Computer Vision and Pattern Recognition}, 2020, pp. 1241--1250.

\bibitem{li2020enhancing}
G.~Li, S.~Ding, J.~Luo, and C.~Liu, ``Enhancing intrinsic adversarial
  robustness via feature pyramid decoder,'' in \emph{Proceedings of the
  IEEE/CVF Conference on Computer Vision and Pattern Recognition}, 2020, pp.
  800--808.

\bibitem{borkar2020defending}
T.~Borkar, F.~Heide, and L.~Karam, ``Defending against universal attacks
  through selective feature regeneration,'' in \emph{Proceedings of the
  IEEE/CVF Conference on Computer Vision and Pattern Recognition}, 2020, pp.
  709--719.

\bibitem{wang2019direct}
H.~Wang and C.-N. Yu, ``A direct approach to robust deep learning using
  adversarial networks,'' \emph{arXiv preprint arXiv:1905.09591}, 2019.

\bibitem{xie2019feature}
C.~Xie, Y.~Wu, L.~v.~d. Maaten, A.~L. Yuille, and K.~He, ``Feature denoising
  for improving adversarial robustness,'' in \emph{Proceedings of the IEEE/CVF
  Conference on Computer Vision and Pattern Recognition}, 2019, pp. 501--509.

\bibitem{lecuyer2019certified}
M.~Lecuyer, V.~Atlidakis, R.~Geambasu, D.~Hsu, and S.~Jana, ``Certified
  robustness to adversarial examples with differential privacy,'' in \emph{2019
  IEEE Symposium on Security and Privacy (SP)}.\hskip 1em plus 0.5em minus
  0.4em\relax IEEE, 2019, pp. 656--672.

\bibitem{liu2018towards}
X.~Liu, M.~Cheng, H.~Zhang, and C.-J. Hsieh, ``Towards robust neural networks
  via random self-ensemble,'' in \emph{Proceedings of the European Conference
  on Computer Vision (ECCV)}, 2018, pp. 369--385.

\bibitem{he2019parametric}
Z.~He, A.~S. Rakin, and D.~Fan, ``Parametric noise injection: Trainable
  randomness to improve deep neural network robustness against adversarial
  attack,'' in \emph{Proceedings of the IEEE/CVF Conference on Computer Vision
  and Pattern Recognition}, 2019, pp. 588--597.

\bibitem{svoboda2018peernets}
J.~Svoboda, J.~Masci, F.~Monti, M.~M. Bronstein, and L.~Guibas, ``Peernets:
  Exploiting peer wisdom against adversarial attacks,'' \emph{ICLR}, 2019.

\bibitem{hosseini2021dsrna}
R.~Hosseini, X.~Yang, and P.~Xie, ``Dsrna: Differentiable search of robust
  neural architectures,'' in \emph{Proceedings of the IEEE/CVF Conference on
  Computer Vision and Pattern Recognition}, 2021, pp. 6196--6205.

\bibitem{cazenavette2021architectural}
G.~Cazenavette, C.~Murdock, and S.~Lucey, ``Architectural adversarial
  robustness: The case for deep pursuit,'' in \emph{Proceedings of the IEEE/CVF
  Conference on Computer Vision and Pattern Recognition}, 2021, pp. 7150--7158.

\bibitem{chan2019jacobian}
A.~Chan, Y.~Tay, Y.~S. Ong, and J.~Fu, ``Jacobian adversarially regularized
  networks for robustness,'' \emph{ICLR}, 2020.

\bibitem{dabouei2020exploiting}
A.~Dabouei, S.~Soleymani, F.~Taherkhani, J.~Dawson, and N.~M. Nasrabadi,
  ``Exploiting joint robustness to adversarial perturbations,'' in
  \emph{Proceedings of the IEEE/CVF Conference on Computer Vision and Pattern
  Recognition}, 2020, pp. 1122--1131.

\bibitem{tadros2019biologically}
T.~Tadros, G.~Krishnan, R.~Ramyaa, and M.~Bazhenov, ``Biologically inspired
  sleep algorithm for increased generalization and adversarial robustness in
  deep neural networks,'' in \emph{International Conference on Learning
  Representations}, 2019.

\bibitem{addepalli2020towards}
S.~Addepalli, A.~Baburaj, G.~Sriramanan, and R.~V. Babu, ``Towards achieving
  adversarial robustness by enforcing feature consistency across bit planes,''
  in \emph{Proceedings of the IEEE/CVF Conference on Computer Vision and
  Pattern Recognition}, 2020, pp. 1020--1029.

\bibitem{qin2019detecting}
Y.~Qin, N.~Frosst, S.~Sabour, C.~Raffel, G.~Cottrell, and G.~Hinton,
  ``Detecting and diagnosing adversarial images with class-conditional capsule
  reconstructions,'' \emph{ICLR}, 2020.

\bibitem{sabour2017dynamic}
S.~Sabour, N.~Frosst, and G.~E. Hinton, ``Dynamic routing between capsules,''
  in \emph{Advances in neural information processing systems}, 2017, pp.
  3856--3866.

\bibitem{Deng_2021_CVPR}
Z.~Deng, X.~Yang, S.~Xu, H.~Su, and J.~Zhu, ``Libre: A practical bayesian
  approach to adversarial detection,'' in \emph{Proceedings of the IEEE/CVF
  Conference on Computer Vision and Pattern Recognition (CVPR)}, June 2021, pp.
  972--982.

\bibitem{li2020connecting}
S.~Li, S.~Zhu, S.~Paul, A.~Roy-Chowdhury, C.~Song, S.~Krishnamurthy, A.~Swami,
  and K.~S. Chan, ``Connecting the dots: Detecting adversarial perturbations
  using context inconsistency,'' in \emph{European Conference on Computer
  Vision}.\hskip 1em plus 0.5em minus 0.4em\relax Springer, 2020, pp. 396--413.

\bibitem{tao2018attacks}
G.~Tao, S.~Ma, Y.~Liu, and X.~Zhang, ``Attacks meet interpretability:
  Attribute-steered detection of adversarial samples,'' \emph{arXiv preprint
  arXiv:1810.11580}, 2018.

\bibitem{qiu2019adversarial}
Y.~Qiu, J.~Leng, C.~Guo, Q.~Chen, C.~Li, M.~Guo, and Y.~Zhu, ``Adversarial
  defense through network profiling based path extraction,'' in
  \emph{Proceedings of the IEEE/CVF Conference on Computer Vision and Pattern
  Recognition}, 2019, pp. 4777--4786.

\bibitem{liu2019detection}
J.~Liu, W.~Zhang, Y.~Zhang, D.~Hou, Y.~Liu, H.~Zha, and N.~Yu, ``Detection
  based defense against adversarial examples from the steganalysis point of
  view,'' in \emph{Proceedings of the IEEE/CVF Conference on Computer Vision
  and Pattern Recognition}, 2019, pp. 4825--4834.

\bibitem{yin2019gat}
X.~Yin, S.~Kolouri, and G.~K. Rohde, ``Gat: Generative adversarial training for
  adversarial example detection and robust classification,'' in
  \emph{International Conference on Learning Representations}, 2019.

\bibitem{liu2019feature}
Z.~Liu, Q.~Liu, T.~Liu, N.~Xu, X.~Lin, Y.~Wang, and W.~Wen, ``Feature
  distillation: Dnn-oriented jpeg compression against adversarial examples,''
  in \emph{2019 IEEE/CVF Conference on Computer Vision and Pattern Recognition
  (CVPR)}.\hskip 1em plus 0.5em minus 0.4em\relax IEEE, 2019, pp. 860--868.

\bibitem{das2017keeping}
N.~Das, M.~Shanbhogue, S.-T. Chen, F.~Hohman, L.~Chen, M.~E. Kounavis, and
  D.~H. Chau, ``Keeping the bad guys out: Protecting and vaccinating deep
  learning with jpeg compression,'' \emph{arXiv preprint arXiv:1705.02900},
  2017.

\bibitem{guo2017countering}
C.~Guo, M.~Rana, M.~Cisse, and L.~Van Der~Maaten, ``Countering adversarial
  images using input transformations,'' \emph{ICLR}, 2018.

\bibitem{raff2019barrage}
E.~Raff, J.~Sylvester, S.~Forsyth, and M.~McLean, ``Barrage of random
  transforms for adversarially robust defense,'' in \emph{Proceedings of the
  IEEE/CVF Conference on Computer Vision and Pattern Recognition}, 2019, pp.
  6528--6537.

\bibitem{taran2019defending}
O.~Taran, S.~Rezaeifar, T.~Holotyak, and S.~Voloshynovskiy, ``Defending against
  adversarial attacks by randomized diversification,'' in \emph{Proceedings of
  the IEEE/CVF Conference on Computer Vision and Pattern Recognition}, 2019,
  pp. 11\,226--11\,233.

\bibitem{jia2019comdefend}
X.~Jia, X.~Wei, X.~Cao, and H.~Foroosh, ``Comdefend: An efficient image
  compression model to defend adversarial examples,'' in \emph{Proceedings of
  the IEEE/CVF Conference on Computer Vision and Pattern Recognition}, 2019,
  pp. 6084--6092.

\bibitem{theagarajan2019shieldnets}
R.~Theagarajan, M.~Chen, B.~Bhanu, and J.~Zhang, ``Shieldnets: Defending
  against adversarial attacks using probabilistic adversarial robustness,'' in
  \emph{Proceedings of the IEEE/CVF Conference on Computer Vision and Pattern
  Recognition}, 2019, pp. 6988--6996.

\bibitem{sun2019adversarial}
B.~Sun, N.-h. Tsai, F.~Liu, R.~Yu, and H.~Su, ``Adversarial defense by
  stratified convolutional sparse coding,'' in \emph{Proceedings of the
  IEEE/CVF Conference on Computer Vision and Pattern Recognition}, 2019, pp.
  11\,447--11\,456.

\bibitem{samangouei2018defense}
P.~Samangouei, M.~Kabkab, and R.~Chellappa, ``Defense-gan: Protecting
  classifiers against adversarial attacks using generative models,''
  \emph{ICLR}, 2018.

\bibitem{gupta2019ciidefence}
P.~Gupta and E.~Rahtu, ``Ciidefence: Defeating adversarial attacks by fusing
  class-specific image inpainting and image denoising,'' in \emph{Proceedings
  of the IEEE/CVF International Conference on Computer Vision}, 2019, pp.
  6708--6717.

\bibitem{kou2019enhancing}
C.~Kou, H.~K. Lee, E.-C. Chang, and T.~K. Ng, ``Enhancing transformation-based
  defenses against adversarial attacks with a distribution classifier,'' in
  \emph{International Conference on Learning Representations}, 2019.

\bibitem{yuan2020ensemble}
J.~Yuan and Z.~He, ``Ensemble generative cleaning with feedback loops for
  defending adversarial attacks,'' in \emph{Proceedings of the IEEE/CVF
  Conference on Computer Vision and Pattern Recognition}, 2020, pp. 581--590.

\bibitem{cohen2020detecting}
G.~Cohen, G.~Sapiro, and R.~Giryes, ``Detecting adversarial samples using
  influence functions and nearest neighbors,'' in \emph{Proceedings of the
  IEEE/CVF Conference on Computer Vision and Pattern Recognition}, 2020, pp.
  14\,453--14\,462.

\bibitem{croce2019scaling}
F.~Croce, J.~Rauber, and M.~Hein, ``Scaling up the randomized gradient-free
  adversarial attack reveals overestimation of robustness using established
  attacks,'' \emph{International Journal of Computer Vision}, pp. 1--19, 2019.

\bibitem{katz2017reluplex}
G.~Katz, C.~Barrett, D.~L. Dill, K.~Julian, and M.~J. Kochenderfer, ``Reluplex:
  An efficient smt solver for verifying deep neural networks,'' in
  \emph{International Conference on Computer Aided Verification}.\hskip 1em
  plus 0.5em minus 0.4em\relax Springer, 2017, pp. 97--117.

\bibitem{tjeng2017evaluating}
V.~Tjeng, K.~Xiao, and R.~Tedrake, ``Evaluating robustness of neural networks
  with mixed integer programming,'' \emph{ICLR}, 2019.

\bibitem{hein2017formal}
M.~Hein and M.~Andriushchenko, ``Formal guarantees on the robustness of a
  classifier against adversarial manipulation,'' in \emph{Advances in Neural
  Information Processing Systems}, 2017, pp. 2266--2276.

\bibitem{raghunathan2018certified}
A.~Raghunathan, J.~Steinhardt, and P.~Liang, ``Certified defenses against
  adversarial examples,'' \emph{ICLR}, 2018.

\bibitem{wong2018provable}
E.~Wong and Z.~Kolter, ``Provable defenses against adversarial examples via the
  convex outer adversarial polytope,'' in \emph{International Conference on
  Machine Learning}.\hskip 1em plus 0.5em minus 0.4em\relax PMLR, 2018, pp.
  5286--5295.

\bibitem{mirman2018differentiable}
M.~Mirman, T.~Gehr, and M.~Vechev, ``Differentiable abstract interpretation for
  provably robust neural networks,'' in \emph{International Conference on
  Machine Learning}, 2018, pp. 3578--3586.

\bibitem{xiao2018training}
K.~Y. Xiao, V.~Tjeng, N.~M. Shafiullah, and A.~Madry, ``Training for faster
  adversarial robustness verification via inducing relu stability,''
  \emph{ICLR}, 2019.

\bibitem{croce2020provable}
F.~Croce and M.~Hein, ``Provable robustness against all adversarial
  l\_p-perturbations for p$\geq$1.'' in \emph{ICLR}, 2020.

\bibitem{tramer2019adversarial}
F.~Tramer and D.~Boneh, ``Adversarial training and robustness for multiple
  perturbations,'' in \emph{Advances in Neural Information Processing Systems},
  2019, pp. 5866--5876.

\bibitem{jia2019certified}
J.~Jia, X.~Cao, B.~Wang, and N.~Z. Gong, ``Certified robustness for top-k
  predictions against adversarial perturbations via randomized smoothing,''
  \emph{ICLR}, 2020.

\bibitem{cao2017mitigating}
X.~Cao and N.~Z. Gong, ``Mitigating evasion attacks to deep neural networks via
  region-based classification,'' in \emph{Proceedings of the 33rd Annual
  Computer Security Applications Conference}, 2017, pp. 278--287.

\bibitem{zhai2020macer}
R.~Zhai, C.~Dan, D.~He, H.~Zhang, B.~Gong, P.~Ravikumar, C.-J. Hsieh, and
  L.~Wang, ``Macer: Attack-free and scalable robust training via maximizing
  certified radius,'' \emph{ICLR}, 2020.

\bibitem{fischer2020certified}
M.~Fischer, M.~Baader, and M.~Vechev, ``Certified defense to image
  transformations via randomized smoothing,'' \emph{NeurIPS}, 2020.

\bibitem{levine2020randomized}
A.~Levine and S.~Feizi, ``(de) randomized smoothing for certifiable defense
  against patch attacks,'' \emph{arXiv preprint arXiv:2002.10733}, 2020.

\bibitem{chiang2020certified}
P.-y. Chiang, R.~Ni, A.~Abdelkader, C.~Zhu, C.~Studor, and T.~Goldstein,
  ``Certified defenses for adversarial patches,'' \emph{ICLR}, 2020.

\bibitem{awasthi2020adversarial}
P.~Awasthi, H.~Jain, A.~S. Rawat, and A.~Vijayaraghavan, ``Adversarial
  robustness via robust low rank representations,'' \emph{NeurIPS}, 2020.

\bibitem{zhang2020black}
D.~Zhang, M.~Ye, C.~Gong, Z.~Zhu, and Q.~Liu, ``Black-box certification with
  randomized smoothing: A functional optimization based framework,''
  \emph{arXiv preprint arXiv:2002.09169}, 2020.

\bibitem{rahnama2020robust}
A.~Rahnama, A.~T. Nguyen, and E.~Raff, ``Robust design of deep neural networks
  against adversarial attacks based on lyapunov theory,'' in \emph{Proceedings
  of the IEEE/CVF Conference on Computer Vision and Pattern Recognition}, 2020,
  pp. 8178--8187.

\bibitem{saralajew2020fast}
S.~Saralajew, L.~Holdijk, T.~Villmann, and U.~Mittweida, ``Fast adversarial
  robustness certification of nearest prototype classifiers for arbitrary
  seminorms,'' \emph{Advances in Neural Information Processing Systems},
  vol.~33, 2020.

\bibitem{zhang2020tightness}
R.~Y. Zhang, ``On the tightness of semidefinite relaxations for certifying
  robustness to adversarial examples,'' \emph{arXiv preprint arXiv:2006.06759},
  2020.

\bibitem{goldwasser2020beyond}
S.~Goldwasser, A.~T. Kalai, Y.~T. Kalai, and O.~Montasser, ``Beyond
  perturbations: Learning guarantees with arbitrary adversarial test
  examples,'' 2020.

\bibitem{salman2020denoised}
H.~Salman, M.~Sun, G.~Yang, A.~Kapoor, and J.~Z. Kolter, ``Denoised smoothing:
  A provable defense for pretrained classifiers,'' \emph{Advances in Neural
  Information Processing Systems}, vol.~33, 2020.

\bibitem{Awasthi_2021_CVPR}
P.~Awasthi, G.~Yu, C.-S. Ferng, A.~Tomkins, and D.-C. Juan, ``Adversarial
  robustness across representation spaces,'' in \emph{Proceedings of the
  IEEE/CVF Conference on Computer Vision and Pattern Recognition (CVPR)}, June
  2021, pp. 7608--7616.

\bibitem{cemgil2019adversarially}
T.~Cemgil, S.~Ghaisas, K.~D. Dvijotham, and P.~Kohli, ``Adversarially robust
  representations with smooth encoders,'' in \emph{International Conference on
  Learning Representations}, 2019.

\bibitem{he2020defending}
Z.~He, A.~S. Rakin, J.~Li, C.~Chakrabarti, and D.~Fan, ``Defending and
  harnessing the bit-flip based adversarial weight attack,'' in
  \emph{Proceedings of the IEEE/CVF Conference on Computer Vision and Pattern
  Recognition}, 2020, pp. 14\,095--14\,103.

\bibitem{carbone2020robustness}
G.~Carbone, M.~Wicker, L.~Laurenti, A.~Patane, L.~Bortolussi, and
  G.~Sanguinetti, ``Robustness of bayesian neural networks to gradient-based
  attacks,'' \emph{arXiv preprint arXiv:2002.04359}, 2020.

\bibitem{sharmin2020inherent}
S.~Sharmin, N.~Rathi, P.~Panda, and K.~Roy, ``Inherent adversarial robustness
  of deep spiking neural networks: Effects of discrete input encoding and
  non-linear activations,'' in \emph{European Conference on Computer
  Vision}.\hskip 1em plus 0.5em minus 0.4em\relax Springer, 2020, pp. 399--414.

\bibitem{zhang2020gnnguard}
X.~Zhang and M.~Zitnik, ``Gnnguard: Defending graph neural networks against
  adversarial attacks,'' \emph{NeurIPS}, 2020.

\bibitem{du2019robust}
M.~Du, R.~Jia, and D.~Song, ``Robust anomaly detection and backdoor attack
  detection via differential privacy,'' \emph{ICLR}, 2020.

\bibitem{mummadi2019defending}
C.~K. Mummadi, T.~Brox, and J.~H. Metzen, ``Defending against universal
  perturbations with shared adversarial training,'' in \emph{Proceedings of the
  IEEE/CVF International Conference on Computer Vision}, 2019, pp. 4928--4937.

\bibitem{akhtar2018defense}
N.~Akhtar, J.~Liu, and A.~Mian, ``Defense against universal adversarial
  perturbations,'' in \emph{Proceedings of the IEEE Conference on Computer
  Vision and Pattern Recognition}, 2018, pp. 3389--3398.

\bibitem{zhang2018cost}
X.~Zhang and D.~Evans, ``Cost-sensitive robustness against adversarial
  examples,'' \emph{arXiv preprint arXiv:1810.09225}, 2018.

\bibitem{chen2019improving}
H.-Y. Chen, J.-H. Liang, S.-C. Chang, J.-Y. Pan, Y.-T. Chen, W.~Wei, and D.-C.
  Juan, ``Improving adversarial robustness via guided complement entropy,'' in
  \emph{Proceedings of the IEEE/CVF International Conference on Computer
  Vision}, 2019, pp. 4881--4889.

\bibitem{jia2020robust}
S.~Jia, C.~Ma, Y.~Song, and X.~Yang, ``Robust tracking against adversarial
  attacks,'' in \emph{European Conference on Computer Vision}.\hskip 1em plus
  0.5em minus 0.4em\relax Springer, 2020, pp. 69--84.

\bibitem{shao2020open}
R.~Shao, P.~Perera, P.~C. Yuen, and V.~M. Patel, ``Open-set adversarial
  defense,'' \emph{ECCV}, 2020.

\bibitem{zhou2020manifold}
J.~Zhou, C.~Liang, and J.~Chen, ``Manifold projection for adversarial defense
  on face recognition,'' in \emph{European Conference on Computer
  Vision}.\hskip 1em plus 0.5em minus 0.4em\relax Springer, 2020, pp. 288--305.

\bibitem{goldblum2020adversarially}
M.~Goldblum, L.~Fowl, and T.~Goldstein, ``Adversarially robust few-shot
  learning: A meta-learning approach,'' \emph{Advances in Neural Information
  Processing Systems}, vol.~33, 2020.

\bibitem{orekondy2019prediction}
T.~Orekondy, B.~Schiele, and M.~Fritz, ``Prediction poisoning: Towards defenses
  against dnn model stealing attacks,'' in \emph{International Conference on
  Learning Representations}, 2019.

\bibitem{kariyappa2020defending}
S.~Kariyappa and M.~K. Qureshi, ``Defending against model stealing attacks with
  adaptive misinformation,'' in \emph{Proceedings of the IEEE/CVF Conference on
  Computer Vision and Pattern Recognition}, 2020, pp. 770--778.

\bibitem{sulam2020adversarial}
J.~Sulam, R.~Muthumukar, and R.~Arora, ``Adversarial robustness of supervised
  sparse coding,'' \emph{arXiv preprint arXiv:2010.12088}, 2020.

\bibitem{yang2020multitask}
J.~Yang and C.~Vondrick, ``Multitask learning strengthens adversarial
  robustness,'' in \emph{European Conference on Computer Vision}.\hskip 1em
  plus 0.5em minus 0.4em\relax Springer, 2020.

\bibitem{kim2020modeling}
E.~Kim, J.~Rego, Y.~Watkins, and G.~T. Kenyon, ``Modeling biological immunity
  to adversarial examples,'' in \emph{Proceedings of the IEEE/CVF Conference on
  Computer Vision and Pattern Recognition}, 2020, pp. 4666--4675.

\bibitem{wang2020once}
H.~Wang, T.~Chen, S.~Gui, T.-K. Hu, J.~Liu, and Z.~Wang, ``Calibratable
  adversarial training: In-situ tradeoff between robustness and accuracy for
  free,'' \emph{NeurIPS}, 2020.

\bibitem{weng2020trade}
C.-H. Weng, Y.-T. Lee, and S.-H.~B. Wu, ``On the trade-off between adversarial
  and backdoor robustness,'' \emph{Advances in Neural Information Processing
  Systems}, vol.~33, 2020.

\bibitem{chen2020anti}
H.~Chen, B.~Zhang, S.~Xue, X.~Gong, H.~Liu, R.~Ji, and D.~Doermann,
  ``Anti-bandit neural architecture search for model defense,'' in
  \emph{European Conference on Computer Vision}.\hskip 1em plus 0.5em minus
  0.4em\relax Springer, 2020, pp. 70--85.

\bibitem{Wu_2021_CVPR}
T.~Wu, Z.~Liu, Q.~Huang, Y.~Wang, and D.~Lin, ``Adversarial robustness under
  long-tailed distribution,'' in \emph{Proceedings of the IEEE/CVF Conference
  on Computer Vision and Pattern Recognition (CVPR)}, June 2021, pp.
  8659--8668.

\bibitem{jalwana2021cameras}
M.~A. Jalwana, N.~Akhtar, M.~Bennamoun, and A.~Mian, ``Cameras: Enhanced
  resolution and sanity preserving class activation mapping for image
  saliency,'' in \emph{Proceedings of the IEEE/CVF Conference on Computer
  Vision and Pattern Recognition}, 2021.

\bibitem{Yu_2021_CVPRLAFEAT}
Y.~Yu, X.~Gao, and C.-Z. Xu, ``Lafeat: Piercing through adversarial defenses
  with latent features,'' in \emph{Proceedings of the IEEE/CVF Conference on
  Computer Vision and Pattern Recognition (CVPR)}, June 2021, pp. 5735--5745.

\bibitem{khan2021transformers}
S.~Khan, M.~Naseer, M.~Hayat, S.~W. Zamir, F.~S. Khan, and M.~Shah,
  ``Transformers in vision: A survey,'' \emph{arXiv preprint arXiv:2101.01169},
  2021.

\bibitem{liu2017trojaning}
Y.~Liu, S.~Ma, Y.~Aafer, W.-C. Lee, J.~Zhai, W.~Wang, and X.~Zhang, ``Trojaning
  attack on neural networks,'' 2017.

\end{thebibliography}





%







\end{document}